\documentclass[a4paper,11pt]{article}
\usepackage{jheppub} 
\usepackage{float}
\usepackage{placeins}
\usepackage{microtype}
\usepackage{subcaption}

\usepackage{xcolor}
\usepackage{ulem}
\definecolor{addcolor}{RGB}{34, 139, 34}
\definecolor{delcolor}{RGB}{178, 34, 34}

\arxivnumber{2503.22498}

\title{Learnable cut flow for high energy physics}

\author{Jing Li}
\author{and Hao Sun}
\affiliation{
  Institute of Theoretical Physics, School of Physics, Dalian University of Technology, \\
  No.2 Linggong Road, Dalian, Liaoning, 116024, P.R.China
}

\emailAdd{jingliphd@mail.dlut.edu.cn, haosun@dlut.edu.cn}

\abstract{
  Neural networks have emerged as a powerful paradigm for tasks in high energy physics, yet their opaque training process renders them as a black box. In contrast, the traditional cut flow method offers simplicity and interpretability but requires extensive manual tuning to identify optimal cut boundaries.
  To merge the strengths of both approaches, we propose the \textit{Learnable Cut Flow} (LCF), a neural network that transforms the traditional cut selection into a fully differentiable, data-driven process.
  LCF implements two cut strategies—parallel, where observable distributions are treated independently, and sequential, where prior cuts shape subsequent ones—to flexibly determine optimal boundaries.
  Building on this strategy, we introduce the \textit{Learnable Importance}, a metric that quantifies feature importance and adjusts their contributions to the loss accordingly, offering model-driven insights unlike ad-hoc metrics.
  To ensure differentiability, a modified loss function replaces hard cuts with mask operations, preserving data shape throughout the training process.
  LCF is tested on six varied mock datasets and a realistic diboson vs. QCD dataset. Results demonstrate that LCF (1) accurately learns cut boundaries across typical feature distributions in both parallel and sequential strategies, (2) assigns higher importance to discriminative features with minimal overlap, (3) handles redundant or correlated features robustly, and (4) performs effectively in real-world scenarios.
  In the diboson dataset, LCF initially underperforms boosted decision trees and multilayer perceptrons when using all observables.
  LCF bridges the gap between traditional cut flow method and modern black-box neural networks, delivering actionable insights into the training process and feature importance.
  Source code and experimental data are available at \url{https://github.com/Star9daisy/learnable-cut-flow}.
}

\begin{document}
\maketitle
\flushbottom

\section{Introduction}
\label{section:introduction}
At the Large Hadron Collider (LHC), vast data from high-energy collisions are collected, allowing physicists to search for signals of Beyond the Standard Model (BSM) physics against overwhelming background noise.
As a common tool, the cut flow (cut-based) method applies a series of manually defined cuts in sequence to filter rare signal events. While straightforward in eliminating background and preserving signals, it struggles with rigidity as the complexity and number of observables grow.

In contrast to the cut flow method, machine learning is increasingly adopted across the field, proving effective in diverse applications: tagging jets with deep networks and jet charge \cite{1511.05190,1908.08256,2012.08526,2305.04372}, speeding up shower simulations with generative models \cite{2210.06204,2301.08128,2305.17169,2312.00042,2401.02248,2407.16704,2410.21611}, enhancing unfolding for full-event reconstruction \cite{2104.03036,2105.09923,2112.08180,2203.16722,2404.14332}, anomaly detection in complex LHC data \cite{2305.15179,2309.13111,2310.06897,2310.13057,2311.17162,2312.10009,2412.03747}, and suppressing pileup with attention-based methods \cite{2107.02779,2203.15823}.

While machine learning excels in high-energy physics, its opacity makes the reinterpretation of the results far more involved, if not impossible in some cases, driving efforts to align such techniques with physical intuition and traditional methods. One approach embeds symmetry principles into neural networks, yielding architectures like Lorentz-equivariant autoencoders \cite{2212.07347}, PELICAN networks \cite{2211.00454}, energy flow networks \cite{1810.05165,2012.00964}, particle cloud jet tagging \cite{1902.08570}, and others \cite{2203.06153,2212.00832,2411.00446}.
Another approach bridges neural networks with high-level observables, with ref. \cite{1704.08249} leveraging ML to quantify jet information and saturate feature space, ref. \cite{1710.01305} deriving novel jet observables, and ref. \cite{1902.07180} automating observable construction. Additionally, researchers map low-level features to high-level observables \cite{2010.11998} and craft interpretable anomaly detectors for jet substructure \cite{2203.01343}, extracting discriminative insights in a human-readable form. For a comprehensive overview, see ref. \cite{2102.02770}.

To overcome the opacity of neural networks while retaining the clarity of traditional cut-based analyses, we introduce the \textit{Learnable Cut Flow} (LCF). It is a neural network that fully emulates human cut-searching operations, ensuring what it learns is entirely transparent and understandable. Recent efforts have also explored making cuts differentiable for optimization. In ref.~\cite{Hance:2025haf}, cuts are represented as bias parameters in a neural network, with losses that are customized for targeted efficiencies and smoothness. While both approaches share similarities in neural network structure, we adopt a different approach for deriving cut expression and calculating losses, emphasizing parallel and sequential strategies via mask operations, as well as a learnable importance mechanism for feature prioritization.

At bottom, we transform cuts into differentiable operations, such that training a neuron is equivalent to searching for an optimal cut. This match reflects LCF's transparency and interpretability. For scenarios where signals lie in the middle or at both edges of an observable distribution, we employ mask operations to split observable distribution into two parts around a specified center. With such learnable cuts, boundaries are automatically identified after training.

Since observables vary in their separation power, contributing unevenly to classification, we introduce \textit{Learnable Importance}. It is a set of softmax-normalized trainable weights that favors stronger observables while suppressing weaker ones. Unlike other ad-hoc metrics, it not only reflects the contribution fraction but also provides a robust way for selecting key observables to build the final cut flow. At inference, we select results based on importance—retaining cuts from high-importance observables and discarding the rest.

We train the LCF model by rethinking human cut-searching strategies: a parallel approach searches each raw observable distribution independently, while a sequential approach searches the distribution altered by prior cuts. In the parallel strategy, we sum the loss across all cuts. For the sequential strategy, we modify the loss function to incorporate prior cuts. Here, mask operations apply binary weights to filter out contributions from already-cut events. This ensures training aligns with LCF's transparent design.

We demonstrate LCF's effectiveness on dedicated mock datasets and a realistic diboson vs. QCD dataset. Results show it accurately learns cut boundaries in both strategies, prioritizes discriminative features with minimal overlap, handles redundant or correlated features robustly, and performs well in real-world cases.

The paper is structured as follows: section~\ref{section:methodology} explores the methodological foundations, section~\ref{section:experiment_setup} details the experimental setups, section~\ref{section:results} presents results and analysis, and section~\ref{section:conclusion} concludes the study.

\section{Methodology}
\label{section:methodology}

\subsection{Learnable cuts}
\label{subsection:learnable_cuts}
In high-energy physics, the traditional cut flow involves a series of manual steps. First, physicists select relevant observables. Next, each observable is visualized as a one-dimensional distribution. Then, boundaries are chosen by inspecting these distributions to separate signal from background events.

To formalize cuts, consider a dataset with $N$ events ($i$ indexes events) and $F$ observables ($j$ indexes observables). A traditional cut on $x_{ij}$ is
\begin{equation}
  \hat{y}_i = \Theta \left( x_{ij} - c_j \right),
\end{equation}
\begin{equation}
  \hat{y}_i = 1 \rightarrow x_{ij} > c_j.
\end{equation}
Here, $\Theta$ is the Heaviside step function, $c_j$ the cut's boundary, and $\hat{y}_i$ the result. $\hat{y}_i=1$ means the event passes the cut $x_{ij} > c_j$, which is taken as a signal. We use the strictly greater than sign for clarity. This is a fundamental case where signal lies right of $c_j$ as a lower limit.

For the left case, with signal below the cut, the operation becomes
\begin{equation}
  \hat{y}_i = \Theta \left( c_j - x_{ij} \right),
\end{equation}
\begin{equation}
  \hat{y}_i = 1 \rightarrow x_{ij} < c_j.
\end{equation}
Here, $\hat{y}_i = 1$ means the event passes the cut $x_{ij} < c_j$, taken as signal.

To make cuts learnable and auto-updated, we replace the non-differentiable $\Theta$ with the logistic function $\sigma_L$, a smooth approximation commonly used with neural networks:
\begin{equation}
  \sigma_L(z) = \frac{1}{1 + e^{-z}}.
\end{equation}

This enables us to define learnable cut operations for both cases:
\begin{equation}
  \hat{y}_i = \sigma_L \left( x_{ij} - b_j \right),
\end{equation}
\begin{equation}
  \hat{y}_i > t \rightarrow x_{ij} > \sigma_L^{-1}(t) + b_j,
\end{equation}
\begin{equation}
  \hat{y}_i = \sigma_L \left( b_j - x_{ij} \right),
\end{equation}
\begin{equation}
  \hat{y}_i > t \rightarrow x_{ij} < b_j - \sigma_L^{-1}(t).
\end{equation}
Here, $b_j$ is a trainable bias, also the learnable cut value. $t$ is a probability threshold (often 0.5) to flag signal events.

To combine the left and right cases in a single operation, we introduce a trainable weight $w_j$:
\begin{equation}
  \hat{y}_i = \sigma_L \left( w_j x_{ij} - b_j \right).
\end{equation}
The cut's behavior becomes:
\begin{equation}
  \hat{y}_i > t \rightarrow
  \begin{cases}
    x_{ij} > \text{boundary}, & \text{if } \text{direction} > 0 \\
    x_{ij} < \text{boundary}, & \text{if } \text{direction} < 0
  \end{cases},
  \label{eq.single-sided_learnable_cut}
\end{equation}
\begin{equation}
  \text{boundary} = \frac{1}{w_j} \left( \sigma_L^{-1}(t) + b_j \right),
  \label{eq.boundary_of_single-sided_learnable_cut}
\end{equation}
\begin{equation}
  \text{direction} = \text{sign}(w_j).
  \label{eq.direction_of_single-sided_learnable_cut}
\end{equation}
Here, the direction determines whether the signal lies above ($w_j > 0$) or below ($w_j < 0$) the boundary. This transforms a static cut to a dynamic, learnable one, mimicking a neuron's operation. Thus, training a neuron to tune $w_j$ and $b_j$ aligns with finding the optimal boundary value—a clear connection from traditional cuts to neural networks.

For all results presented in Section 4, the threshold $t$ is set to 0.5, ensuring a balanced classification boundary consistent with standard binary classification practices. If we raise its value, through eq.~(\ref{eq.boundary_of_single-sided_learnable_cut}) and eq.~(\ref{eq.direction_of_single-sided_learnable_cut}), $\sigma_L^{-1}(t)$ is increased since this function is monotonically increasing. Now the boundary change depends on the sign of $w_j$: if it's positive (signal on the right), the boundary increases, shrinking the cut range; if it's negative (signal on the left), the boundary decreases, similarly reducing the cut range. Therefore, higher $t$ values impose stricter cuts, potentially decreasing the number of events classified as signal, while lower $t$ values relax the cuts, expanding the signal region but risking increased background events.

Beyond single-sided cuts, double-sided cases—middle case (signal between two boundaries) and edge case (signal outside two boundaries)—are also prevalent. To manage these, we divide the observable's distribution at $\text{center}_j$ (set manually from the distribution) into two ranges and then feed them into two learnable cuts:
\begin{equation}
  \begin{aligned}
    \hat{y}_{i j}^{\text{lower}} &= \sigma_L \left( w_j^{\text{lower}} x_{ij} - b_j^{\text{lower}} \right), \\
    \hat{y}_{i j}^{\text{upper}} &= \sigma_L \left( w_j^{\text{upper}} x_{ij} - b_j^{\text{upper}} \right).
  \end{aligned}
\end{equation}

Accordingly, the loss function shifts as follows. The loss $L_{ij}$ blends the masked average of both cut losses:
\begin{equation}
  L_{ij} = \frac{1}{2} \left( L ^{\text{lower}}_{ij} \operatorname{Mask}_{ij}^{\text{lower}} + L^{\text{upper}}_{ij} \operatorname{Mask}_{ij}^{\text{upper}} \right),
\end{equation}
\begin{equation}
  L_{ij} \left( y_i, \hat{y}_{ij} \right) = - \left[ y_i \log \left( \hat{y}_{ij} \right) + (1 - y_i) \log \left( 1 - \hat{y}_{ij} \right) \right],
\end{equation}
where $L$ is the binary cross-entropy loss between the true label $y_i$ and predicted output $\hat{y}_{ij}$. The masks are defined as:
\begin{equation}
  \text{Mask}^{\text{lower}}_{ij} =
  \begin{cases}
    1, & \text{if } x_{ij} < \text{center}_j \\
    0, & \text{otherwise}
  \end{cases},
\end{equation}
\begin{equation}
  \text{Mask}^{\text{upper}}_{ij} =
  \begin{cases}
    1, & \text{if } x_{ij} > \text{center}_j \\
    0, & \text{otherwise}
  \end{cases}.
\end{equation}

To tie this together, we generalize all four cases (left, right, middle, and edge) using two sets of trainable parameters: $w_j^{\text{lower}}$, $b_j^{\text{lower}}$ for the lower range and $w_j^{\text{upper}}$, $b_j^{\text{upper}}$ for the upper range. These define a learnable cut on each observable, whether it's a single-sided cut (left or right) or a double-sided one (middle or edge). For the simpler left and right cases, a single set of parameters can suffice, as shown in eq.~(\ref{eq.single-sided_learnable_cut}), but the two-set approach fully accommodates the more complex middle and edge scenarios. Based on later experimental results, we summarize the combination rules for the final learned cut:
\begin{itemize}
  \item The specified $\text{center}_j$ separates the lower cut and upper cut. A cut is invalid if its boundary crosses $\text{center}_j$; otherwise, it is valid.
  \item If one cut is valid and the other invalid, the final cut follows the valid cut's boundary and direction.
  \item If both cuts are valid, we check their directions. If they align, the final cut is the intersection of both cuts' ranges. If they differ, a positive lower ($w_j^{\text {lower }}>0$) and negative upper ($w_j^{\text {upper}} < 0$) yield the middle case; otherwise, the edge case.
  \item If both cuts are invalid, single-sided cuts (left or right) remain equivalent to their valid forms. Double-sided cuts are always taken as the edge case, forming the union of ranges to bypass observables with poor separation.
\end{itemize}

With this framework, we complete the transition from traditional cuts to fully learnable cuts.

\subsection{Learnable importance}
\label{subsection:learable_importance}
As the number of observables grows, determining their optimal combination becomes increasingly difficult. Unlike neural networks, which adapt to input features flexibly, traditional cut flows are sensitive to the order and selection of observables. Usually, highly-correlated observables are dropped since they function similarly with each other. Those that have a large overlap area between signal and background are also dropped since they don't provide distinct characteristics between the two classes. We aim to let the LCF model to learn such an importance during the training process by providing a proper tool.

To achieve this, we introduce the \textit{Learnable Importance} for each observable. For the $j$-th observable, we define a trainable parameter $s_j$, and the importance score $s'_j$ is computed using the softmax function:
\begin{equation}
  s'_j = \sigma_S(\mathbf{s})_j = \frac{e^{s_j}}{\sum_{k=1}^{F} e^{s_k}}.
\end{equation}
Here $\mathbf{s}=\left[s_1, s_2, \ldots, s_F\right]$ is the vector of trainable parameters across all observables, $\sigma_S$ denotes the softmax function, and $F$ is the number of observables. While softmax is not the only option—any differentiable function ensuring $\sum_{j=1}^F s_j^{\prime}=1$ could work—its prevalence in machine learning makes it a natural choice.

This importance score acts as a scaling factor for each observable's contribution. For a data point  $x_{ij}$, we define a scaled input:
\begin{equation}
  x_{i j}^{\prime}=s_j^{\prime} x_{i j},
\end{equation}
which feeds into the learnable cut operation:
\begin{equation}
  \hat{y}_{i j}=\sigma_L\left(w_j x_{i j}^{\prime}-b_j\right).
\end{equation}
Here, a higher $s_j^{\prime}$ retains a more complete range, emphasizing the correspondent observable's role in signal-background separation, while a lower $s_j^{\prime}$ narrows it, reducing its influence when separation is poor due to highly-overlapped distributions.

To fully understand this mechanism, we analyze the gradient descent process, focusing on a single cut per observable for simplicity (noting that the two-cut setup from section~\ref{subsection:learnable_cuts} follows similarly). Using binary cross-entropy as the loss function, the total loss across $N$ events and $F$ observables is:
\begin{equation}
  L=\frac{1}{N} \sum_{i=1}^N \sum_{j=1}^F L_{i j},
\end{equation}
where:
\begin{equation}
  L_{ij} = - [y_i \log(\hat{y}_{ij}) + (1 - y_i) \log(1 - \hat{y}_{ij})],
\end{equation}
\begin{equation}
  \hat{y}_{ij} = \sigma_L(z_{ij}),
\end{equation}
\begin{equation}
  z_{ij} = w_j x'_{ij} + b_j,
\end{equation}
\begin{equation}
  x'_{ij} = s'_j x_{ij},
\end{equation}
\begin{equation}
  s'_j = \sigma_S(\boldsymbol{s})_j.
\end{equation}

The gradient of $L$ with respect to $s_k$ reveals how importance scores adjust during training. We compute:
\begin{equation}
  \frac{\partial L}{\partial s_k}=\frac{1}{N} \sum_{i=1}^N \sum_{j=1}^F \frac{\partial L_{i j}}{\partial \hat{y}_{i j}} \cdot \frac{\partial \hat{y}_{i j}}{\partial z_{i j}} \cdot \frac{\partial z_{i j}}{\partial s_j^{\prime}} \cdot \frac{\partial s_j^{\prime}}{\partial s_k},
\end{equation}
where:
\begin{equation}
  \frac{\partial L_{i j}}{\partial \hat{y}_{i j}}=\frac{\hat{y}_{i j}-y_i}{\hat{y}_{i j}\left(1-\hat{y}_{i j}\right)},
\end{equation}
\begin{equation}
  \frac{\partial \hat{y}_{i j}}{\partial z_{i j}}=\hat{y}_{i j}\left(1-\hat{y}_{i j}\right),
\end{equation}
\begin{equation}
  \frac{\partial z_{i j}}{\partial s_j^{\prime}}=w_j x_{i j}.
\end{equation}
Since $s'_j$ depends on all $s_k$ via softmax, its gradient is a Jacobian matrix:
\begin{equation}
  \frac{\partial s_j^{\prime}}{\partial s_k}=
  \begin{cases}
    s_j^{\prime}\left(1-s_j^{\prime}\right) & k=j \\
    -s_k^{\prime} s_j^{\prime} & k \neq j
  \end{cases}.
\end{equation}
Define:
\begin{equation}
  \delta_j=\frac{1}{N} \sum_{i=1}^N\left(\hat{y}_{i j}-y_i\right) w_j x_{i j},
\end{equation}
then:
\begin{equation}
  \frac{\partial L}{\partial s_k}=\sum_{j=1}^F \delta_j \cdot \frac{\partial s_j^{\prime}}{\partial s_k}=s_k^{\prime}\left(\delta_k-\sum_{j=1}^F s_j^{\prime} \delta_j\right).
  \label{eq:learning_process_of_importance}
\end{equation}
This gradient drives the learning dynamics. Here, $\delta_k$ reflects the $k$-th observable's misalignment between predictions and labels, while $\sum_{j=1}^F s_j^{\prime} \delta_j$ is a weighted baseline across all observables. If $\delta_k>$ baseline, the gradient is positive, reducing $s_k$ and thus $s_k^{\prime}$. This shrinks the value of the observable towards zero and the $s_k^\prime$ outside the parenthesis reduces its gradient making it rely more on other influential ones.  If $\delta_k<$ baseline, the gradient is negative, increasing $s_k^{\prime}$, boosting a reliable observable to keep the full range of it. This process embeds importance directly into training. Such a learned importance offers insights into each observable's role and plays as a potential threshold for dropping negligible ones.

\subsection{Learnable cut flow}
\label{subsection:learnable_cut_flow}
\begin{figure}[htbp]
  \centering
  \includegraphics[width=1.0\textwidth]{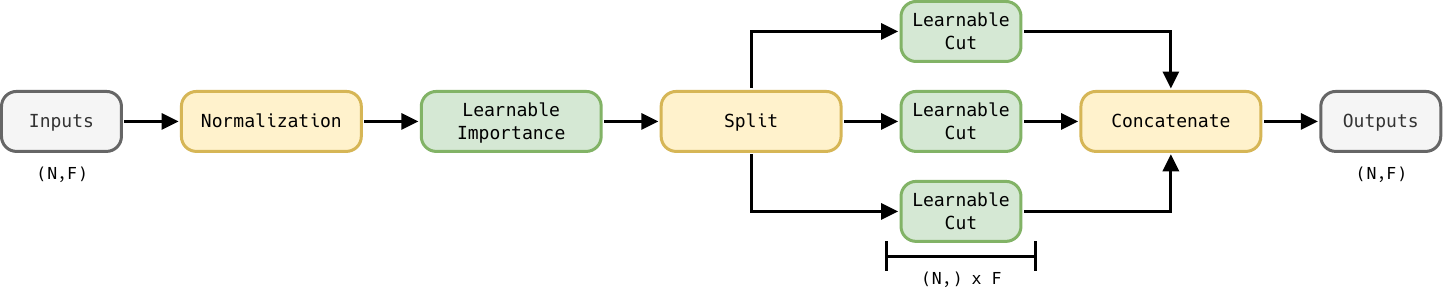}
  \caption{The structure of the learnable cut flow model.}
  \label{figure:learnable_cut_flow}
\end{figure}

With the learnable cuts from section~\ref{subsection:learnable_cuts} and the learnable importance from section~\ref{subsection:learable_importance}, we now build the LCF model on them, as shown in figure~\ref{figure:learnable_cut_flow}.

The model takes the original data as inputs. Then data are normalized with the mean and standard deviation of the training data. Then data are split along the observable axis to feed into each learnable cut. The output of each learnable cut is collected at the end. When the LCF model performs inference on new data, a minimum importance ratio is applied, set to 0.05 by default. This means it automatically ignores cuts with observables below 5\% of the average (defined as the baseline in the context of the learned importance), where the average is defined as 1 divided by the number of features.

There are two strategies commonly adopted in the literature:
\begin{itemize}
  \item \textbf{Parallel strategy}: Each cut is optimized independently, focusing solely on the original data distribution to determine cut values, ignoring inter-observable correlations.
  \item  \textbf{Sequential strategy}: Cut values are optimized sequentially, with each cut adjusting based on the updated data distribution after applying the previous cut, fully considering correlations among different observables.
\end{itemize}

The parallel strategy is straightforward because it independently optimizes each cut, requiring all collected cut results to exceed a threshold at the end of the flow for an event to pass all cuts during optimization.

In the sequential strategy, we cannot change the input data shape during forward propagation, as neurons must preserve the original structure for stacking. Instead, we mask events by setting their contribution to the loss function to zero, effectively excluding them without altering the data shape. We modify the loss per event $i$ and observable $j$ for the sequential strategy as:
\begin{equation}
  L_{ij} = \frac{1}{2} \left(L^{\text{lower}}_{ij} + L^{\text{upper}}_{ij}\right) \cdot \text{Mask}_{i,j-1},
\end{equation}
where $L_{i j}^{\text{lower}}$ and $L_{i j}^{\text{upper}}$ are the binary cross-entropy losses for the two ranges (as in section~\ref{subsection:learnable_cuts}), and $\text{Mask}_{i, j-1}$ is the mask from the previous cut, defined as:
\begin{equation}
  \text{Mask}_{i,j} =
  \begin{cases}
    1 & \text{if } j = 0 \\
    \Theta(\hat{y}^{\text{lower}}_{i,j-1} - t) \cdot \Theta(\hat{y}^{\text{upper}}_{i,j-1} - t) & \text{if } j > 0
  \end{cases}.
\end{equation}
The mask takes the intersection of the outputs from both ranges, ensuring the loss accounts for the effect of the previous cut's decisions.

With these parallel and sequential strategies, we've constructed a learnable cut flow model that fully mimics the traditional approach, enhancing it with the adaptability and interpretability of neural networks.

\section{Experiment setup}
\label{section:experiment_setup}

\subsection{Datasets}
\label{subsection:datasets}
This study employs two types of datasets: a series of synthetic mock datasets to demonstrate the effectiveness of the proposed neural network under controlled and simplified conditions, and a real dataset produced in the context of the studies presented in ref.~\cite{1603.09349} to benchmark its performance against baseline models.

\begin{figure}[htbp]
  \centering
  \includegraphics[width=.32\textwidth]{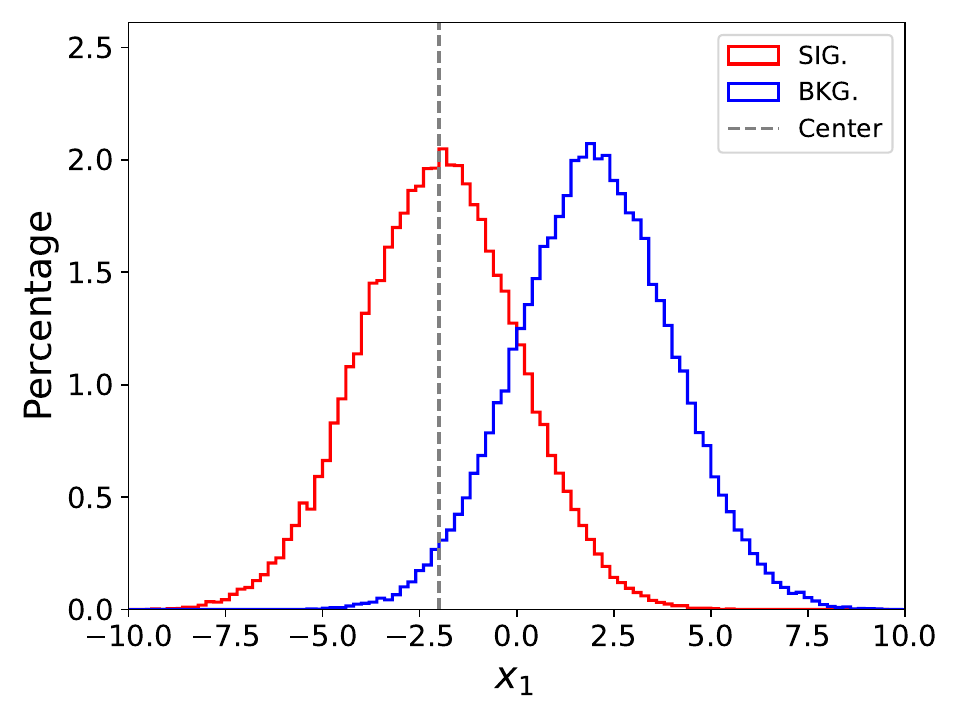}
  \includegraphics[width=.32\textwidth]{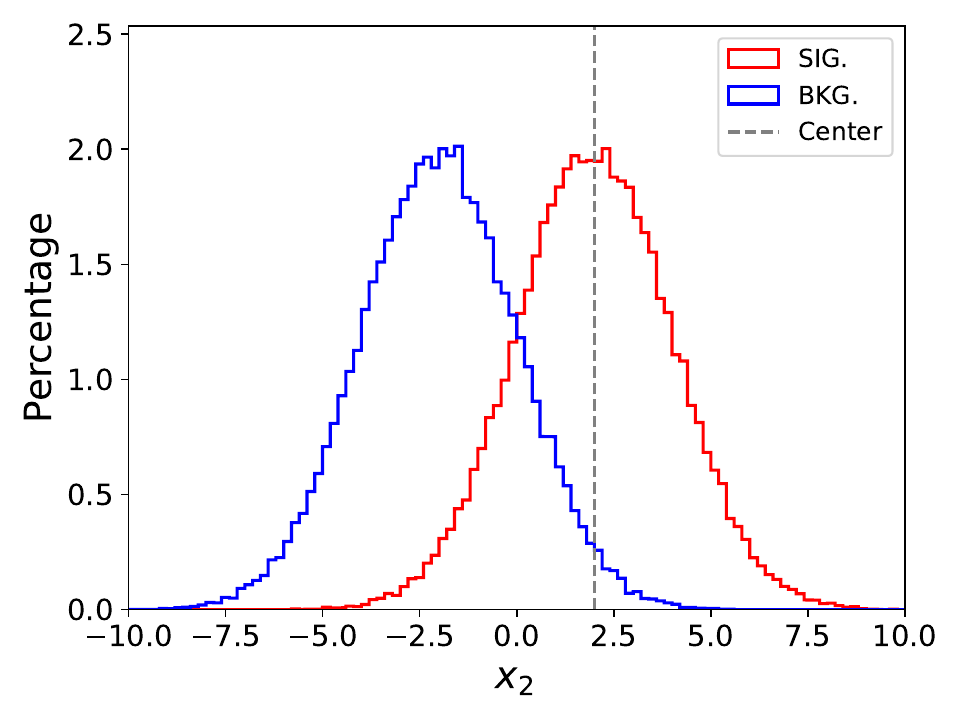}
  \includegraphics[width=.32\textwidth]{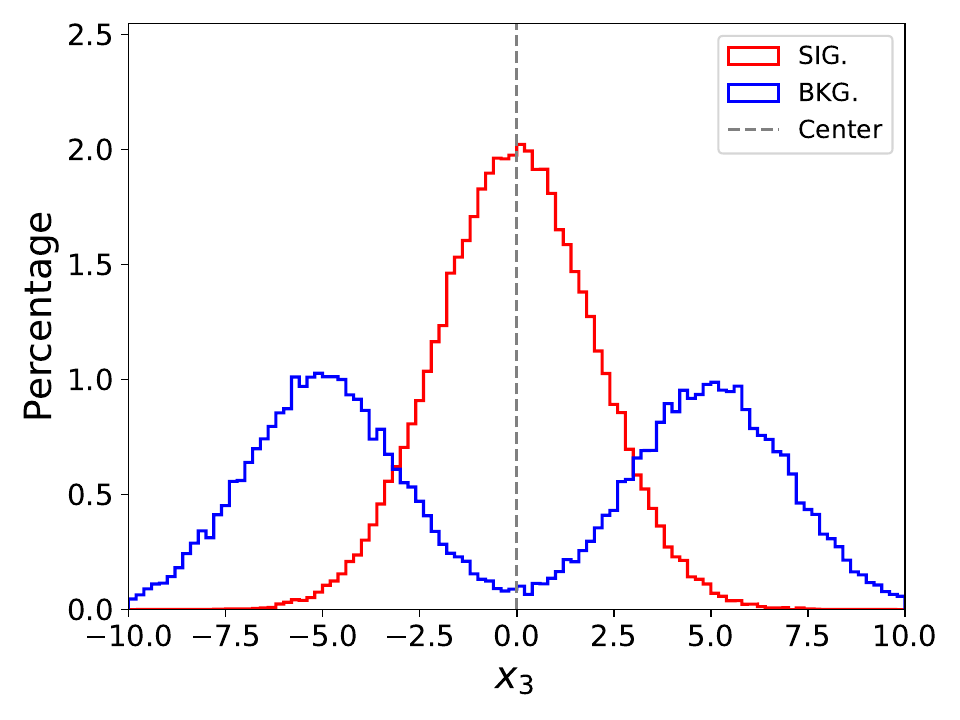}
  \includegraphics[width=.32\textwidth]{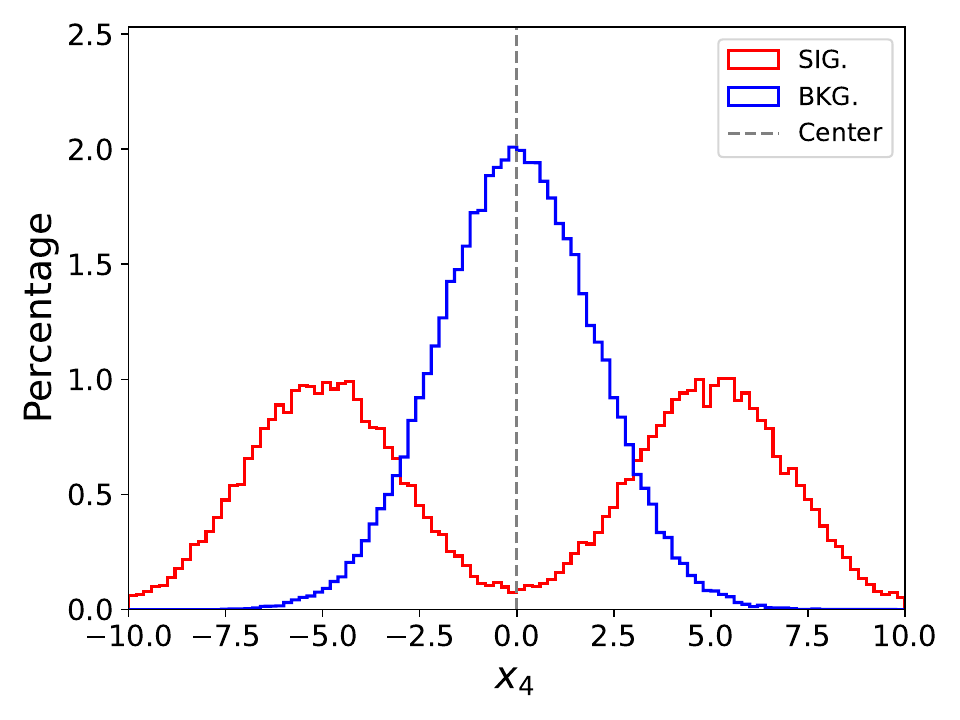}
  \includegraphics[width=.32\textwidth]{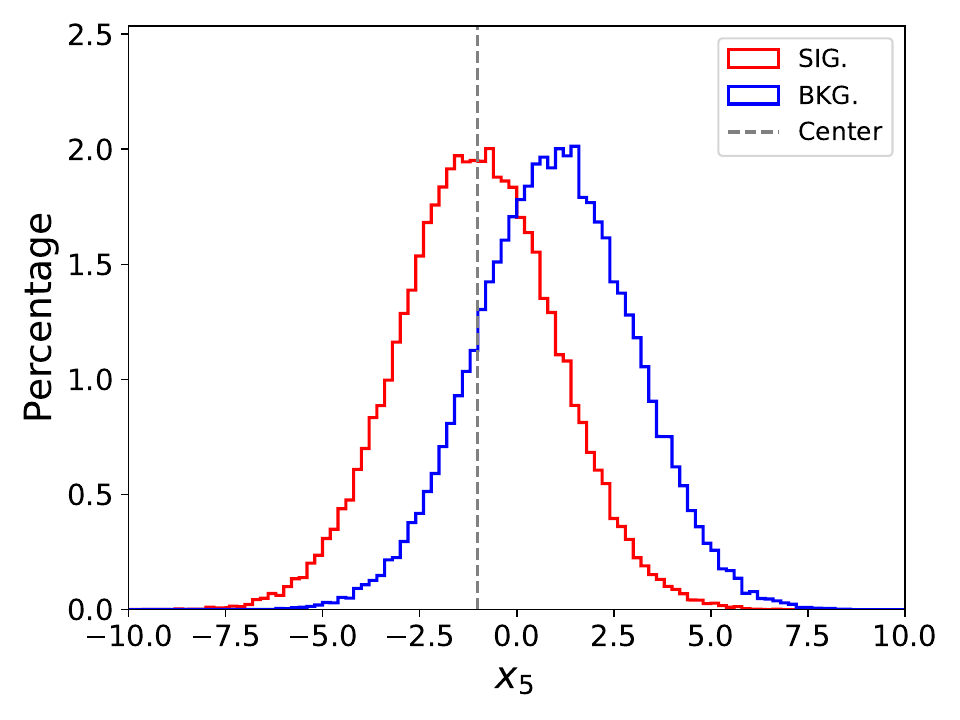}
  \includegraphics[width=.32\textwidth]{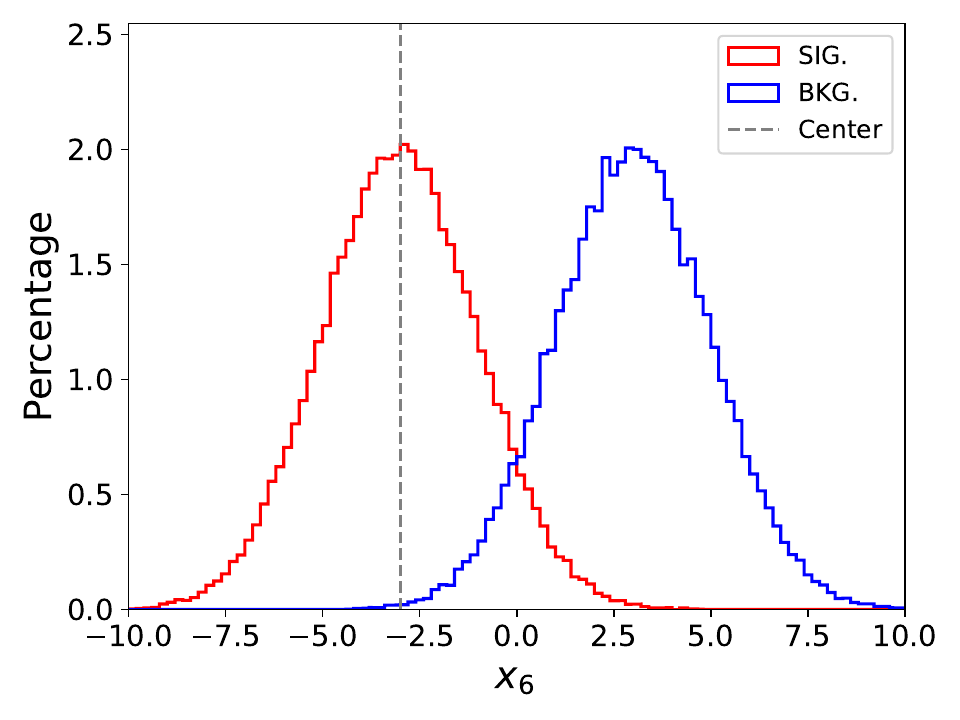}
  \includegraphics[width=.32\textwidth]{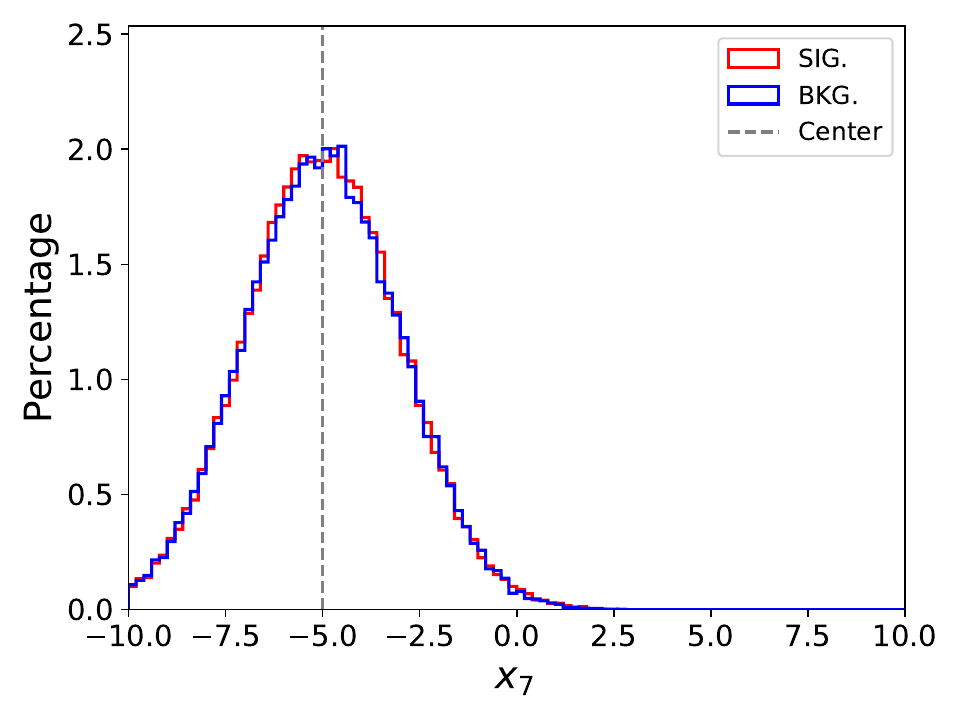}
  \includegraphics[width=.32\textwidth]{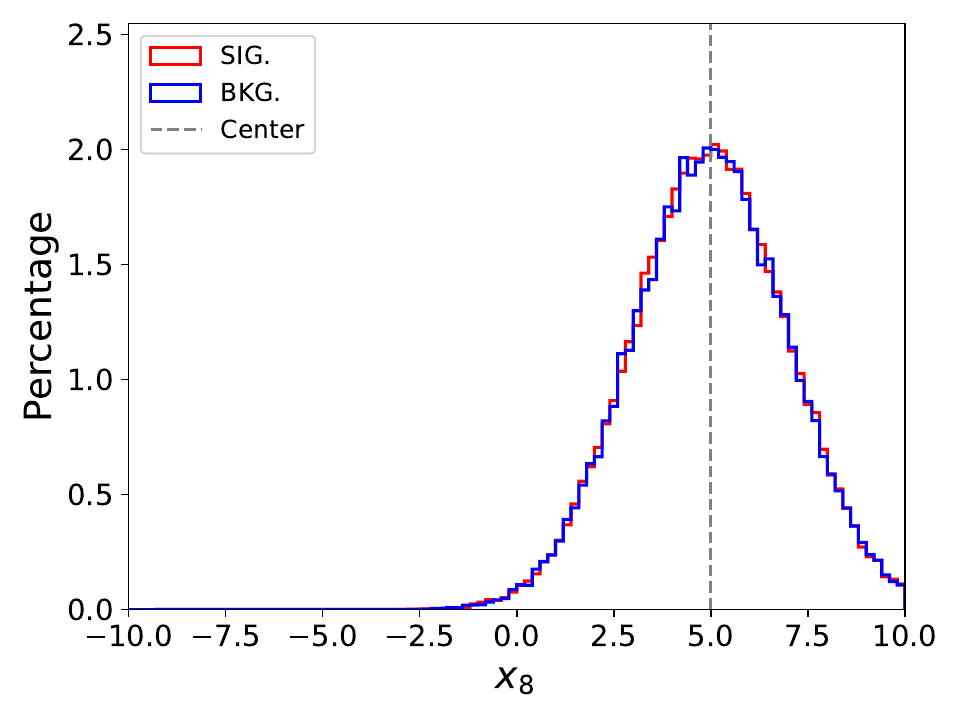}
  \includegraphics[width=.32\textwidth]{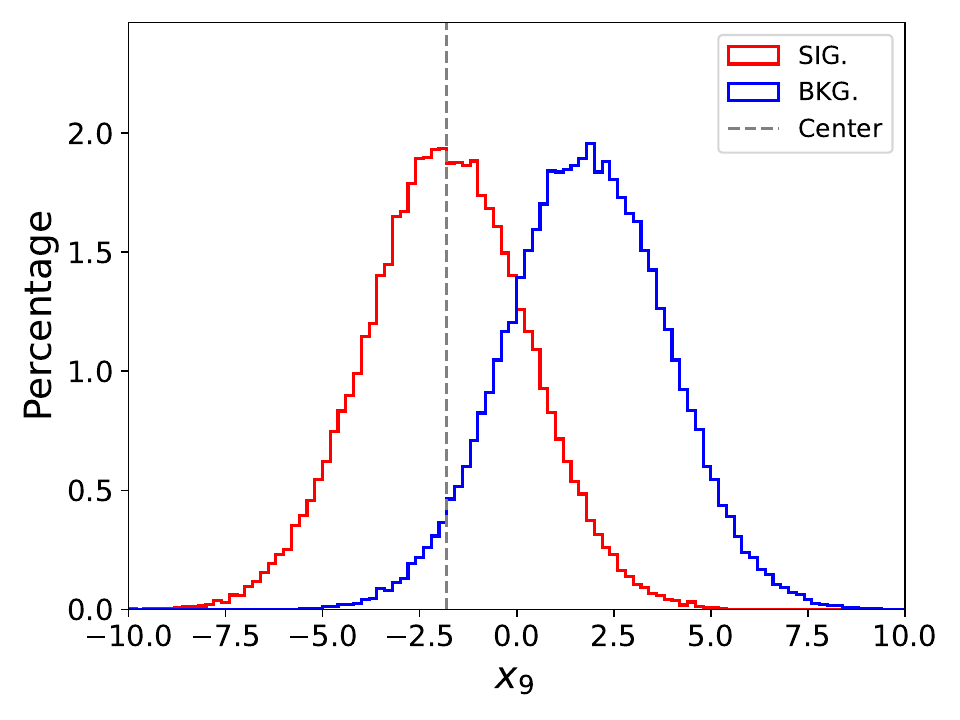}
  \includegraphics[width=.32\textwidth]{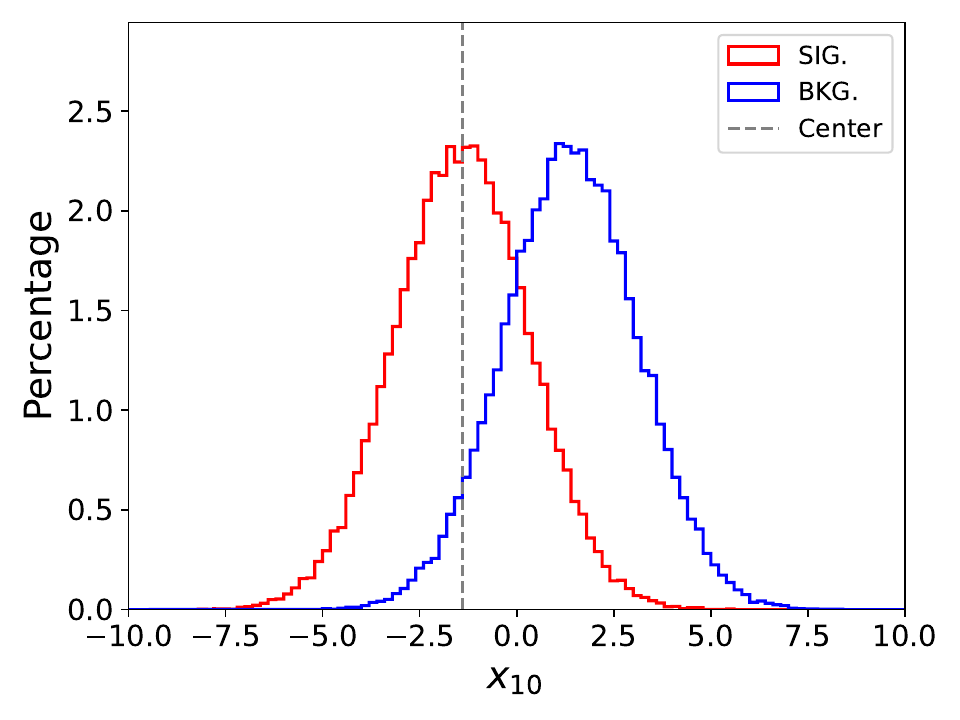}
  \caption{Mock feature distributions from $x_1$ to $x_{10}$. $x_1$, $x_2$, $x_3$, and $x_4$ represent four basic cases of signal relative location. $x_5$ and $x_6$ have weaker and stronger separation compared with $x_1$. $x_7$ and $x_8$ are two redundant features. $x_9$ and $x_{10}$ are two features highly-correlated with $x_1$. The center value divides a distribution into two parts to train two sets of trainable weights to form one learned cut.}
  \label{figure:mock_feature_distributions}
\end{figure}

The synthetic datasets are constructed from different combinations of ten features sampled from distinct normal distributions ($N\left(\mu, \sigma^2\right)$ where $\mu$ is mean and $\sigma$ is the standard deviation). The parameters are specified in table~\ref{table:features} and distributions are presented in figure~\ref{figure:mock_feature_distributions}.

\begin{table}[htbp]
  \centering
  \begin{tabular}{c|c|c|c}
    \hline
    Feature & Signal & Background & Comment \\
    \hline
    $x_1$ & $N(-2,4)$ & $N(2,4)$ & Left case \\
    $x_2$ & $N(2,4)$ & $N(-2,4)$ & Right case \\
    $x_3$ & $N(0,4)$ & $N(5,4)+N(-5,4)$ & Middle case \\
    $x_4$ & $N(5,4)+N(-5,4)$ & $N(0,4)$ & Edge case \\
    $x_5$ & $N(-1,4)$ & $N(1,4)$ & Weakly separated \\
    $x_6$ & $N(-3,4)$ & $N(3,4)$ & Strongly separated \\
    $x_7$ & $N(-5,4)$ & $N(-5,4)$ & Redundant \\
    $x_8$ & $N(5,4)$ & $N(5,4)$ & Redundant \\
    $x_9$ & $0.9 \times x_1^{\text {signal }}+N(0,1)$ & $0.9 \times x_1^{\text {background }}+N(0,1)$ & Highly correlated \\
    $x_{10}$ & $0.7 \times x_1^{\text {signal }}+N(0,1)$ & $0.7 \times x_1^{\text {background }}+N(0,1)$ & Highly correlated \\
    \hline
  \end{tabular}
  \caption{Parameters of signal and background features for synthetic mock datasets.}
  \label{table:features}
\end{table}

\begin{figure}[htbp]
  \centering
  \begin{subfigure}{0.32\textwidth}
    \includegraphics[width=\textwidth]{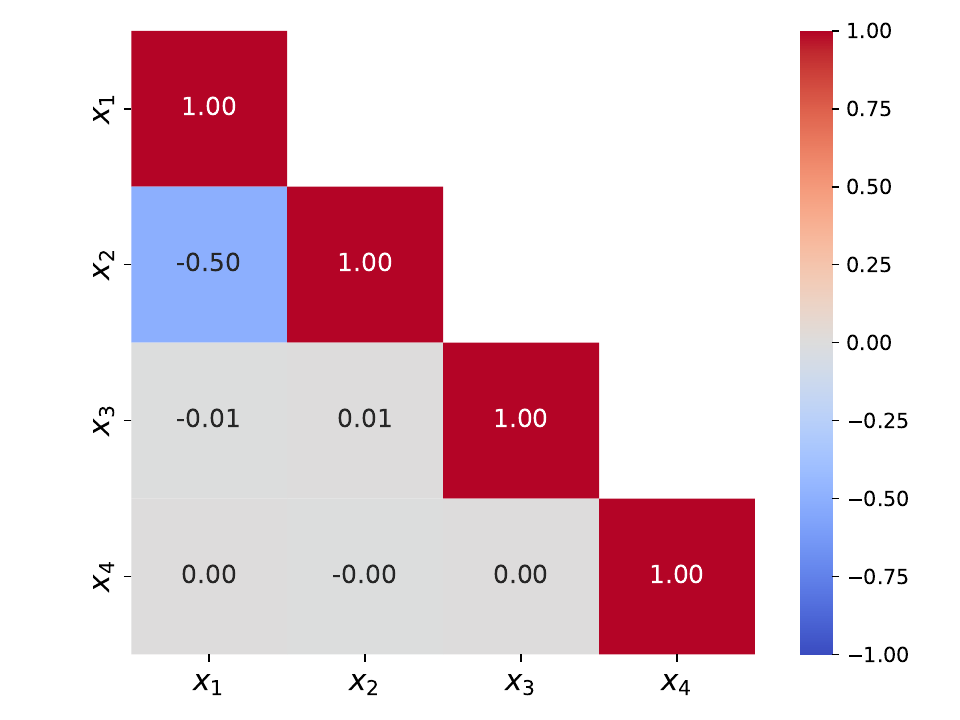}
    \caption{Mock1}
  \end{subfigure}
  \hfill
  \begin{subfigure}{0.32\textwidth}
    \includegraphics[width=\textwidth]{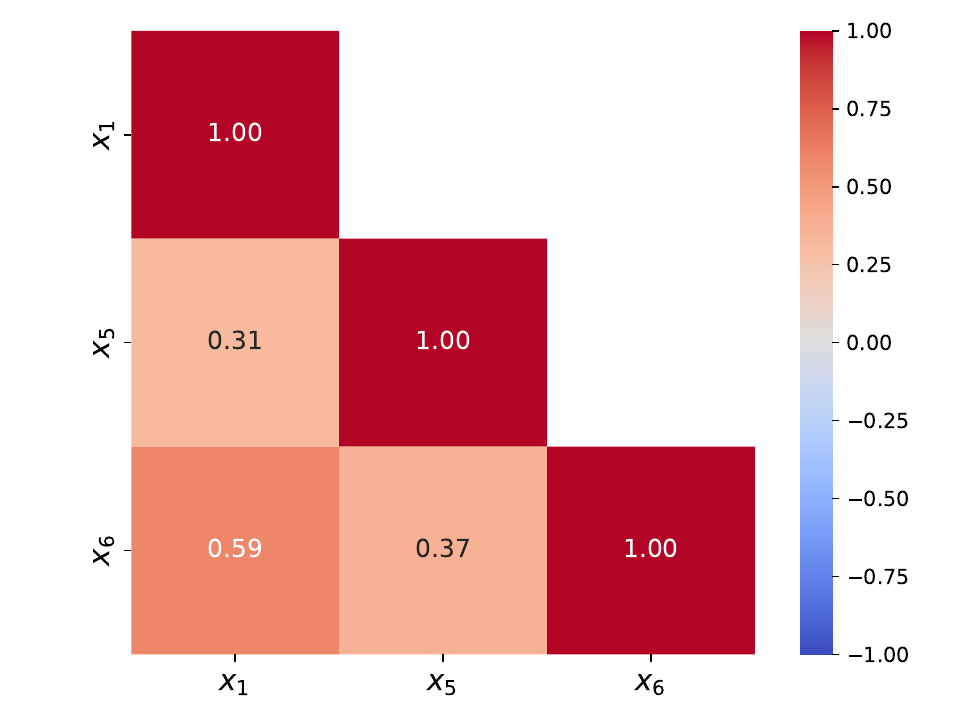}
    \caption{Mock2}
  \end{subfigure}
  \hfill
  \begin{subfigure}{0.32\textwidth}
    \includegraphics[width=\textwidth]{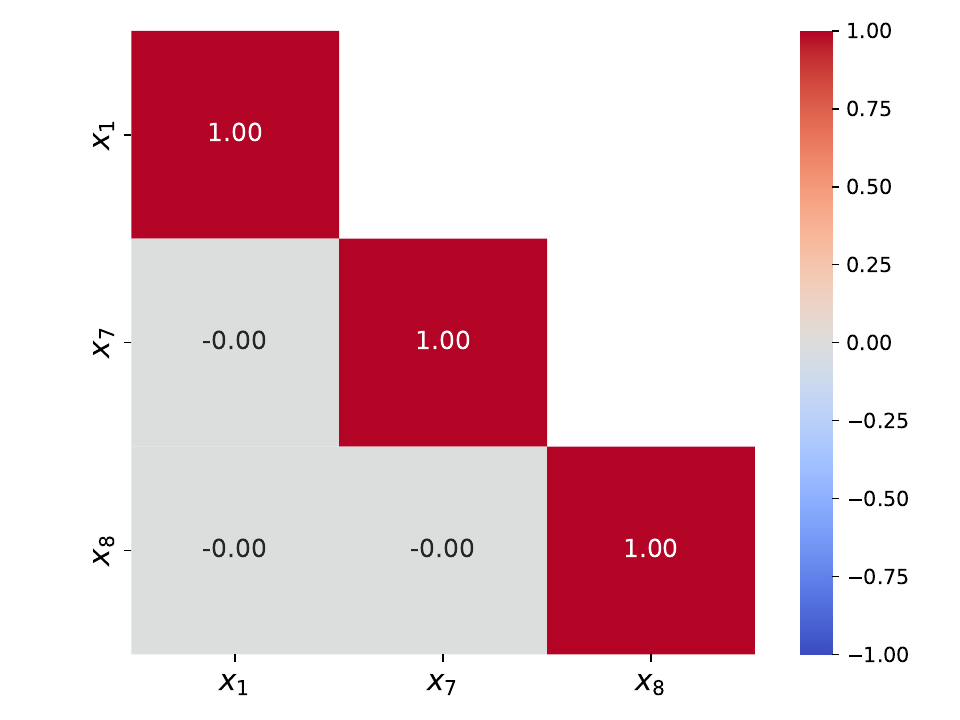}
    \caption{Mock3}
  \end{subfigure}

  \vspace{0.3cm}
  \begin{subfigure}{0.32\textwidth}
    \includegraphics[width=\textwidth]{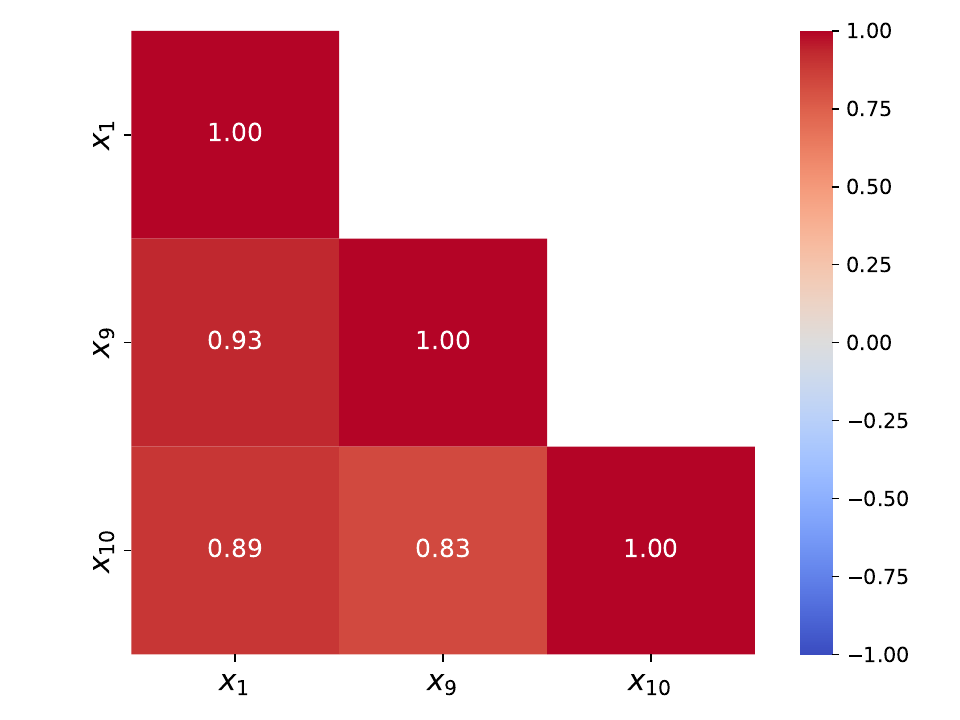}
    \caption{Mock4}
  \end{subfigure}
  \hfill
  \begin{subfigure}{0.32\textwidth}
    \includegraphics[width=\textwidth]{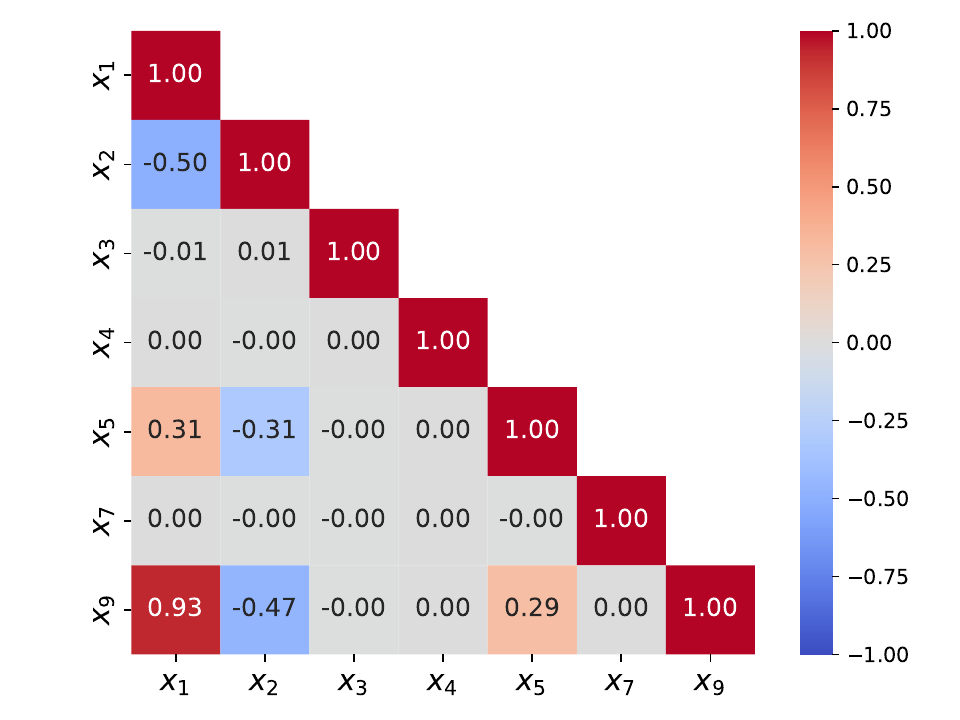}
    \caption{Mock5}
  \end{subfigure}
  \hfill
  \begin{subfigure}{0.32\textwidth}
    \includegraphics[width=\textwidth]{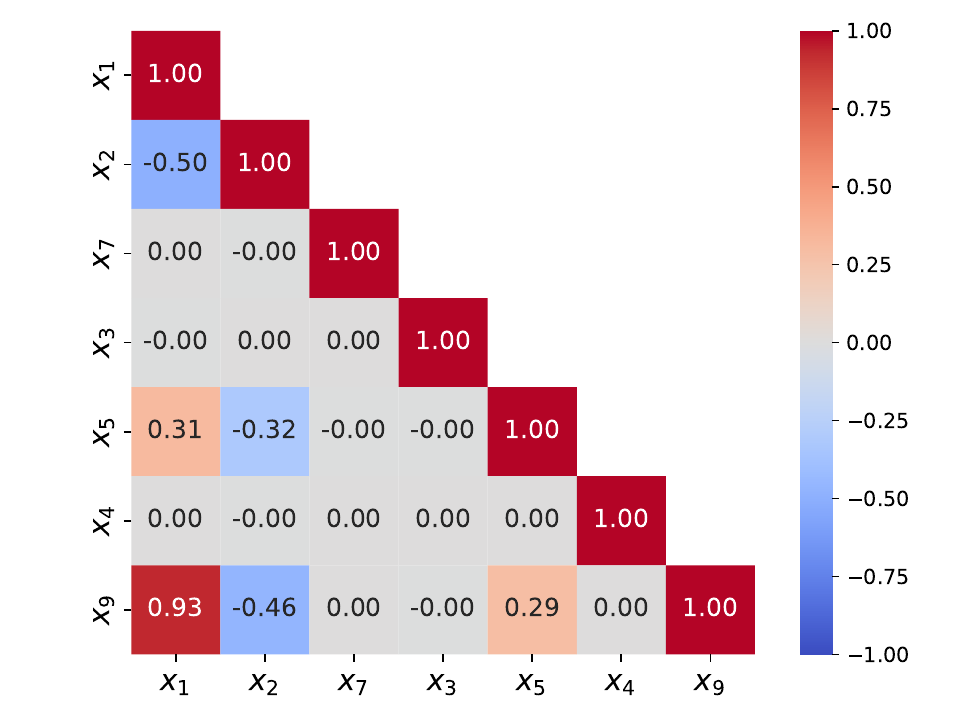}
    \caption{Mock6}
  \end{subfigure}

  \caption{Feature correlations of mock datasets. Mock4 dataset contains two features $x_9$ and $x_{10}$ highly-correlated to $x_1$.}
  \label{figure:mock_correlations}
\end{figure}

Each dataset is designed for a specific purpose and with correlations shown in figure~\ref{figure:mock_correlations}:
\begin{itemize}
  \item \textbf{Mock1} consists of four features ($x_1$, $x_2$, $x_3$, $x_4$) to validate whether the learned decision boundaries are reasonable and intuitive across four fundamental cases: signals located at the left, right, middle, and edge.

  \item \textbf{Mock2} comprises three features ($x_1$, $x_5$, $x_6$) to assess the interpretability of learned feature importance by adjusting the signal-background overlap in $x_1$, with $x_5$ reducing the overlap and $x_6$ increasing it.

  \item \textbf{Mock3} includes two redundant features ($x_7$, $x_8$) alongside $x_1$ to assess the model's robustness to irrelevant inputs.

  \item \textbf{Mock4} combines $x_1$ with two highly correlated features ($x_9$, $x_{10}$) to simulate a common challenge in data analysis, where correlated observables often complicate signal-background separation.

  \item \textbf{Mock5} incorporates a typical set of features ($x_1$, $x_2$, $x_3$, $x_4$, $x_5$, $x_7$, $x_9$) to represent a realistic dataset.

  \item \textbf{Mock6} randomly reorders the features of Mock5 to evaluate the model's sensitivity to feature order. The order of features is $x_1, x_2, x_7, x_3, x_5, x_4, x_9$ since the random seed is fixed.
\end{itemize}

The real dataset was created using standard simulation tools in high energy physics. The signal, diboson production $p p \rightarrow W^{+} W^{-} \rightarrow qqqq$, produces two fat jets exhibiting a 2-prong substructure. In contrast, the background, QCD dijet production $p p \rightarrow qq, qg, gg,$ predominantly yields 1-prong jets.

Events were generated using \textsc{MadGraph} 5 v2.2.3 for hard scattering, \textsc{Pythia} v6.426 for showering and hadronization, and \textsc{Delphes} v3.2.0 for detector response. Jets were reconstructed with the anti-$k_t$ algorithm and parameter $R=1.2$ using \textsc{FastJet} v3.1.2. Their constituents were then reclustered with the $k_t$ algorithm and parameter $R=0.2$ to form subjets. Subjets with $p_T^{\text{sub}} < p_T^{\text{jet}} \times 3\%$ were discarded and the remaining subjets were used to construct the trimmed jets \cite{0912.1342}. To ensure the collinear $W$ bosons decays, only jets with $p_T^{\text{trim}} \in[300,400]$ GeV were retained.

\begin{figure}[htbp]
  \centering
  \includegraphics[width=.45\textwidth]{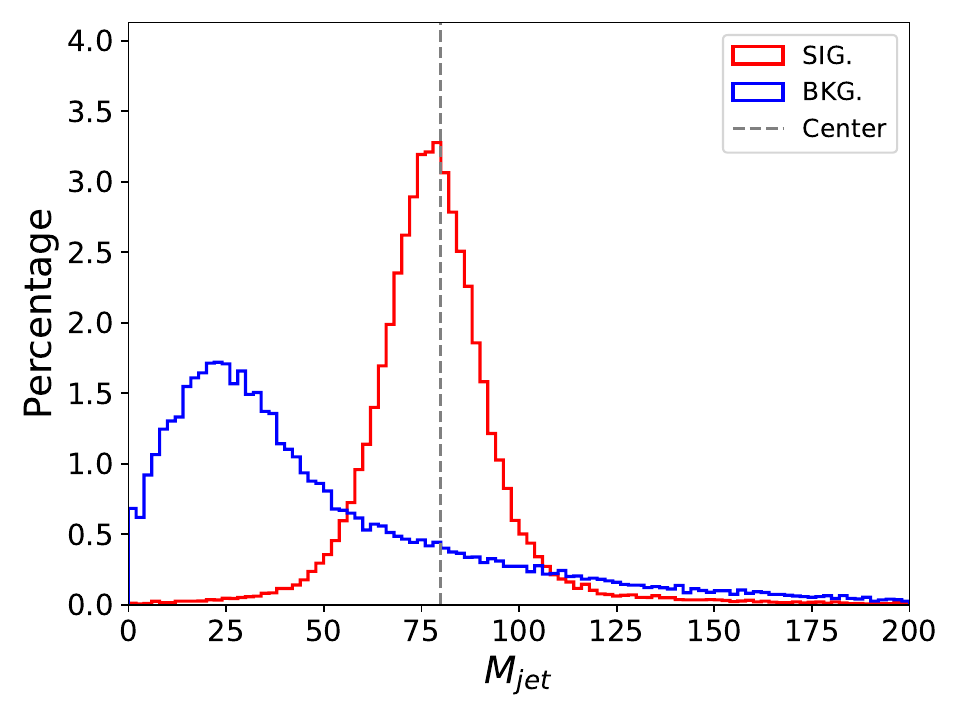}
  \qquad
  \includegraphics[width=.45\textwidth]{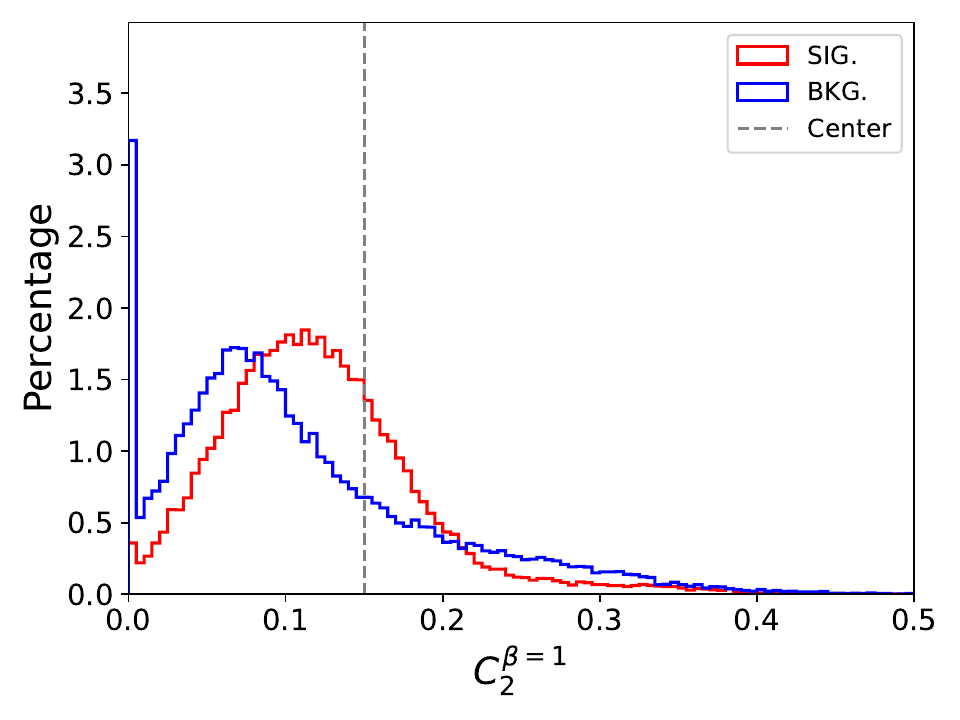}
  \qquad
  \includegraphics[width=.45\textwidth]{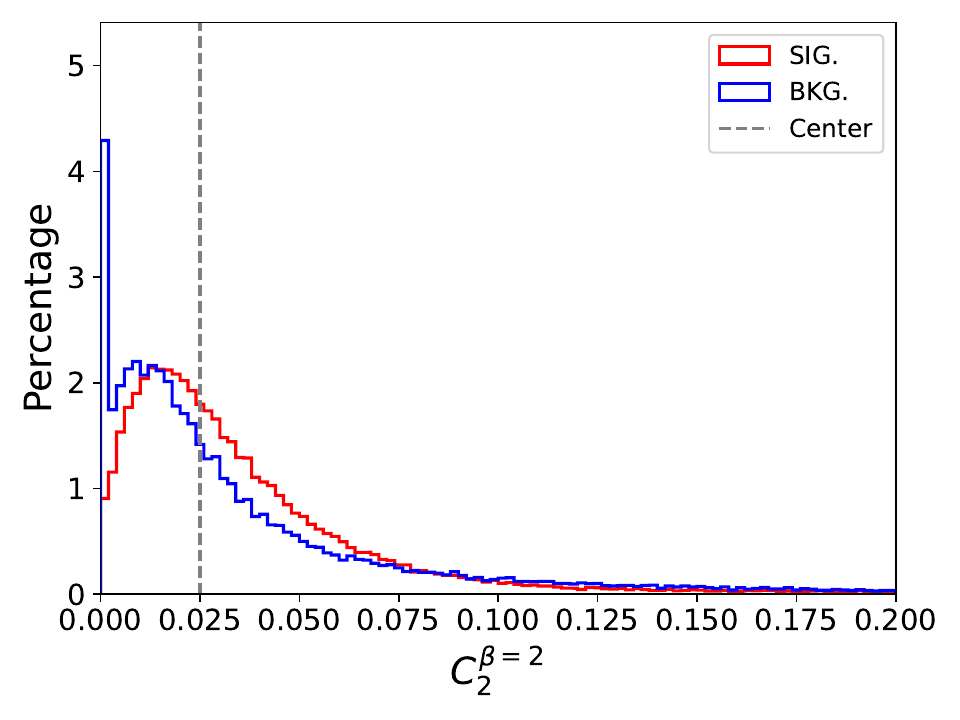}
  \qquad
  \includegraphics[width=.45\textwidth]{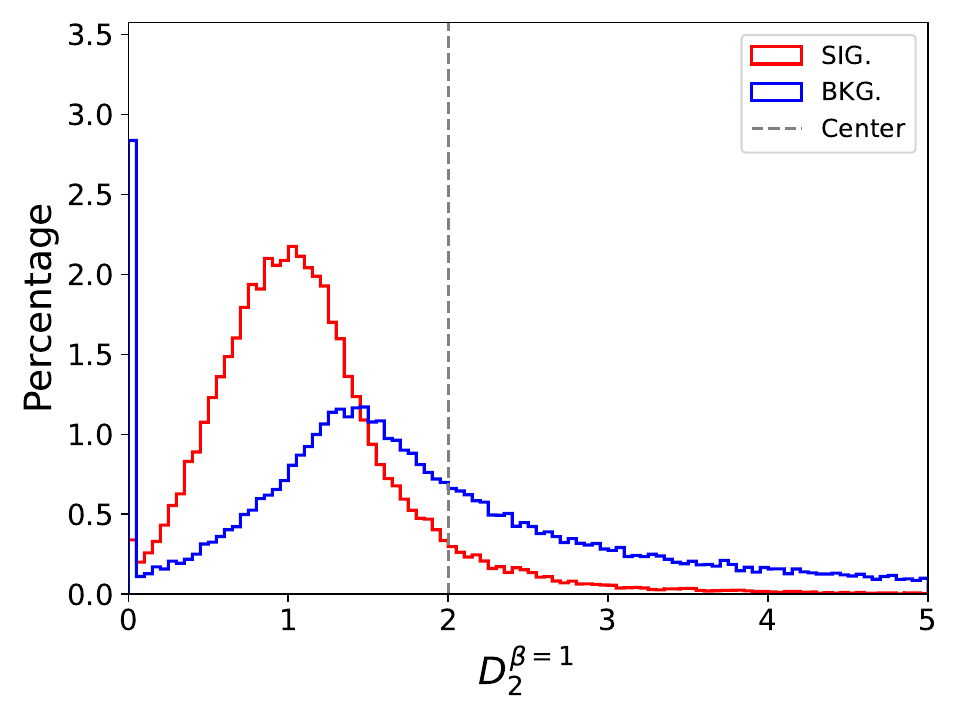}
  \qquad
  \includegraphics[width=.45\textwidth]{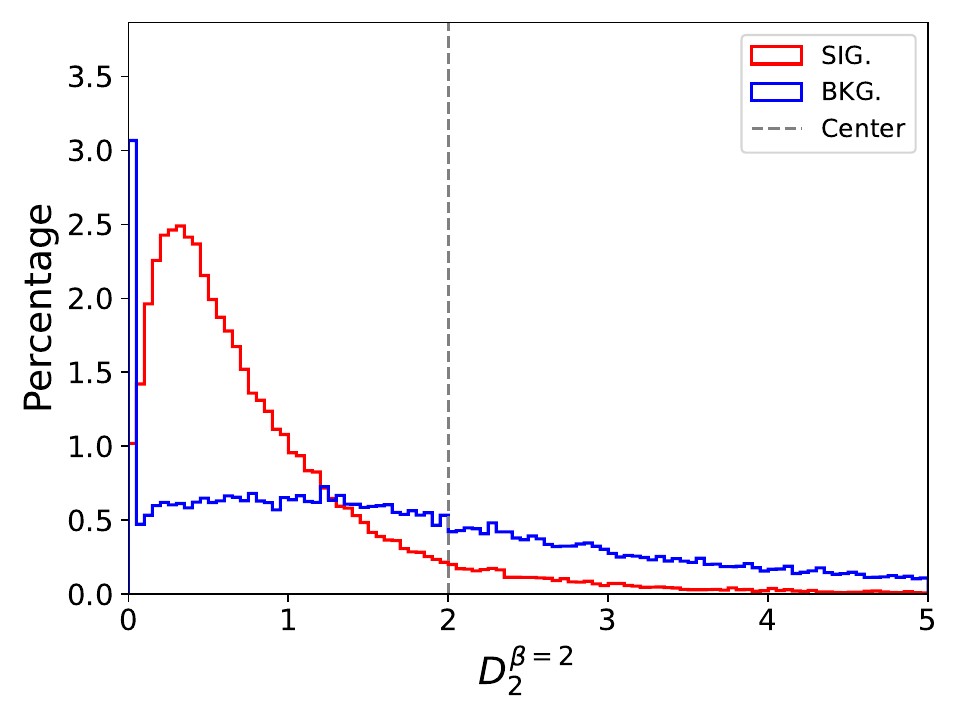}
  \qquad
  \includegraphics[width=.45\textwidth]{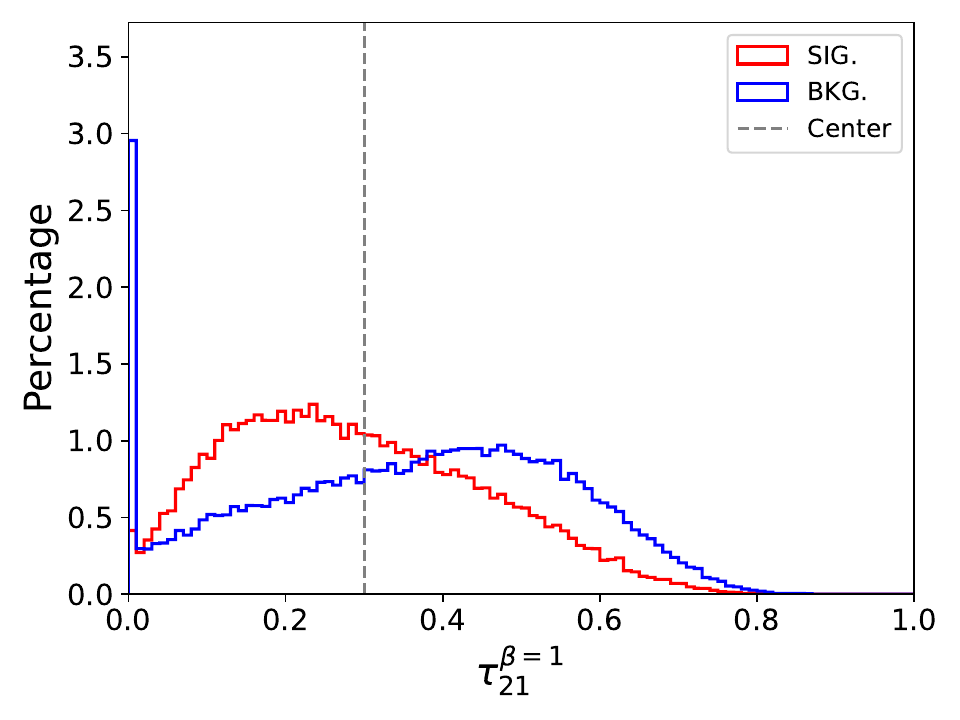}
  \caption{Distributions of jet substructure variables in the real diboson vs. QCD dataset. The center value divides a distribution into two parts to train two sets of trainable weights to form one learned cut.}
  \label{figure:real_observable_distributions}
\end{figure}

\begin{figure}[htbp]
  \centering
  \includegraphics[width=.8\textwidth]{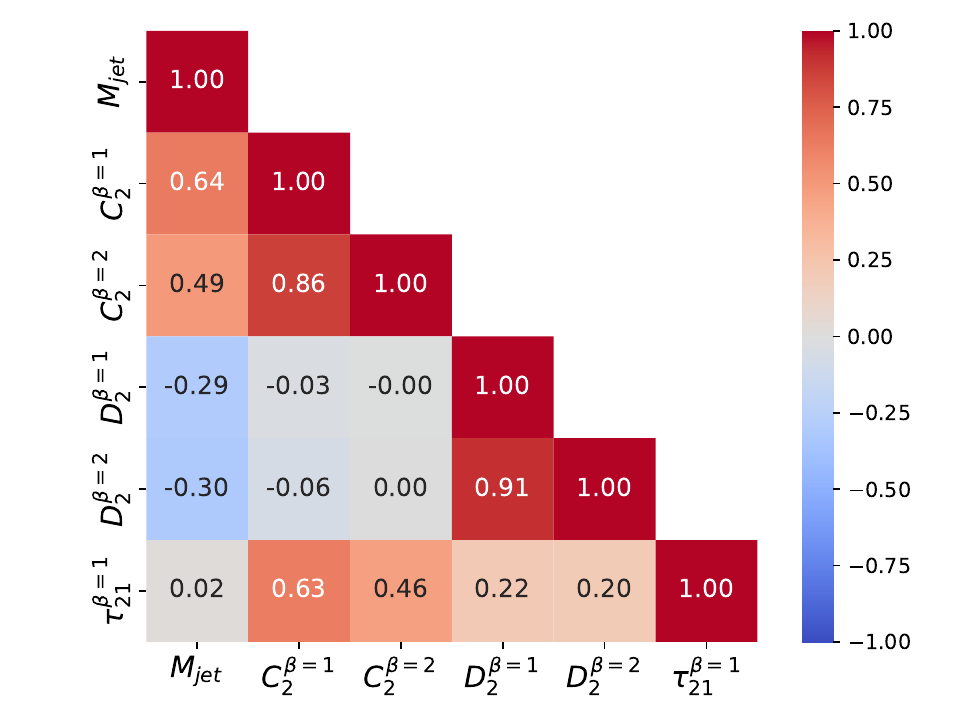}
  \caption{Observable correlations of the diboson dataset.}
  \label{figure:real1_correlations}
\end{figure}

Six high-level observables were calculated: the invariant mass of the trimmed jet ($M_{\text{jet}}$),  the N-subjettiness ratio ($\tau_{21}^{\beta=1}$), and four ratios of energy correlation functions ($C_2^{\beta=1}, C_2^{\beta=2}, D_2^{\beta=1}, D_2^{\beta=2}$). Their distributions are shown in figure~\ref{figure:real_observable_distributions} and their correlations in figure~\ref{figure:real1_correlations}.

Each dataset contains 200,000 events evenly divided into signal and background. The data are split into a training set (50\%) and a test set (50\%). The real dataset, sourced from \cite{1603.09349}, is originally split into training and test sets, each further divided into pile-up and non-pileup subsets. To ensure computational efficiency, we didn't use the full sets; instead, our events are randomly drawn from the original non-pileup test set to reduce computational load while maintaining representativity. Furthermore, the original observable distributions show a significant presence of low values for all observables except $M_{\text{jet}}$. This is a physical effect that arises from the one-pronged nature of QCD background jets. Unlike signal jets from diboson decays, which exhibit multi-pronged substructure, QCD jets typically originate from single quarks or gluons and retain this characteristic even when clustered using algorithms with relatively large R parameters. For further details on the distributions of specifically selected dijet QCD events, see refs.~\cite{1011.2268,1305.0007}. To mitigate outliers and ensure stable training, only events within the 5th to 95th percentile range of all observables are retained. Both dataset types define a binary classification task, allowing the learnable cut flow models to be well trained and provide insights into the neural network's decision-making process.

\subsection{Baseline models and training}
\label{subsection:baseline_models}
Two baseline models are implemented: a boosted decision tree (BDT) and a multilayer perceptron (MLP):
\begin{itemize}
  \item The BDT, built with Scikit-Learn \cite{scikit-learn}, consists of 100 estimators with a maximum depth of 3. It is trained directly on raw input data without preprocessing as it is insensitive to feature scaling. The output represents the probability of the signal event.
  \item The MLP, built with Keras \cite{keras}, includes a standard normalization layer for input preprocessing and five hidden layers with 16, 32, 64, 32, and 16 units respectively. All hidden layers use the rectified linear unit (ReLU) activation function. The output layer applies a sigmoid function to produce the signal probability.
\end{itemize}

All models involved (BDT, MLP, LCF in two strategies) are trained using the same configuration: binary cross entropy as the loss function, Adam as the optimizer with a learning rate 0.001, a batch size of 512, and 200 training epochs. Early stopping is not applied as training of LCF exhibits plateau phases in loss. The centers for the LCF model are set as -2, 2, 0, 0, -1, -3, -5, 5, -1.8, -1.4 for each mock feature, as shown in figure~\ref{figure:mock_feature_distributions}, and 80, 0.15, 0.025, 2, 2, 0.3 for each observable in the diboson dataset, as shown in figure \ref{figure:real_observable_distributions}. Each center is chosen to divide the respective observable's distribution into two regions, enabling the LCF to learn boundaries on one side (for left or right cases) or both sides (for middle or edge cases). Models are trained continuously ten times to demonstrate robustness and stability.

\subsection{Performance evaluation}
\label{subsection:performance_evaluation}
To comprehensively assess the performance of the LCF model and compare it with baseline models, we employ the following metrics, evaluated on the test set, that capture different aspects of classification performance. For each metric, higher values are preferable unless otherwise noted:

\begin{itemize}
  \item \textbf{True Positive (TP)}: The number of simulated signal events correctly identified (i.e., retained by the cuts).
  \item \textbf{False Positive (FP)}: The number of simulated background events incorrectly identified as signal (i.e., passing the cuts).
  \item \textbf{Accuracy}: The fraction of events (both signal and background) that are correctly classified. It's calculated as (TP + TN) / total events, where TN is true negatives (the number of background events correctly identified).
  \item \textbf{Precision}: The ratio of correctly identified signal events to all events predicted as signal, calculated as TP / (TP + FP).
  \item \textbf{Significance}: A measure of sensitivity to a rare signal against a dominant background in high-energy physics. It's computed as $S / \sqrt{B}$, where $S=\epsilon_s \times \sigma_s \times \mathcal{L}$ is the expected number of signal events and $B=\epsilon_b \times \sigma_b \times \mathcal{L}$ is the expected number of background events after all learned cuts. Here, $\epsilon_s =$TP / total test signal events is the signal efficiency, $\epsilon_b =$FP / total test background events is the background efficiency. The integrated luminosity $\mathcal{L}$ is assumed to be 3000~$\text{fb}^{-1}$. For the mock datasets, we assume: $\sigma_s=1$ pb (rare signal) and $\sigma_b=10^6$ pb (abundant background). For the real diboson dataset, we estimate the cross sections based on the original processes: approximately 0.7644 pb for the diboson signal and $1.806 \times 10^5$ pb for the QCD background. Since ref.~\cite{1603.09349} does not provide the cross sections after their full processing pipeline, we derived them independently using only anti-k$_t$ jet clustering ($R=1.2$). These nominal cross sections are used for basic significance comparisons and differ from those in the reference due to the omitted trimming and selection steps.
\end{itemize}

\section{Results}
\label{section:results}

\subsection{Learned cuts}
\label{subsection:learned_cuts}
\begin{figure}[htbp]
  \centering
  \includegraphics[width=.45\textwidth]{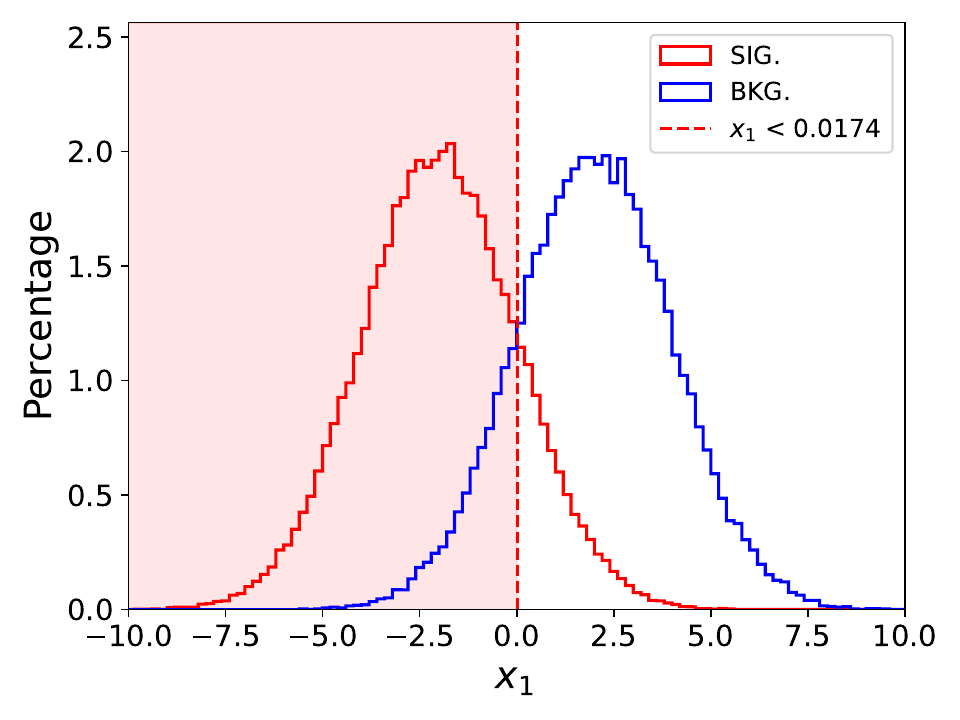} \qquad
  \includegraphics[width=.45\textwidth]{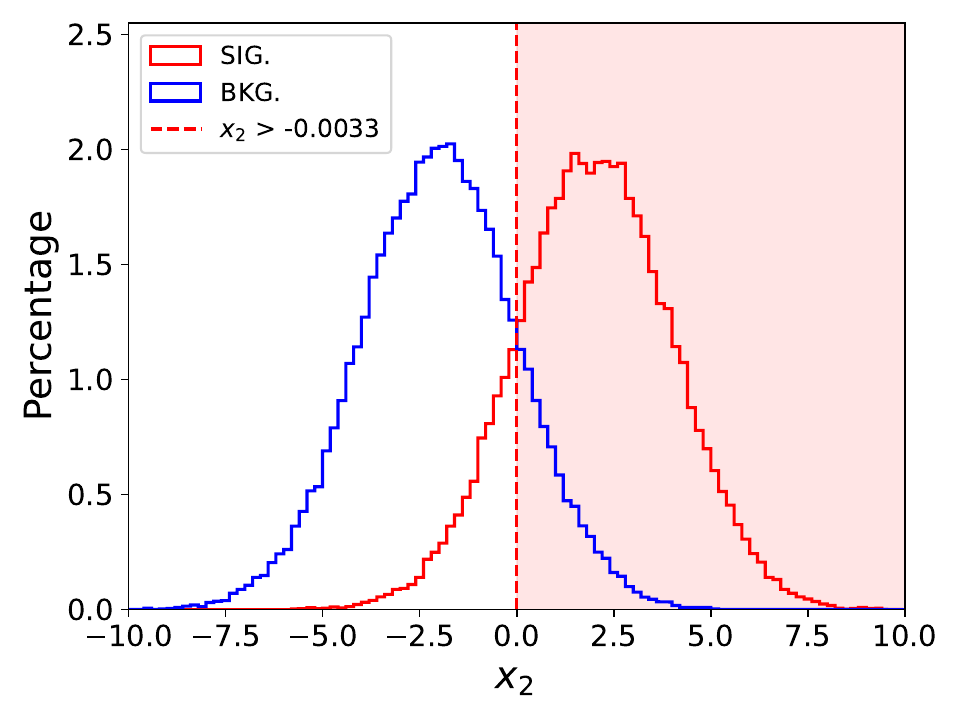} \qquad
  \includegraphics[width=.45\textwidth]{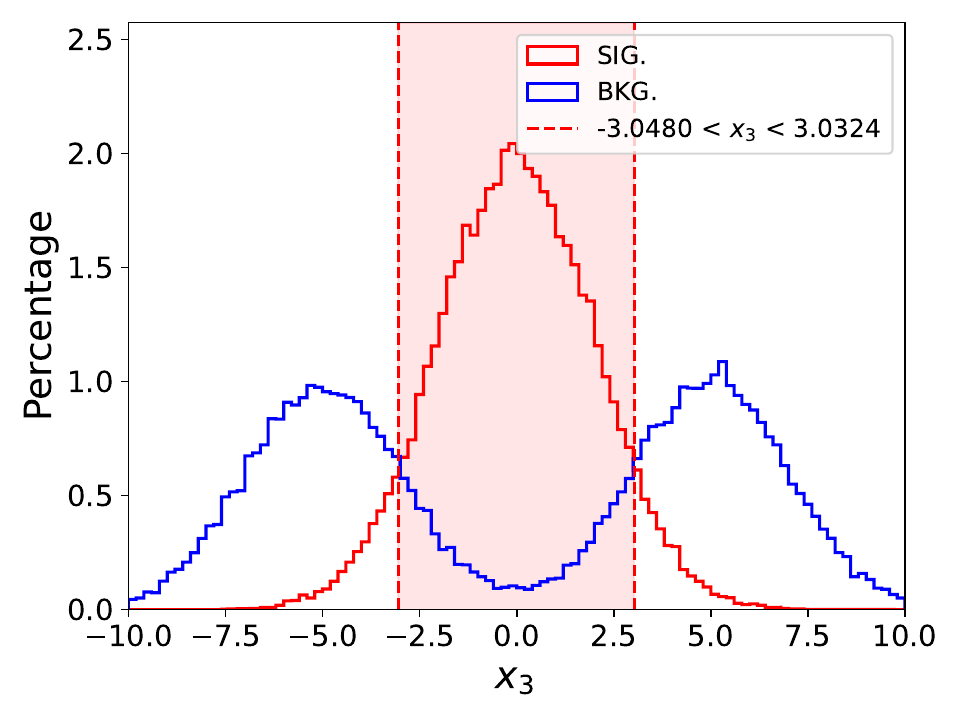} \qquad
  \includegraphics[width=.45\textwidth]{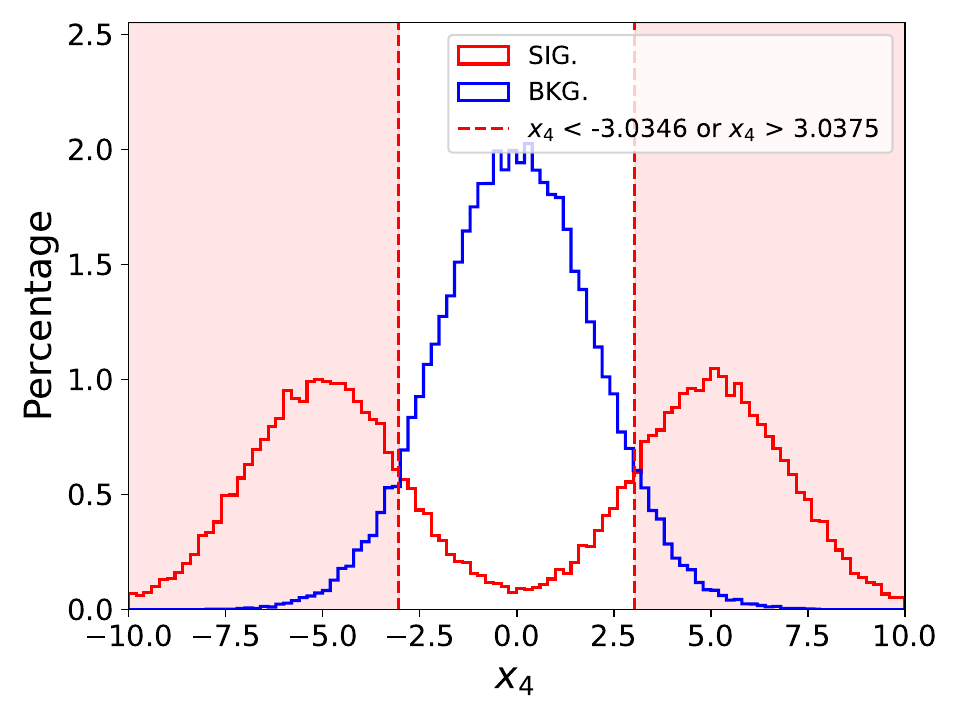}
  \caption{Learned cuts from the parallel LCF on the Mock1 dataset. The error bars on the learned cut boundaries are negligible compared to the bin width and are therefore not shown.}
  \label{figure:learned_cuts-mock1-lcf_par}
\end{figure}

\begin{figure}[htbp]
  \centering
  \includegraphics[width=.45\textwidth]{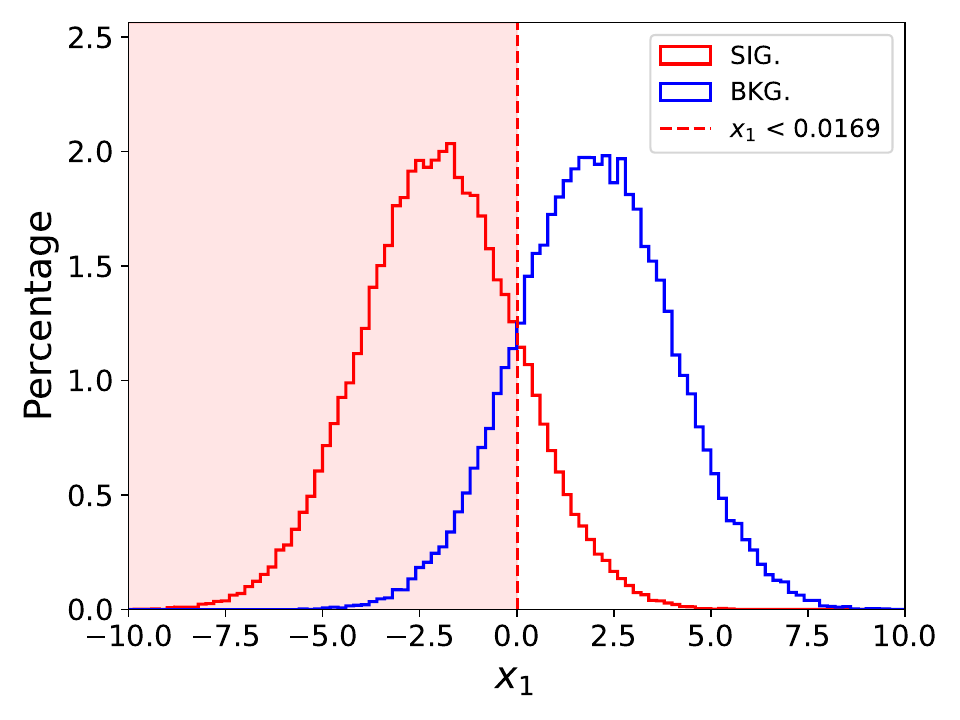} \qquad
  \includegraphics[width=.45\textwidth]{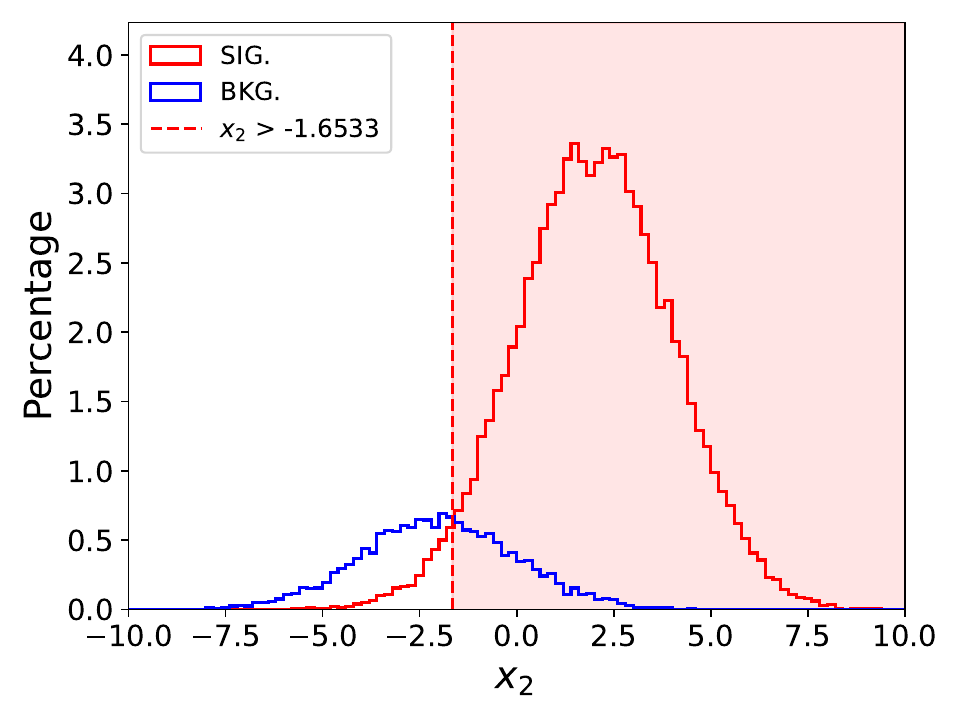} \qquad
  \includegraphics[width=.45\textwidth]{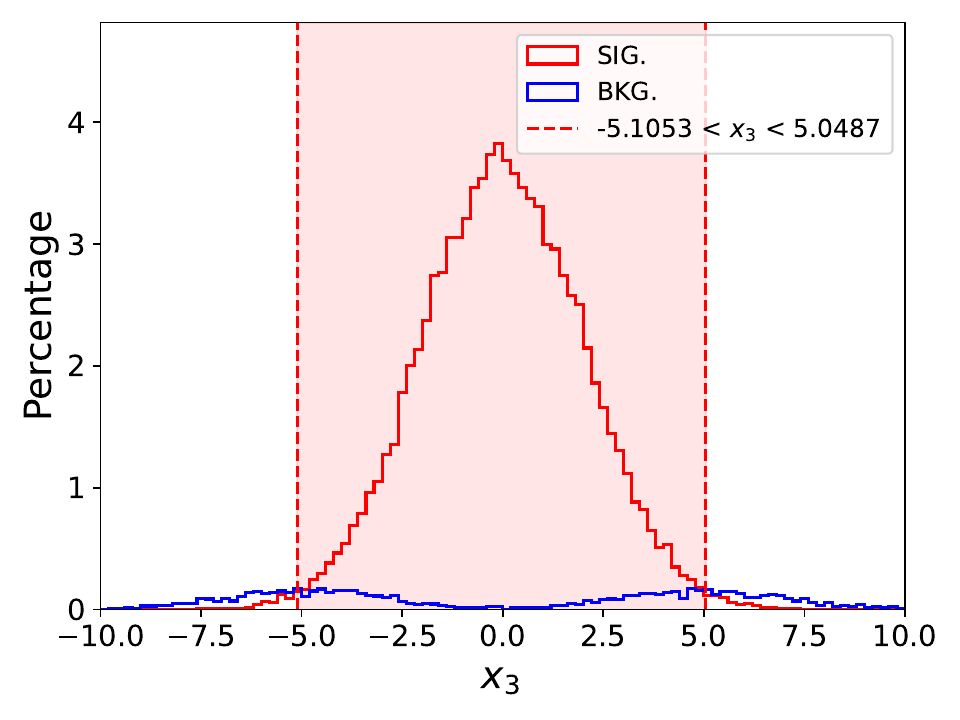} \qquad
  \includegraphics[width=.45\textwidth]{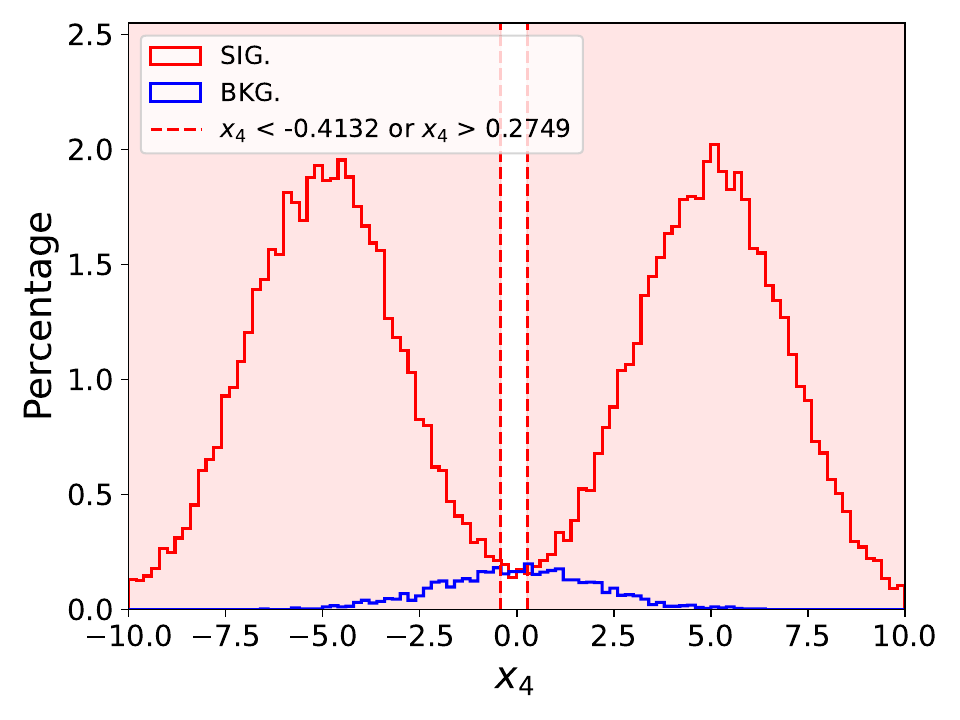}
  \caption{Learned cuts from the sequential LCF on the Mock1 dataset. The error bars on the learned cut boundaries are negligible compared to the bin width and are therefore not shown.}
  \label{figure:learned_cuts-mock1-lcf_seq}
\end{figure}

To validate the ability of LCF models to learn optimal cut positions, we evaluate their performance on the Mock1 dataset, which includes four features illustrating fundamental cut scenarios: left, right, middle, and edge cases. The learned cuts are shown alongside signal and background distributions in figure~\ref{figure:learned_cuts-mock1-lcf_par} (parallel LCF) and figure~\ref{figure:learned_cuts-mock1-lcf_seq} (sequential LCF).

In the parallel strategy, the model analyzes each feature's distribution independently to optimize its cut positions. As shown in figure~\ref{figure:learned_cuts-mock1-lcf_par}, the resulting cuts effectively isolate signal-rich regions: $x_1$ captures the left tail of the signal, $x_2$ the right tail, $x_3$ the central peak, and $x_4$ the outer tails. These positions align well with the boundaries designed for Mock1. In contrast, the sequential strategy applies cuts iteratively, modifying the distribution after each cut to account for inter-feature correlations. Figure~\ref{figure:learned_cuts-mock1-lcf_seq} illustrates how the cut positions adapt to these updates, with subsequent cuts refining the remaining signal region to improve signal-background separation as the feature space is progressively filtered by earlier cuts.

\begin{figure}[htbp]
  \centering
  \includegraphics[width=.45\textwidth]{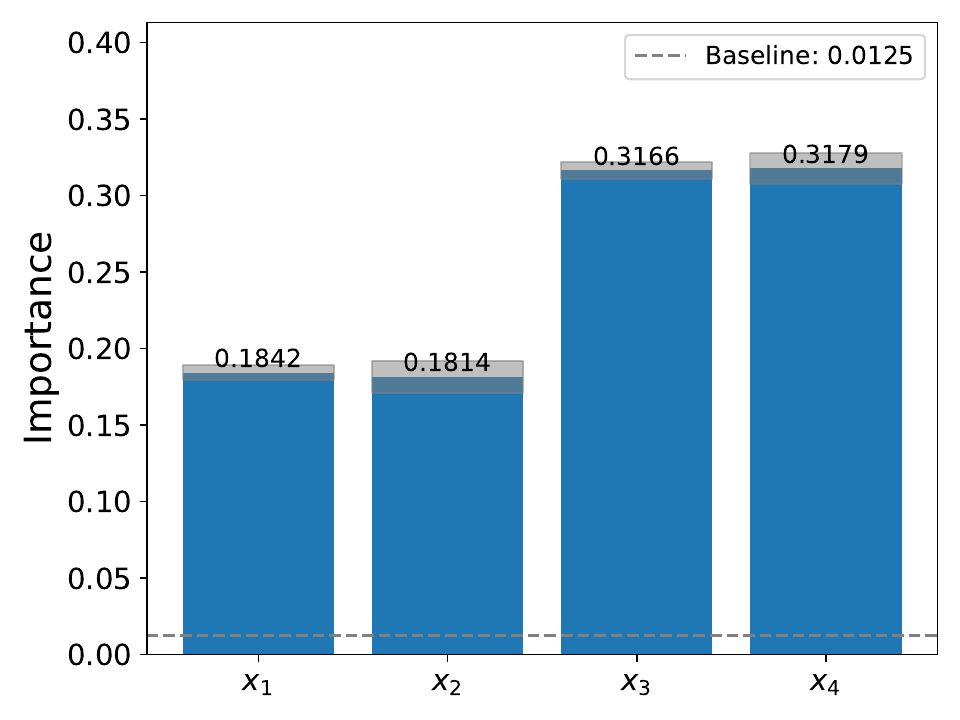} \qquad
  \includegraphics[width=.45\textwidth]{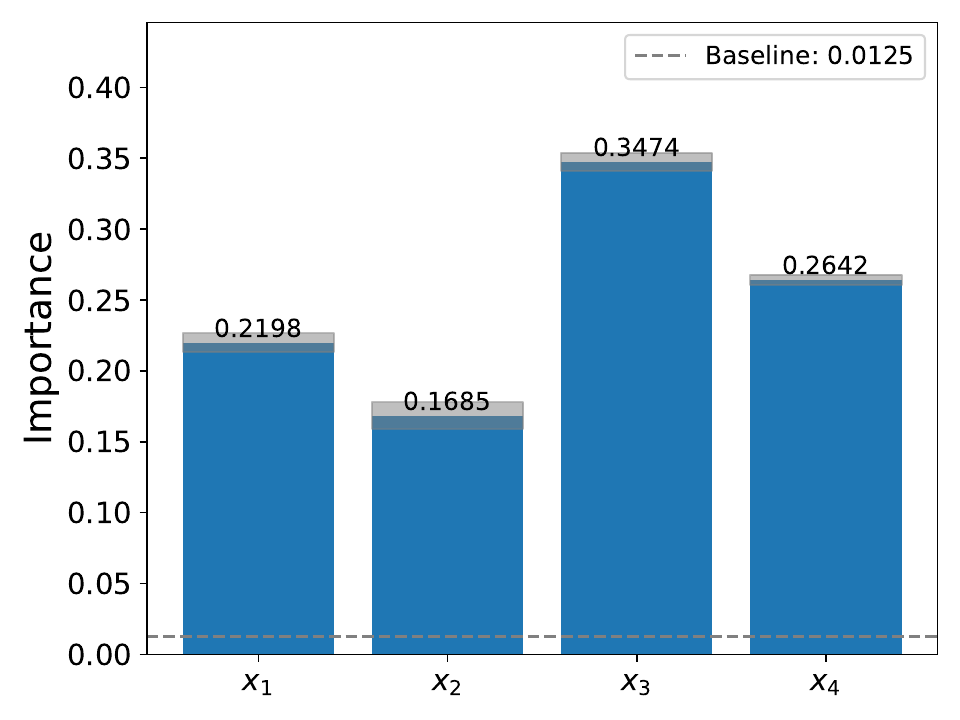}
  \caption{Learned importance from parallel (left) and sequential (right) LCFs on the Mock1 dataset. The baseline indicates the minimum importance (default: 5\% of average importance = $1/F \times 0.05$), below which the features are ignored during inference.}
  \label{figure:learned_importance-mock1}
\end{figure}

The importance scores shown in figure~\ref{figure:learned_importance-mock1} reflect how the LCF prioritizes each feature. The baseline shown in the same figure represents the minimum threshold (5\% of average importance) below which features are automatically excluded during inference. In the parallel strategy, scores are relatively balanced: features with single-sided cuts ($x_1$ and $x_2$) have scores 0.1842 and 0.1814, respectively, compared to features with double-sided cuts ($x_3$ and $x_4$) at 0.3166 and 0.3179. In contrast, the importance shifts in the sequential strategy as prior cuts filter out a fraction of background events: $x_1$ increases to 0.2198 while $x_2$ decreases to 0.1685; similarly, $x_3$ increases to 0.3474 and $x_4$ decreases to 0.2642.

\begin{table}[htbp]
  \centering
  \begin{tabular}{|l|c|c|c|c|c|}
    \hline
    \textbf{Model} & \textbf{TP} & \textbf{FP} & \textbf{Accuracy} & \textbf{Precision} & \textbf{Significance} \\
    \hline
    LCF (Parallel) & $25734 \pm 66$ & $34 \pm 0$ & $75.8 \pm 0.1\%$ & $99.9 \pm 0.0\%$ & $34.3 \pm 0.1$ \\
    LCF (Sequential) & $39759 \pm 23$ & $1577 \pm 19$ & $88.3 \pm 0.0\%$ & $96.2 \pm 0.0\%$ & $7.8 \pm 0.0$ \\
    \hline
  \end{tabular}
  \caption{Performance metrics of LCF models on the Mock1 dataset.}
  \label{table:metrics_mock1}
\end{table}

In table~\ref{table:metrics_mock1}, we observe that the sequential strategy captures more true signal events and achieves higher accuracy because it examines the distributions more carefully by considering the relationships among features. However, this strategy also retains more background events, resulting in lower precision and significance. The parallel LCF significantly reduces the number of background events at the cost of modest signal loss, achieving higher precision and remarkably high significance.

The learned cuts and importance scores demonstrate that LCF models can replicate the traditional cut-based methods while also dynamically adapting to the evolving feature space during training.

\subsection{Learned importance}
\label{subsection:learned_importance}

\begin{figure}[htbp]
  \centering
  \includegraphics[width=.32\textwidth]{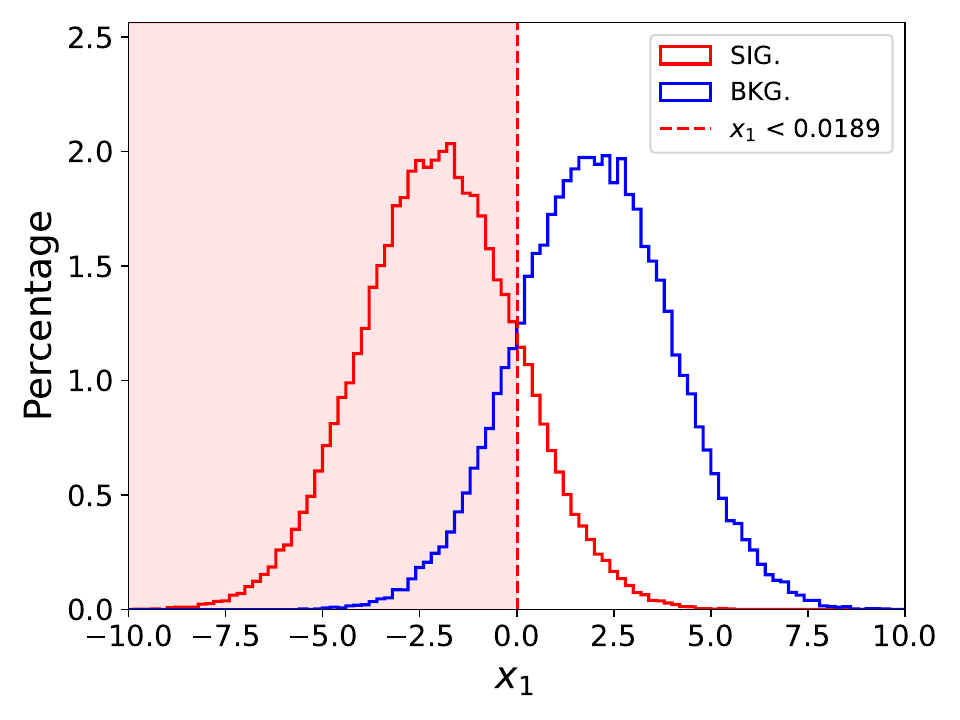}
  \includegraphics[width=.32\textwidth]{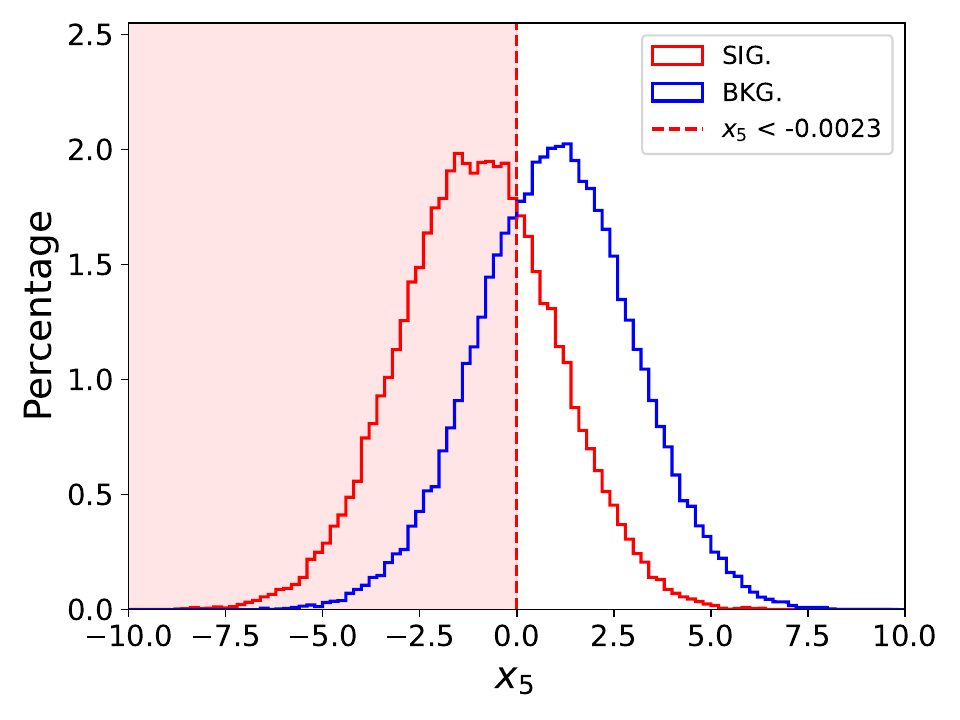}
  \includegraphics[width=.32\textwidth]{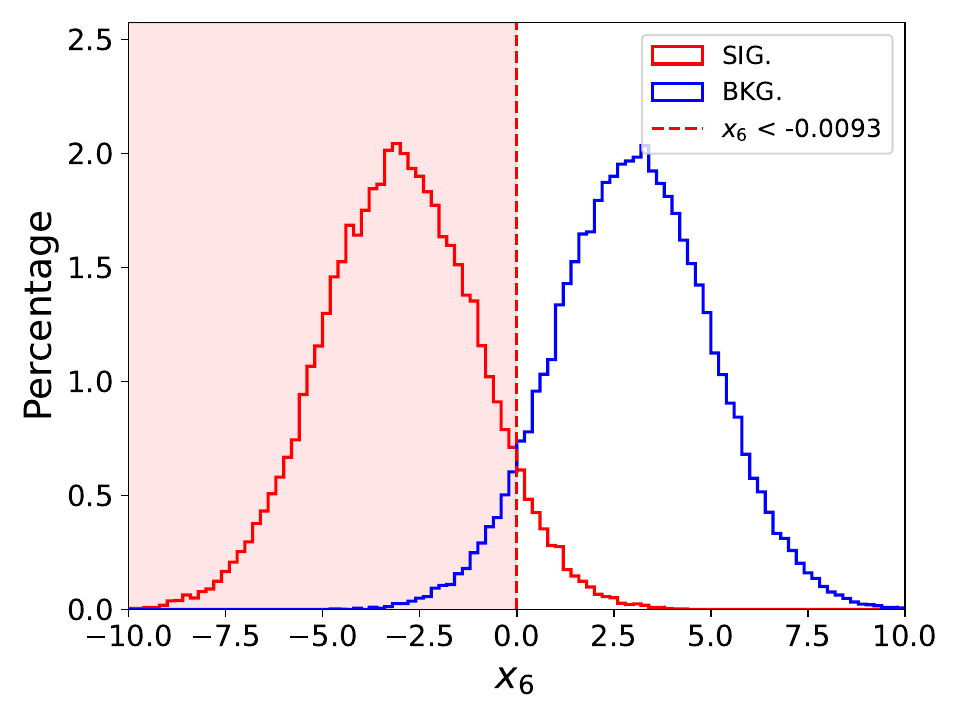}
  \caption{Learned cuts from the parallel LCF on the Mock2 dataset. The error bars on the learned cut boundaries are negligible compared to the bin width and are therefore not shown.}
  \label{figure:learned_cuts-mock2-lcf_par}
\end{figure}

\begin{figure}[htbp]
  \centering
  \includegraphics[width=.32\textwidth]{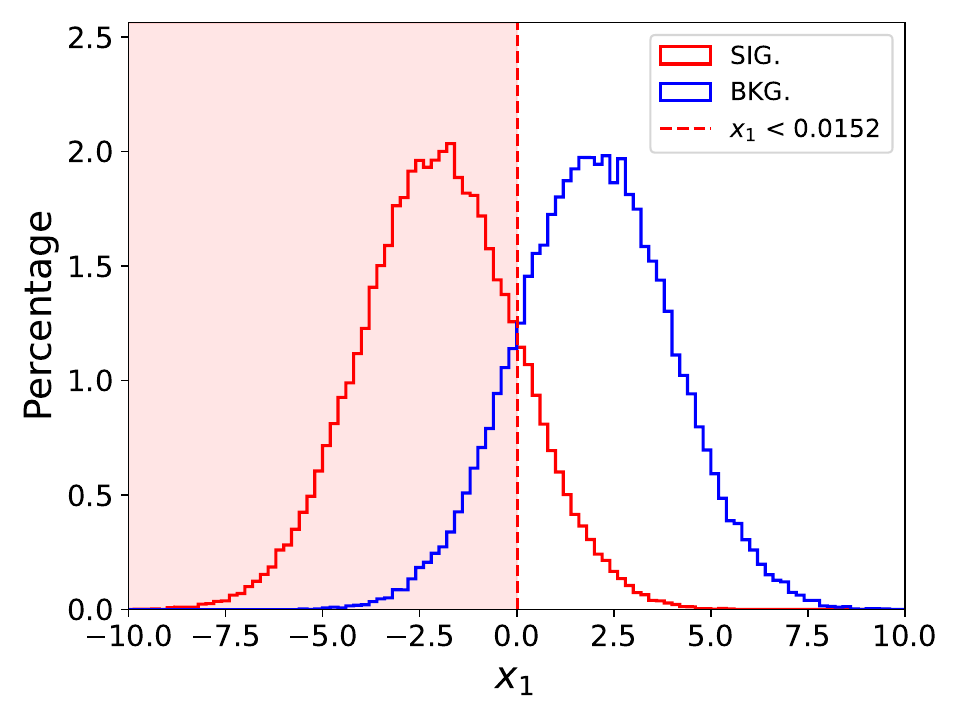}
  \includegraphics[width=.32\textwidth]{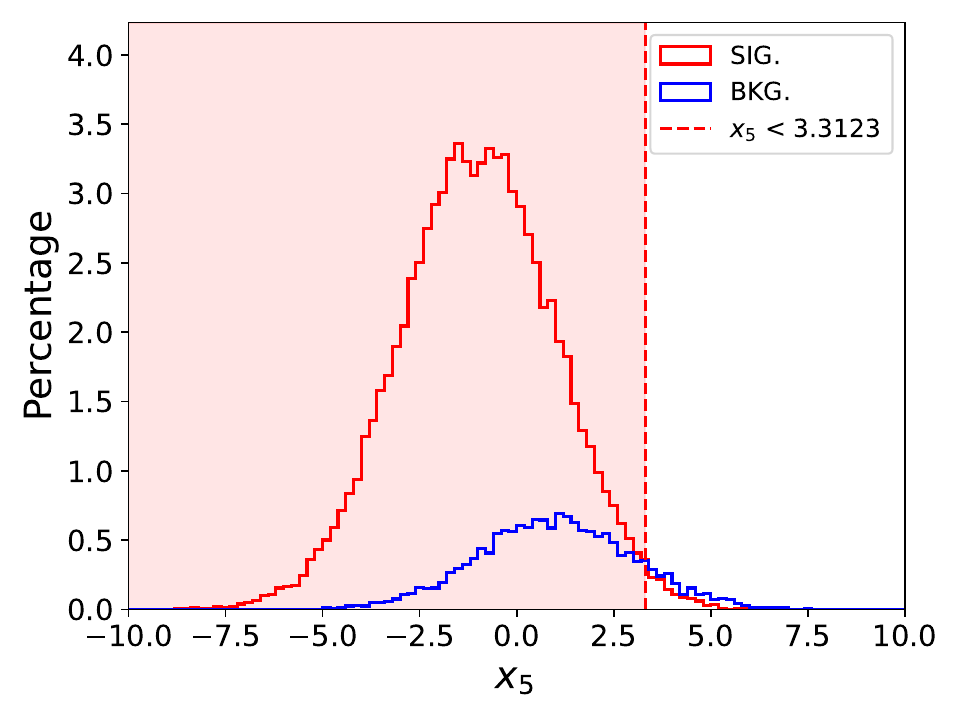}
  \includegraphics[width=.32\textwidth]{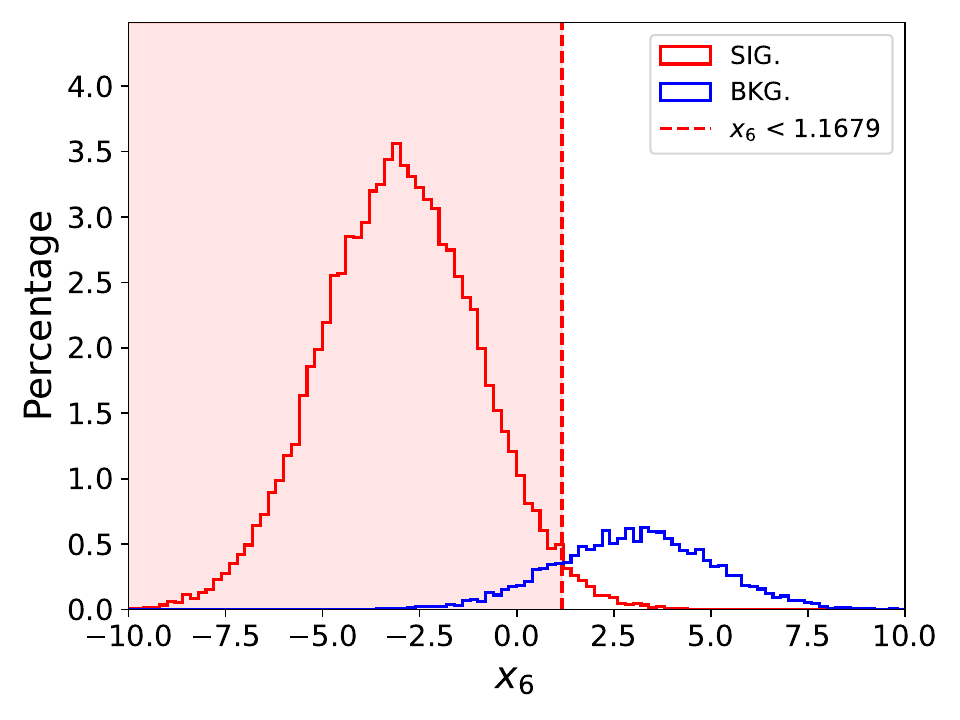}
  \caption{Learned cuts from the sequential LCF on the Mock2 dataset. The error bars on the learned cut boundaries are negligible compared to the bin width and are therefore not shown.}
  \label{figure:learned_cuts-mock2-lcf_seq}
\end{figure}

\begin{figure}[htbp]
  \centering
  \includegraphics[width=.45\textwidth]{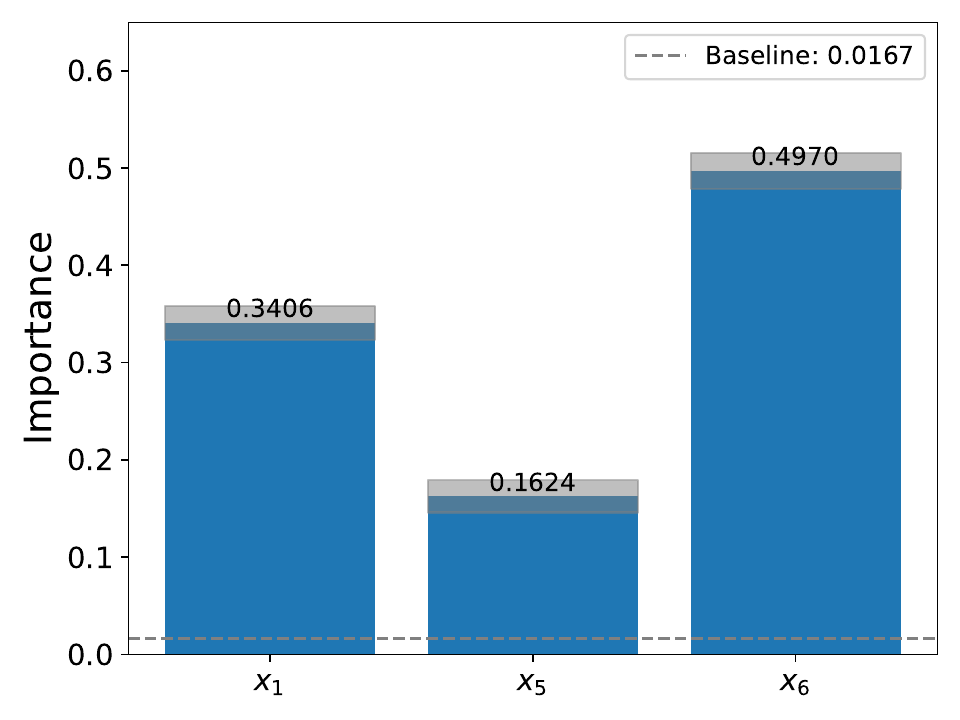} \qquad
  \includegraphics[width=.45\textwidth]{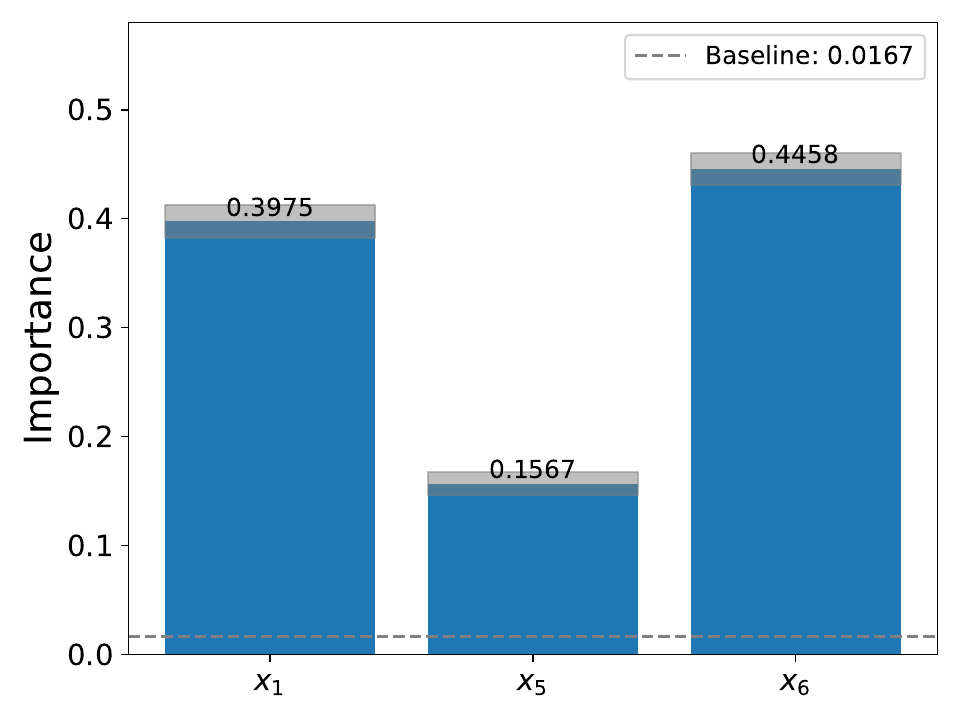}
  \caption{Learned importance from parallel (left) and sequential (right) LCFs on the Mock2 dataset. The baseline indicates the minimum importance (default: 5\% of average importance = $1/F \times 0.05$), below which the features are ignored during inference.}
  \label{figure:learned_importance-mock2}
\end{figure}

To further evaluate the LCF model's ability to prioritize features, we analyze the importance scores learned on the Mock2 dataset, which includes three features ($x_1$, $x_5$, $x_6$) with regular, weak, and strong signal-background separation. The signal and background distributions, along with the learned cuts, are shown in figure~\ref{figure:learned_cuts-mock2-lcf_par} (parallel LCF) and figure~\ref{figure:learned_cuts-mock2-lcf_seq} (sequential LCF). The corresponding importance scores are shown in figure~\ref{figure:learned_importance-mock2}.

In the parallel strategy, the LCF model assigns importance scores that closely match each feature's intended role: compared to $x_1$'s regular signal-background separation, $x_5$ decreases it while $x_6$ increases it. The sequential strategy doesn't change much the rank of the features. The score of the earlier cut on $x_1$ is raised a lot and later cuts' scores are lowered a little bit, which matches what we find in section~\ref{subsection:learned_cuts}.

\begin{table}[htbp]
  \centering
  \begin{tabular}{|l|c|c|c|c|c|}
    \hline
    \textbf{Model} & \textbf{TP} & \textbf{FP} & \textbf{Accuracy} & \textbf{Precision} & \textbf{Significance} \\
    \hline
    LCF (Parallel) & $27117 \pm 58$ & $155 \pm 1$ & $77.1 \pm 0.1\%$ & $99.4 \pm 0.0\%$ & $17.0 \pm 0.0$ \\
    LCF (Sequential) & $40553 \pm 24$ & $1275 \pm 7$ & $89.4 \pm 0.0\%$ & $97.0 \pm 0.0\%$ & $8.8 \pm 0.0$ \\
    \hline
  \end{tabular}
  \caption{Performance metrics of LCF models on the Mock2 dataset.}
  \label{table:metrics_mock2}
\end{table}

Table~\ref{table:metrics_mock2} shows the same pattern observed in section~\ref{subsection:learned_cuts}. These results demonstrate that the learnable importance mechanism captures each features' discriminative power and adapts dynamically under both the parallel and sequential strategies.

\subsection{Robustness to redundancy}
\label{subsection:robustness_to_redundancy}

\begin{figure}[htbp]
  \centering
  \includegraphics[width=.32\textwidth]{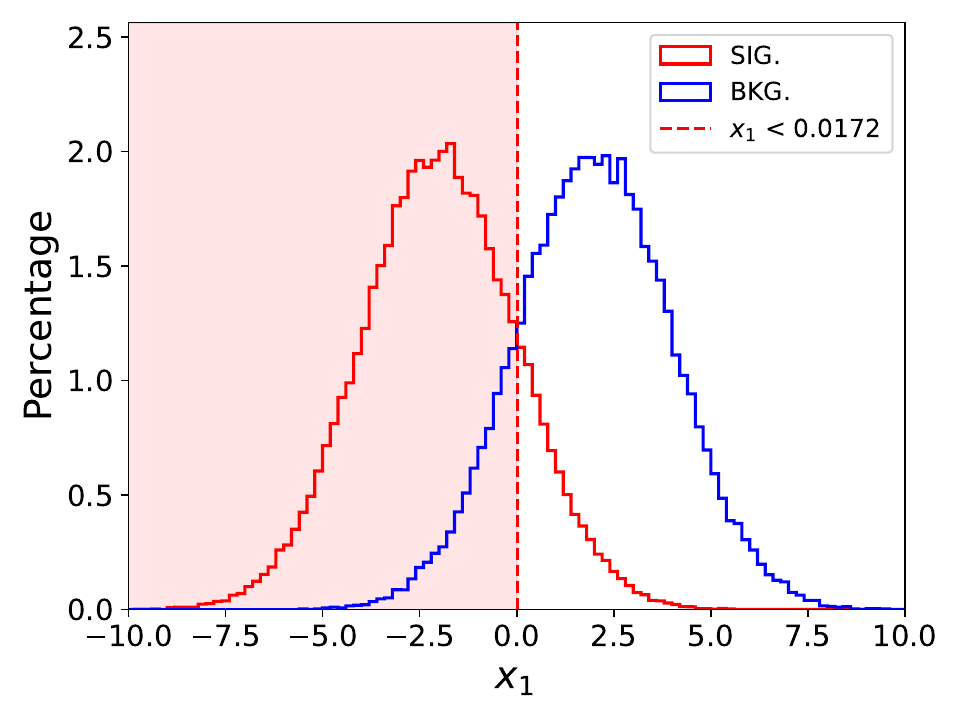}
  \includegraphics[width=.32\textwidth]{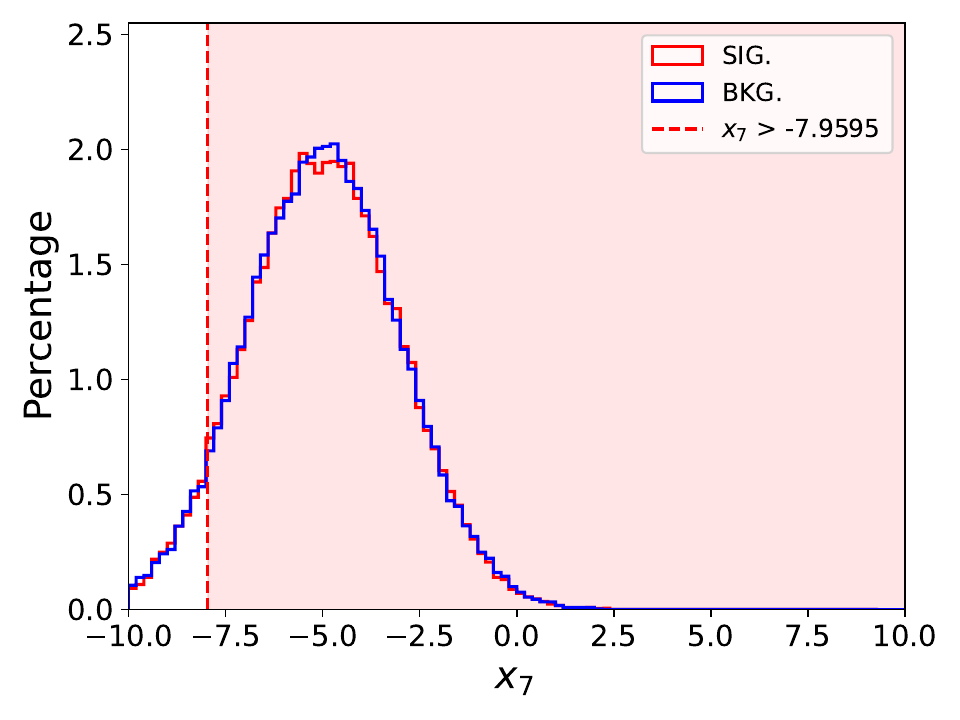}
  \includegraphics[width=.32\textwidth]{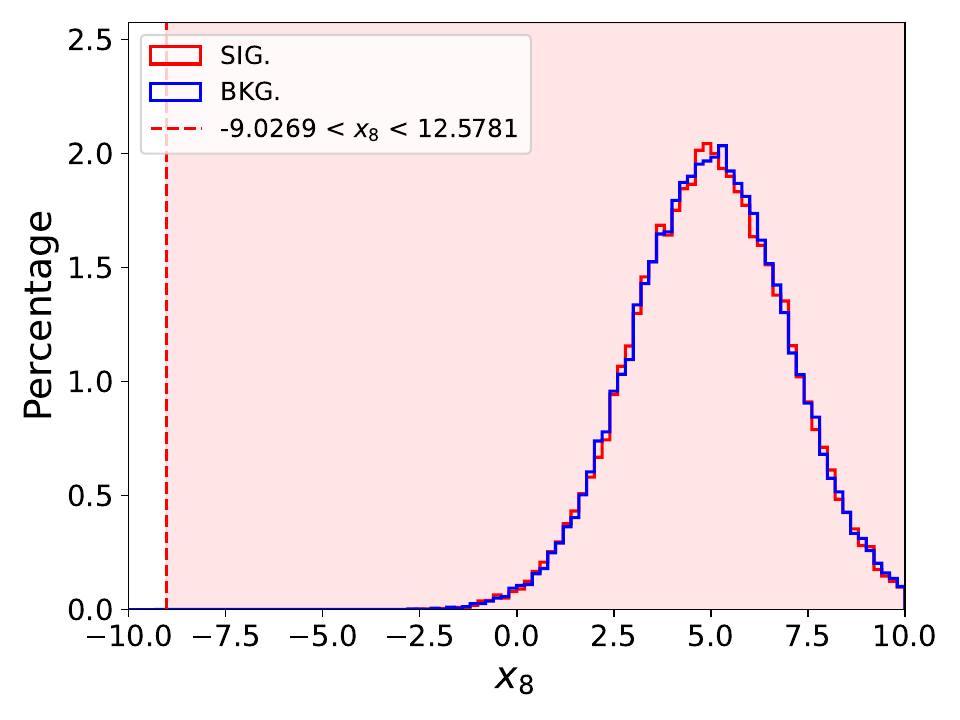}
  \caption{Learned cuts from the parallel LCF on the Mock3 dataset. The error bars are not shown because $x_1$'s error is negligible compared to the bin width while $x_7$'s and $x_8$'s errors have no meaning when the importance is close to zero.}
  \label{figure:learned_cuts-mock3-lcf_par}
\end{figure}

\begin{figure}[htbp]
  \centering
  \includegraphics[width=.32\textwidth]{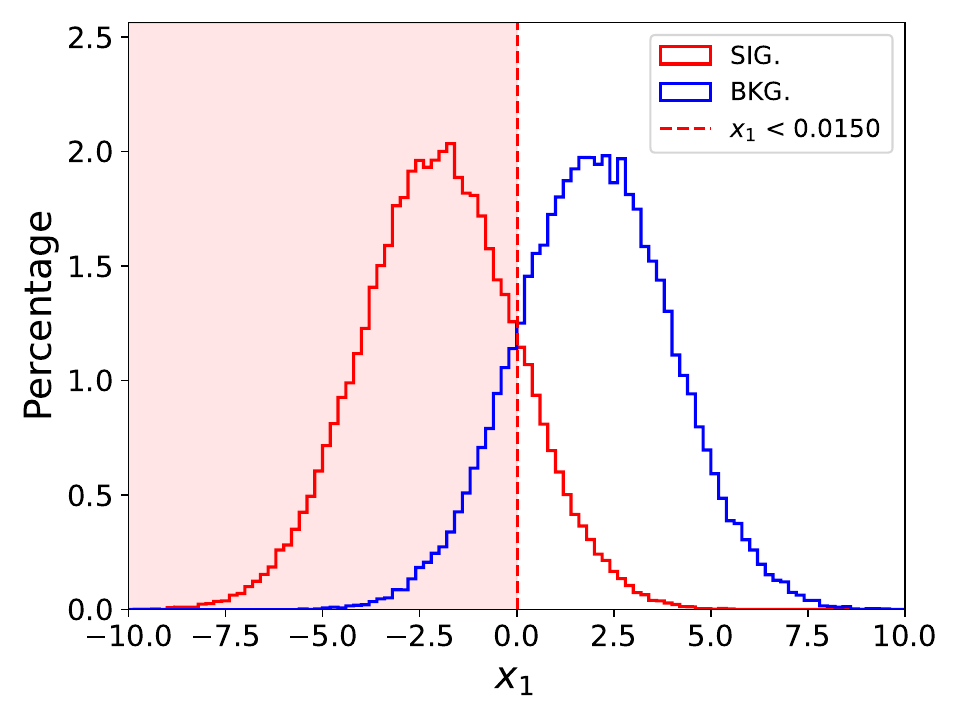}
  \includegraphics[width=.32\textwidth]{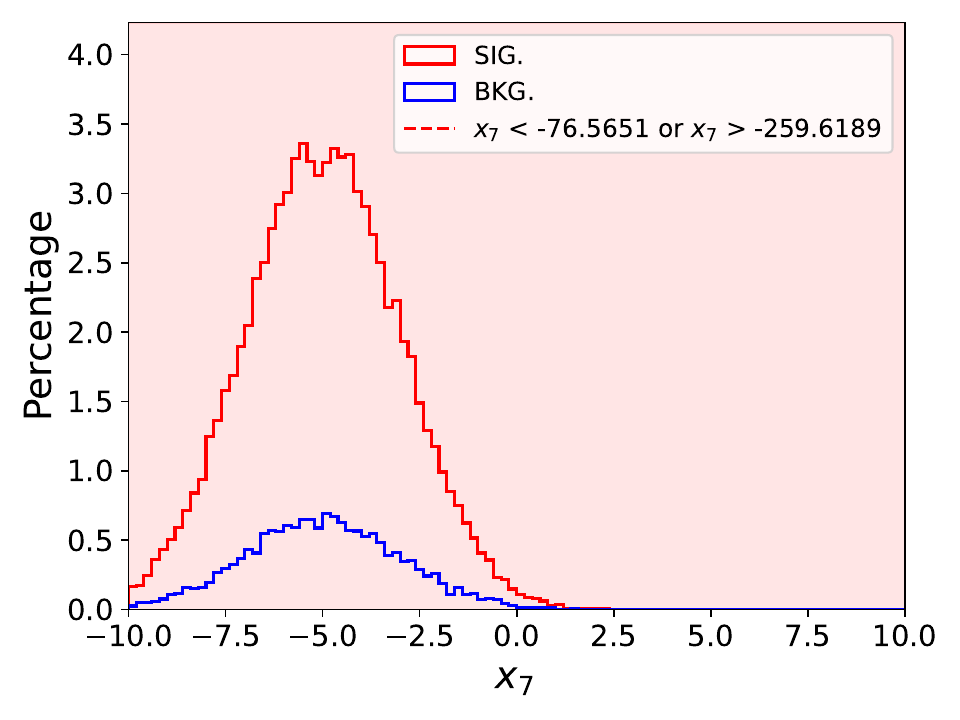}
  \includegraphics[width=.32\textwidth]{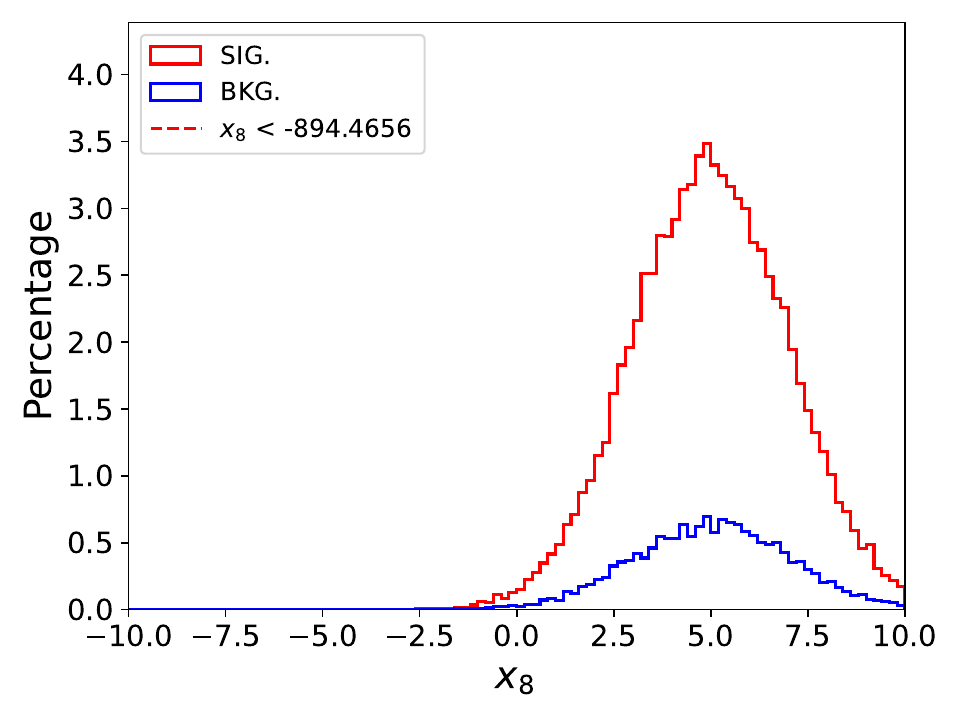}
  \caption{Learned cuts from the sequential LCF on the Mock3 dataset. The error bars are not shown because $x_1$'s error is negligible compared to the bin width while $x_7$'s and $x_8$'s errors have no meaning when the importance is close to zero.}
  \label{figure:learned_cuts-mock3-lcf_seq}
\end{figure}

\begin{figure}[htbp]
  \centering
  \includegraphics[width=.45\textwidth]{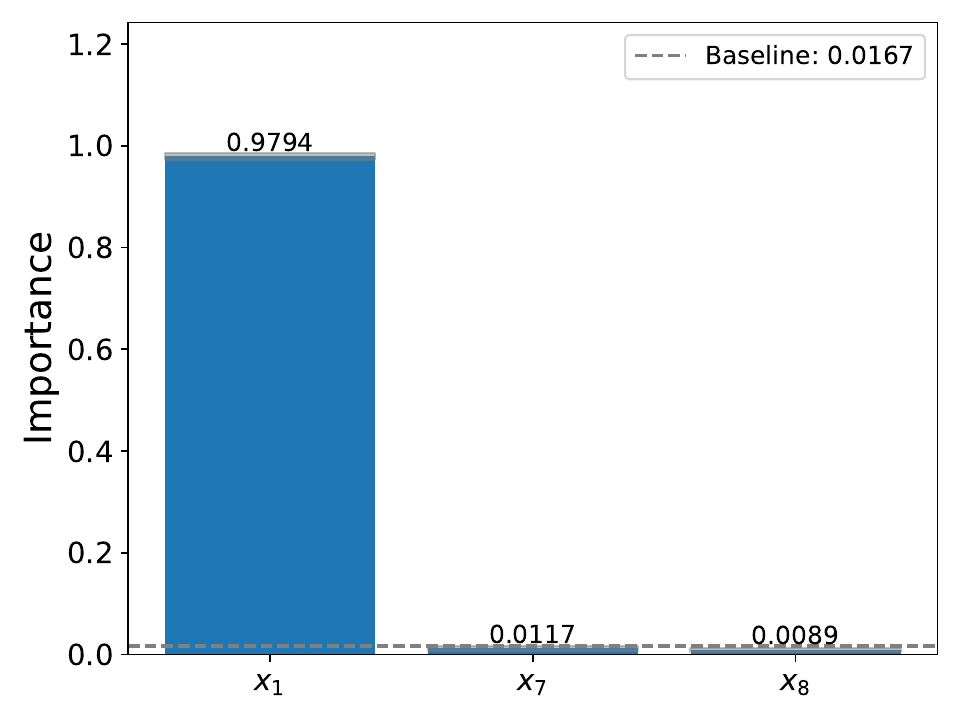} \qquad
  \includegraphics[width=.45\textwidth]{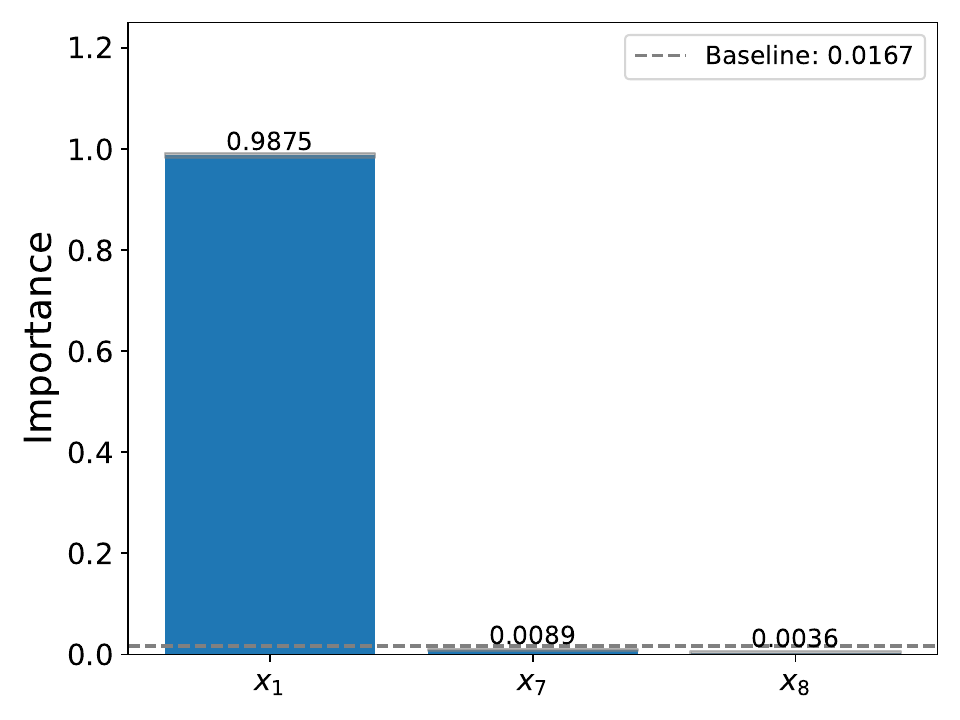}
  \caption{Learned importance from parallel (left) and sequential (right) LCFs on the Mock3 dataset. The baseline indicates the minimum importance (default: 5\% of average importance = $1/F \times 0.05$), below which the features are ignored during inference.}
  \label{figure:learned_importance-mock3}
\end{figure}

To assess how robust the LCF models are when presented with redundant features, we train them on the Mock3 dataset. In this setup, $x_1$ provides a clear signal-background separation, whereas $x_7$ and $x_8$ offer little to no additional discriminative power. As shown in figure~\ref{figure:learned_cuts-mock3-lcf_par} and figure~\ref{figure:learned_cuts-mock3-lcf_seq}, the learned cuts on $x_7$ and $x_8$ appear almost randomly placed, reflecting their limited ability to distinguish signal and background. The corresponding importance scores in figure~\ref{figure:learned_importance-mock3} confirm that both parallel and sequential strategies identify these two features contribute minimally to separation.

\begin{table}[htbp]
  \centering
  \begin{tabular}{|l|c|c|c|c|c|}
    \hline
    \textbf{Model} & \textbf{TP} & \textbf{FP} & \textbf{Accuracy} & \textbf{Precision} & \textbf{Significance} \\
    \hline
    LCF (Parallel) & $38673 \pm 8392$ & $7443 \pm 1624$ & $81.4 \pm 6.8\%$ & $83.9 \pm 0.1\%$ & $3.5 \pm 0.5$ \\
    LCF (Sequential) & $41924 \pm 21$ & $8060 \pm 21$ & $84.0 \pm 0.0\%$ & $83.9 \pm 0.0\%$ & $3.6 \pm 0.0$ \\
    \hline
  \end{tabular}
  \caption{Performance metrics of LCF models on the Mock3 dataset.}
  \label{table:metrics_mock3}
\end{table}

Table~\ref{table:metrics_mock3} shows that both strategies achieve almost the same performance. This finding highlights the LCF models' robustness in detecting and effectively ignoring features with negligible separation power.

\subsection{Robustness to high correlation}
\label{subsection:robustness_to_high_correlation}

\begin{figure}[htbp]
  \centering
  \includegraphics[width=.32\textwidth]{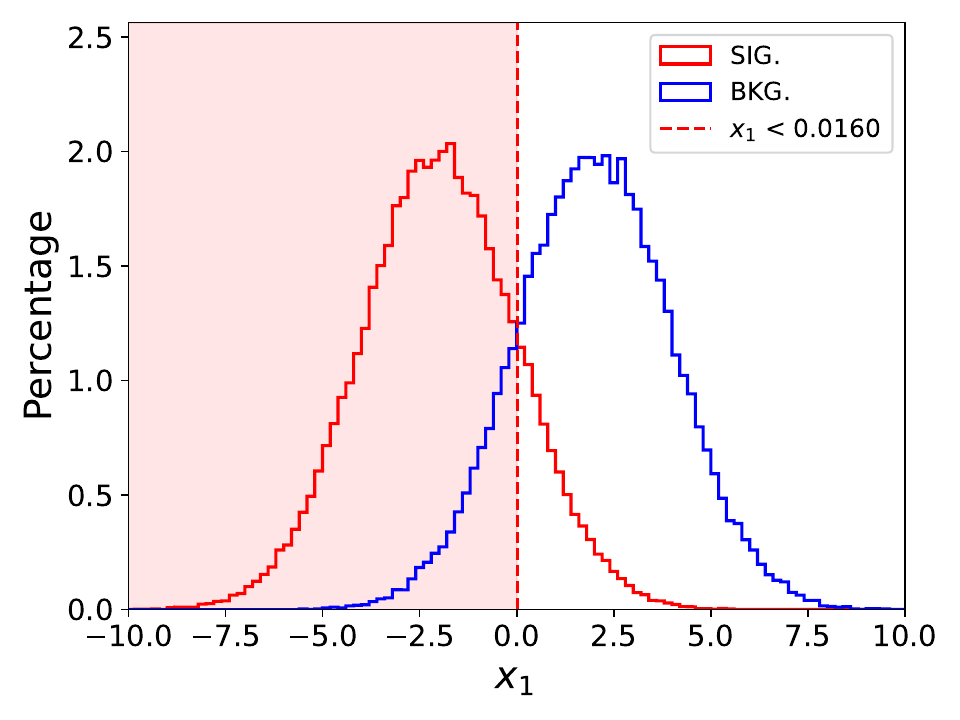}
  \includegraphics[width=.32\textwidth]{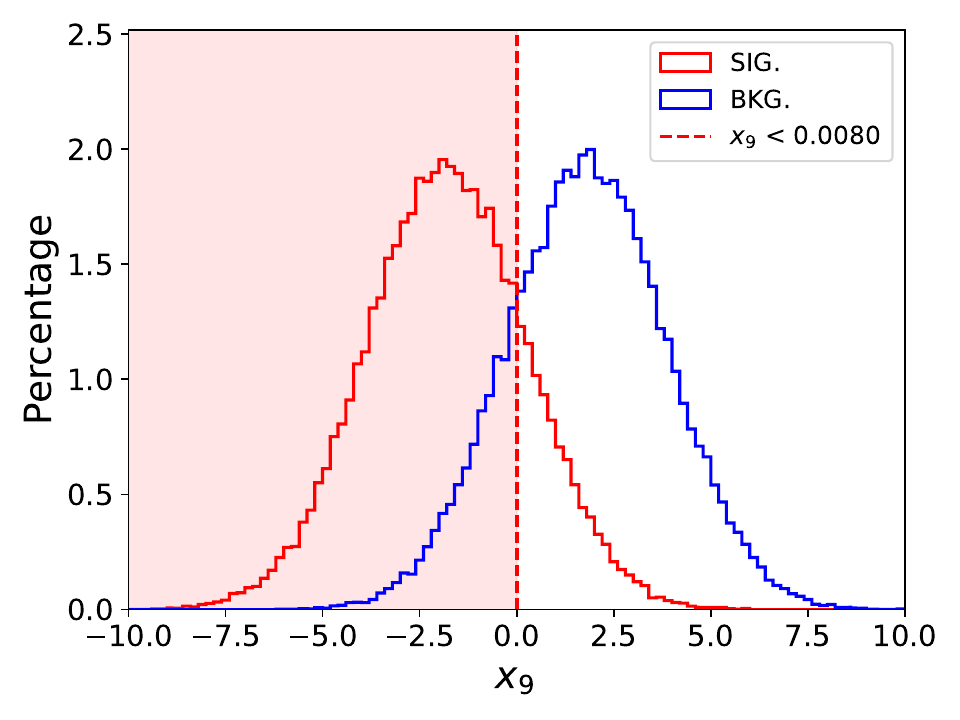}
  \includegraphics[width=.32\textwidth]{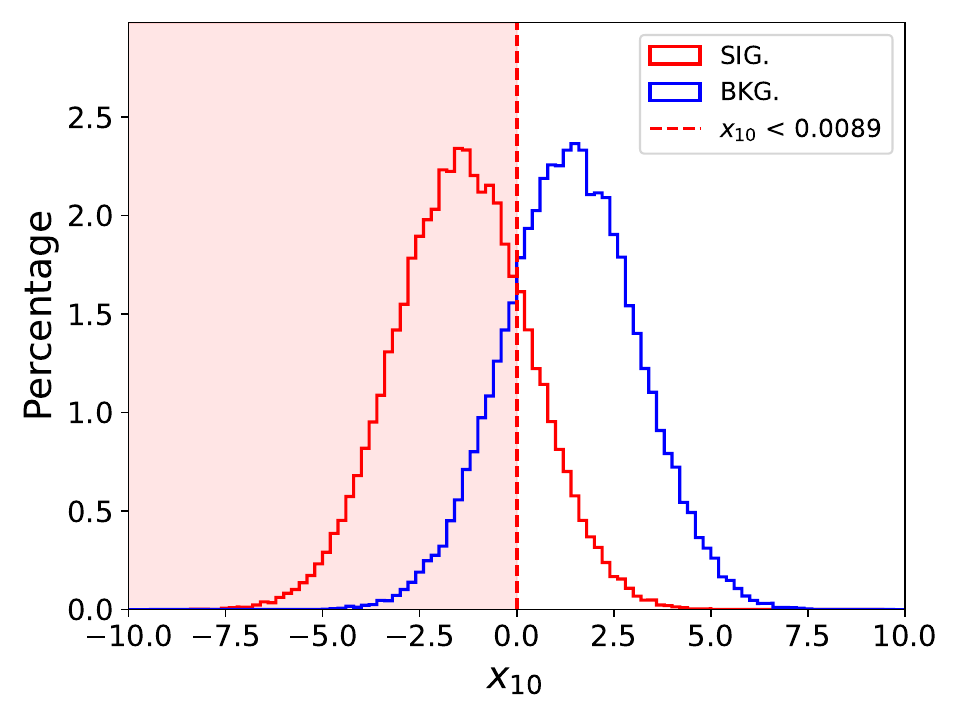}
  \caption{Learned cuts from the parallel LCF on the Mock4 dataset. The error bars on the learned cut boundaries are negligible compared to the bin width and are therefore not shown.}
  \label{figure:learned_cuts-mock4-lcf_par}
\end{figure}

\begin{figure}[htbp]
  \centering
  \includegraphics[width=.32\textwidth]{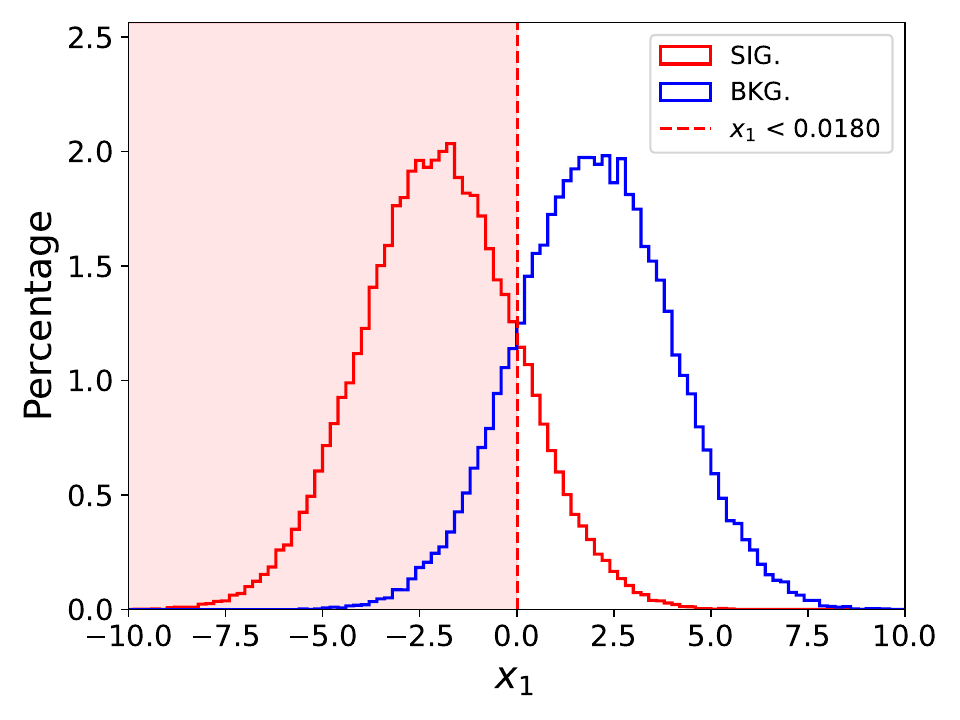}
  \includegraphics[width=.32\textwidth]{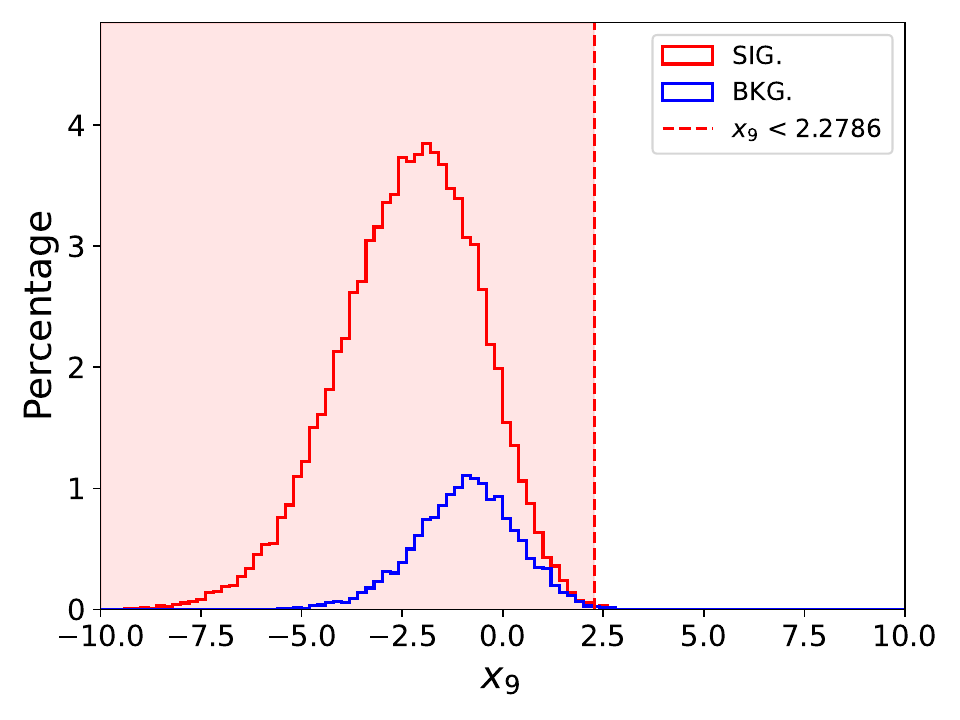}
  \includegraphics[width=.32\textwidth]{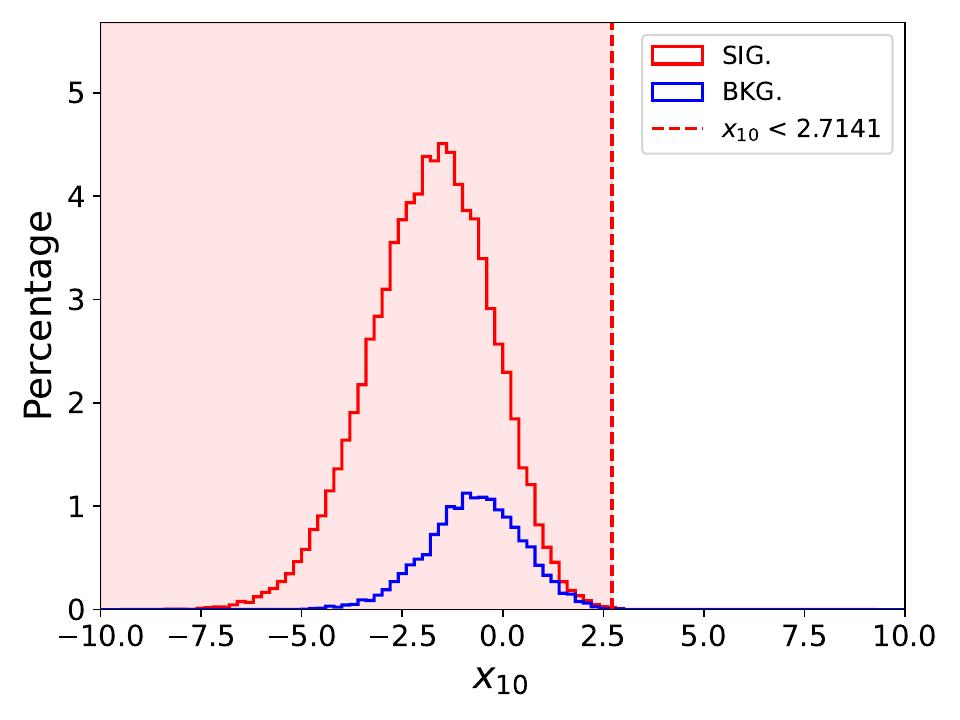}
  \caption{Learned cuts from the sequential LCF on the Mock4 dataset. The error bars on the learned cut boundaries are negligible compared to the bin width and are therefore not shown.}
  \label{figure:learned_cuts-mock4-lcf_seq}
\end{figure}

\begin{figure}[htbp]
  \centering
  \includegraphics[width=.45\textwidth]{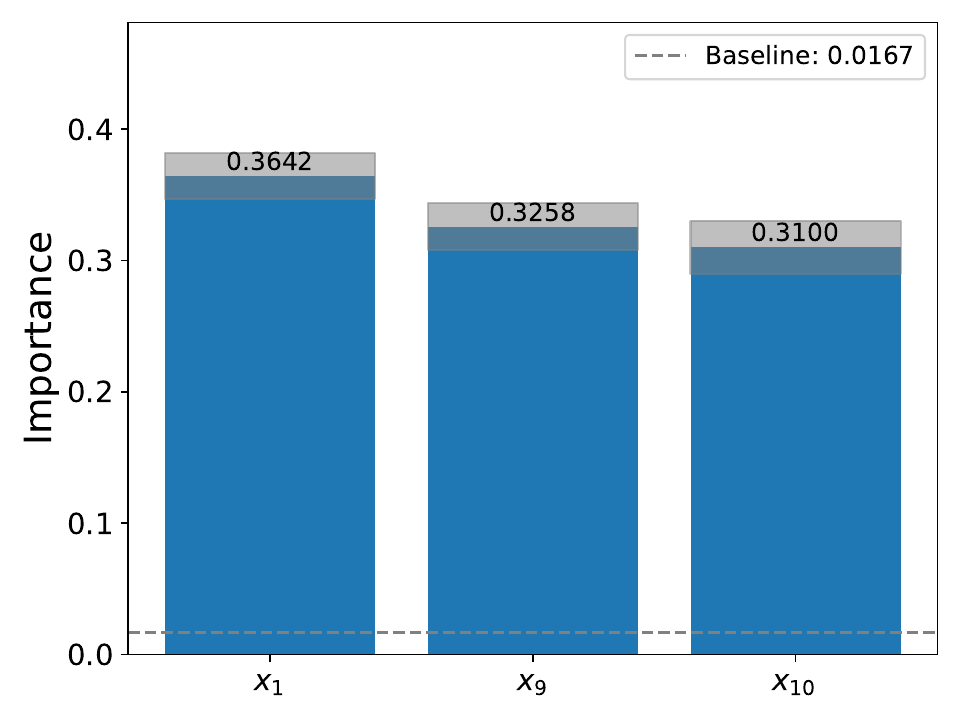} \qquad
  \includegraphics[width=.45\textwidth]{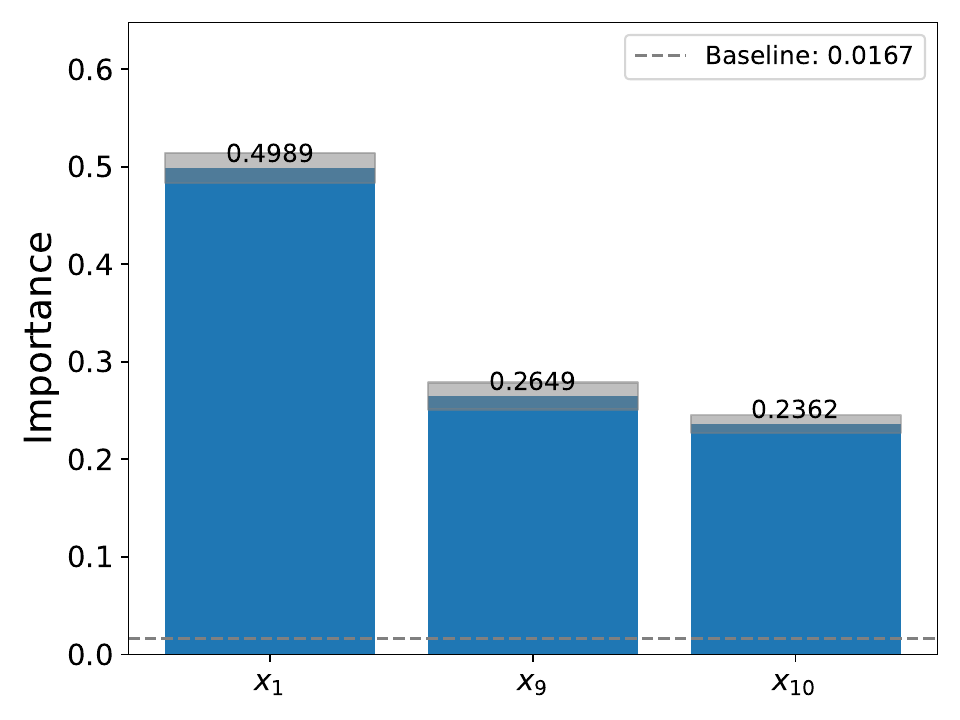}
  \caption{Learned importance from parallel (left) and sequential (right) LCFs on the Mock4 dataset. The baseline indicates the minimum importance (default: 5\% of average importance = $1/F \times 0.05$), below which the features are ignored during inference.}
  \label{figure:learned_importance-mock4}
\end{figure}

High correlation between observables implies that a cut on one feature effectively constrains the others, posing a challenge for cut-based methods. To evaluate the LCF models' robustness in this scenario, we train them on the Mock4 dataset which includes $x_1$ alongside highly correlated features $x_9$ and $x_{10}$. The resulting learned cuts are shown in figure~\ref{figure:learned_cuts-mock4-lcf_par} (parallel LCF) and figure~\ref{figure:learned_cuts-mock4-lcf_seq} (sequential LCF), with corresponding importance scores shown in figure~\ref{figure:learned_importance-mock4}.

In the parallel strategy, the LCF model treats each feature independently, so the high correlation doesn't significantly affect the learned cuts or importance scores. These scores remain consistent with the dataset's design. In contrast, since the first cut in the sequential strategy reshapes the subsequent feature distributions, $x_9$ and $x_{10}$ end up with lower importance. Although the learned cuts of the sequential strategy almost cover the full range of the features, the importance scores are still not zero. It is because the cuts on $x_9$ and $x_{10}$ generate reasonable losses compared to $x_7$ and $x_8$ in the Mock3 dataset.

\begin{table}[htbp]
  \centering
  \begin{tabular}{|l|c|c|c|c|c|}
    \hline
    \textbf{Model} & \textbf{TP} & \textbf{FP} & \textbf{Accuracy} & \textbf{Precision} & \textbf{Significance} \\
    \hline
    LCF (Parallel) & $35010 \pm 19$ & $4710 \pm 9$ & $80.4 \pm 0.0\%$ & $88.1 \pm 0.0\%$ & $4.0 \pm 0.0$ \\
    LCF (Sequential) & $41903 \pm 18$ & $8044 \pm 16$ & $84.0 \pm 0.0\%$ & $83.9 \pm 0.0\%$ & $3.6 \pm 0.0$ \\
    \hline
  \end{tabular}
  \caption{Performance metrics of LCF models on the Mock4 dataset.}
  \label{table:metrics_mock4}
\end{table}

Combined with the results in sections~\ref{subsection:learned_cuts}, \ref{subsection:learned_importance}, and \ref{subsection:robustness_to_redundancy}, we observe that the parallel strategy typically achieves higher precision and significance, while the sequential strategy achieves higher accuracy. This difference most likely arises from a fundamental consequence of the parallel strategy's independent feature processing. For features $x_1$, $x_2$, $x_3$, $x_4$, $x_5$, $x_6$, $x_9$, and $x_{10}$, portions of the background have already been filtered out by other cuts operating in parallel. This causes the cross point of the signal and background distributions to shift outward relative to the signal distribution. Consequently, the cuts learned from the original distributions happen to be more stringent and more conservative than the ones learned from the distributions after applying cuts on previous variables. This makes the parallel strategy achieve better performance on the Mock1, Mock2, and Mock4 datasets where such cuts remove a large portion of the background at a fair cost of signal efficiency. However, in the Mock3 dataset, random cuts learned from $x_7$ and $x_8$ make the reduction of signal and background an almost even trade-off, which has a negative impact on the rare signal events that are far less frequent than the background events. So the sequential strategy achieves better performance on the Mock3 dataset. Similarly, in other datasets where the parallel strategy's tighter cut boundaries may incur signal efficiency costs that outweigh the background reduction benefits, the sequential strategy would be expected to achieve better significance.

Both strategies in the Mock4 dataset successfully identify the proper cut boundaries under high correlation, demonstrating that the LCF models remain robust in this setting.

\subsection{Generalization across feature sets}
\label{subsection:generalization_across_feature_sets}

\begin{figure}[htbp]
  \centering
  \includegraphics[width=.24\textwidth]{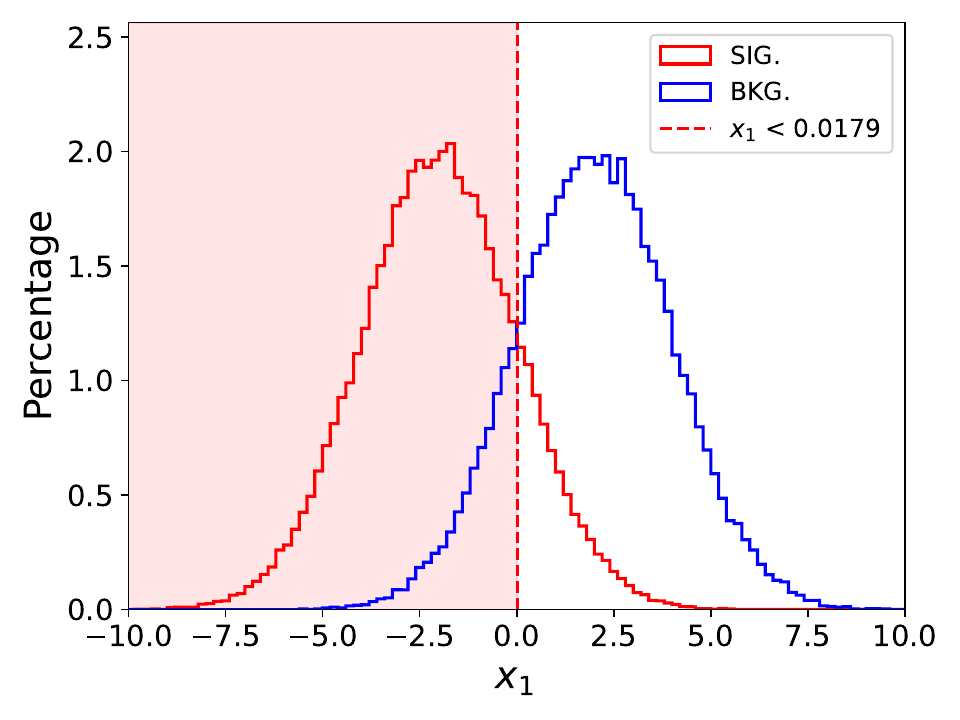}
  \includegraphics[width=.24\textwidth]{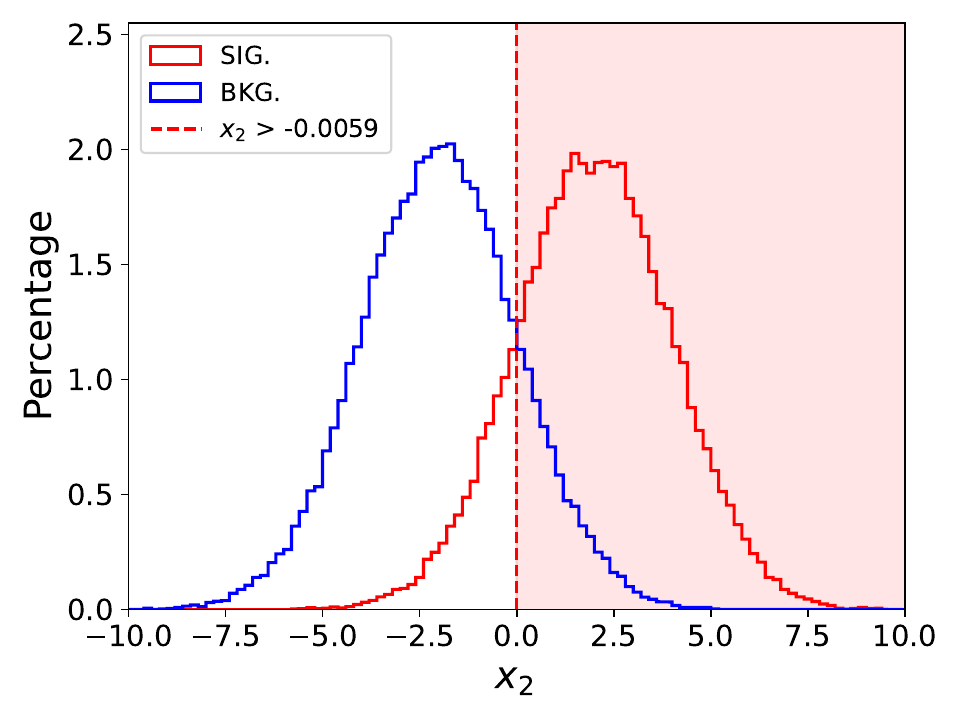}
  \includegraphics[width=.24\textwidth]{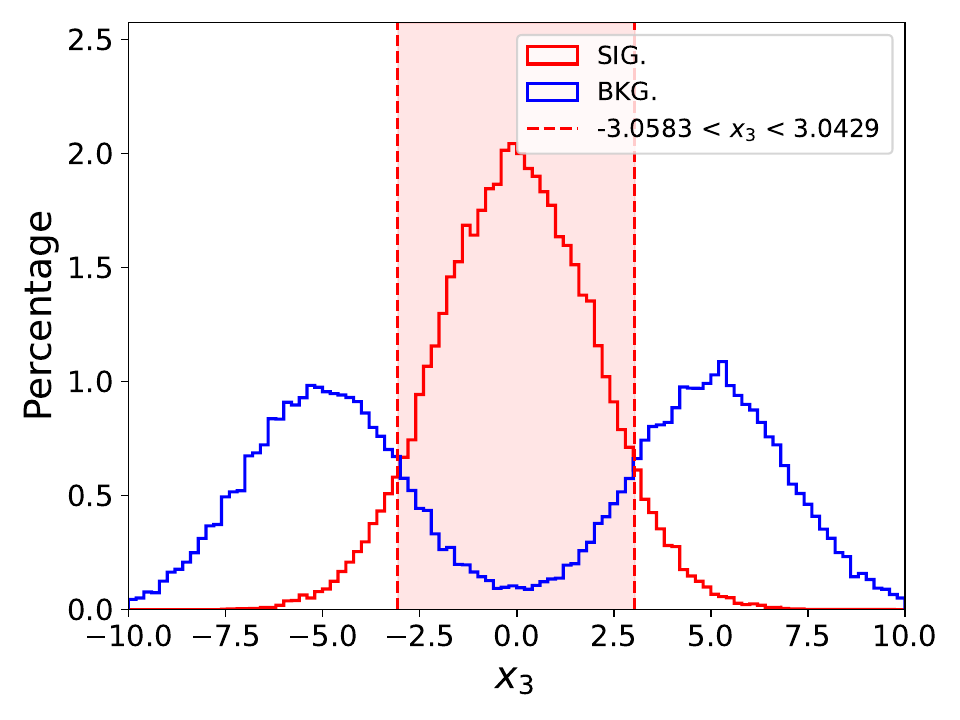}
  \includegraphics[width=.24\textwidth]{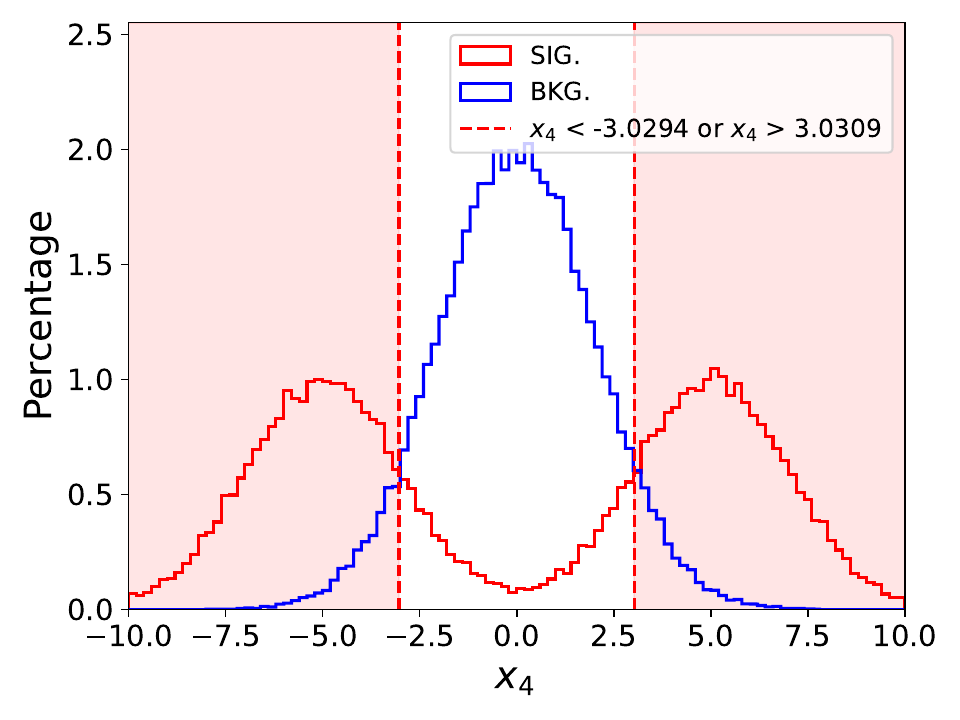}
  \includegraphics[width=.24\textwidth]{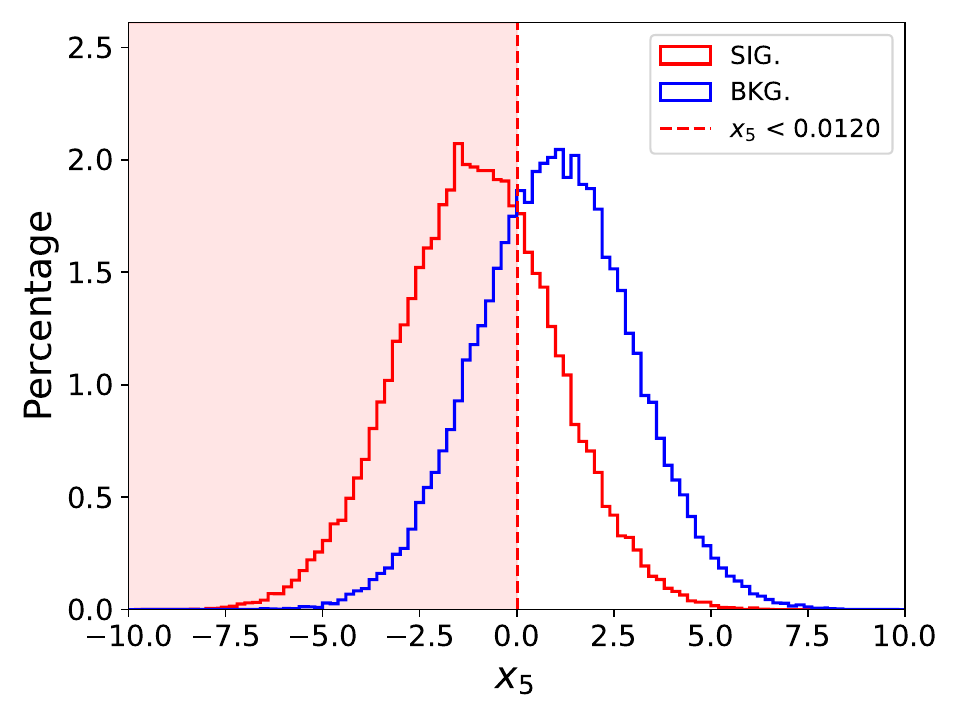}
  \includegraphics[width=.24\textwidth]{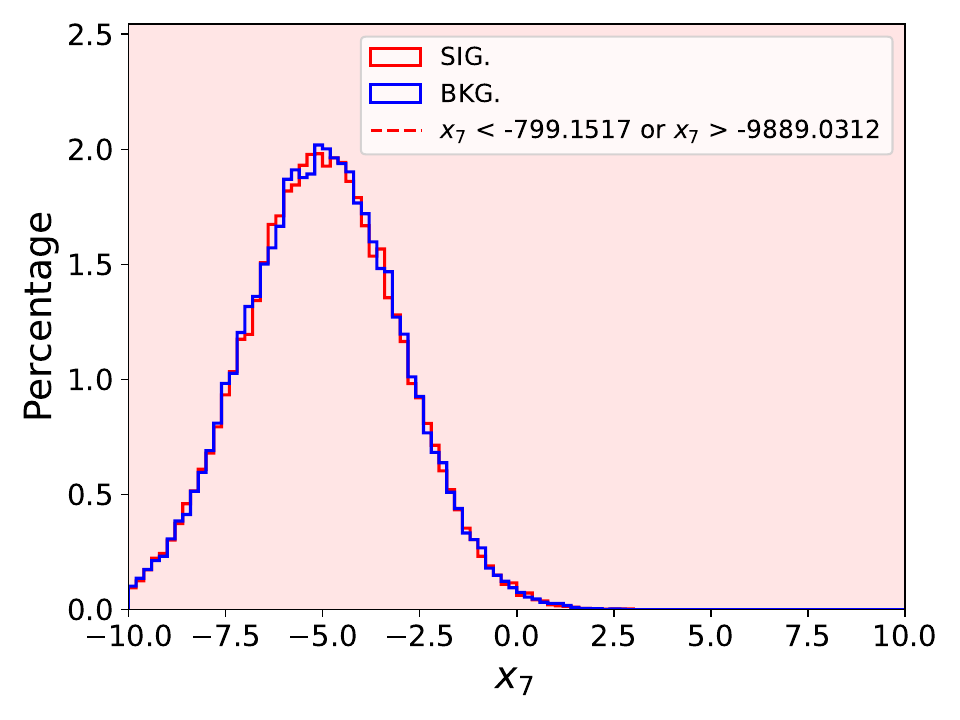}
  \includegraphics[width=.24\textwidth]{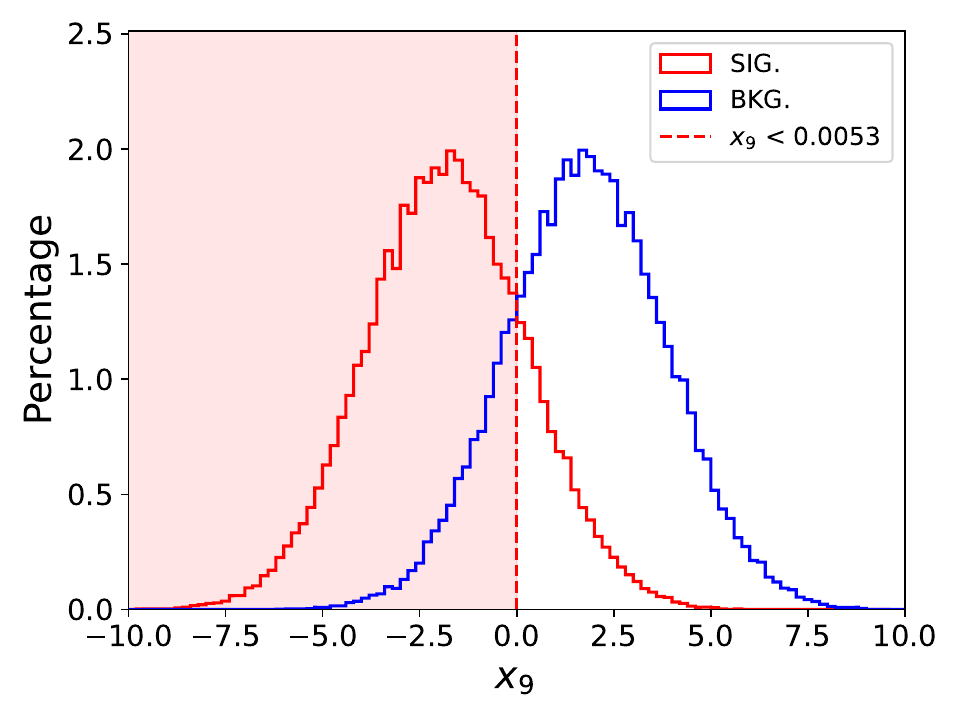}
  \caption{Learned cuts from the parallel LCF on the Mock5 dataset. The error bars are not shown due to the small error or no meaning under the zero importance.}
  \label{figure:learned_cuts-mock5-lcf_par}
\end{figure}

\begin{figure}[htbp]
  \centering
  \includegraphics[width=.24\textwidth]{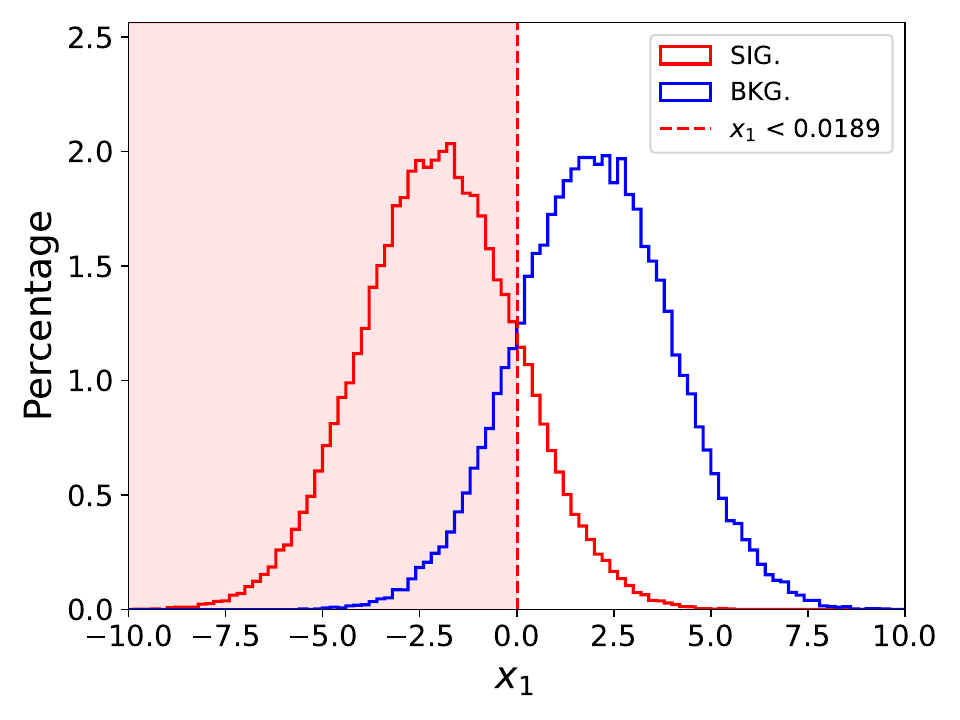}
  \includegraphics[width=.24\textwidth]{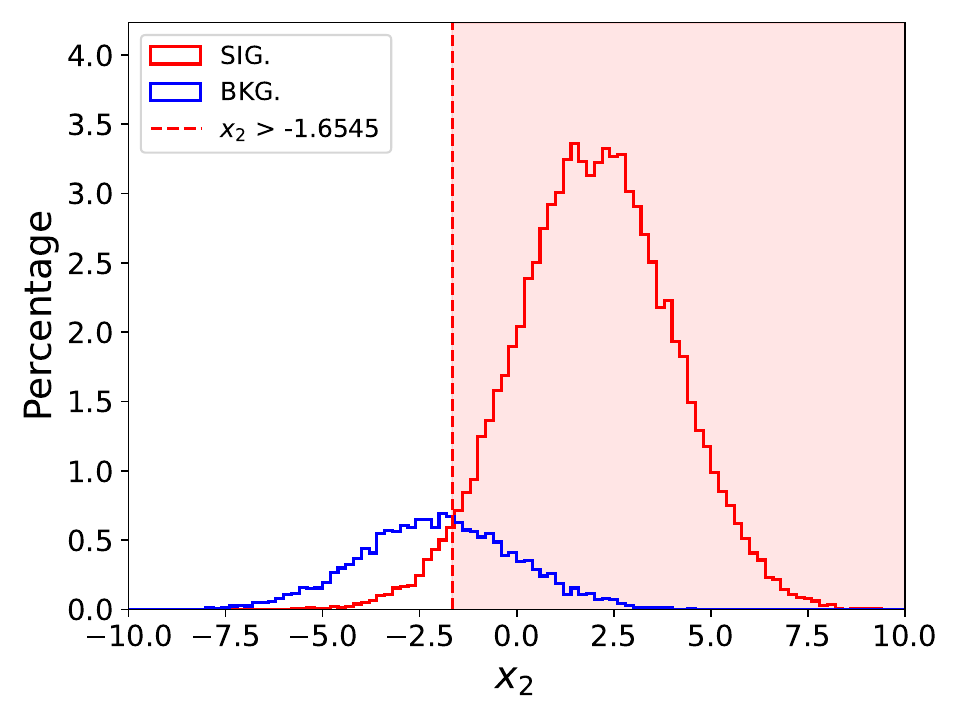}
  \includegraphics[width=.24\textwidth]{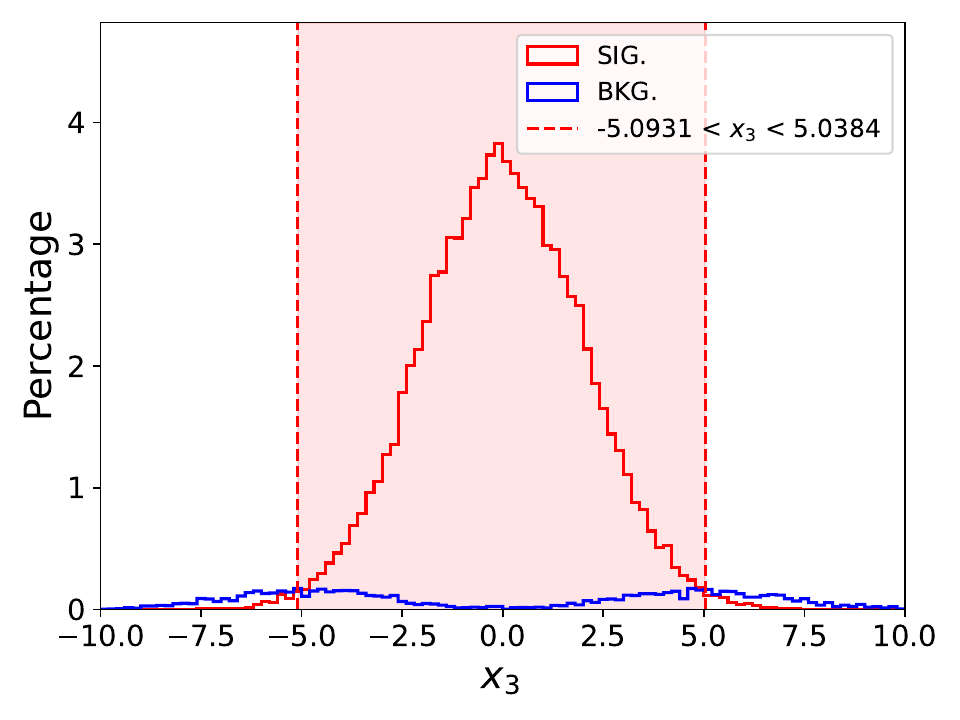}
  \includegraphics[width=.24\textwidth]{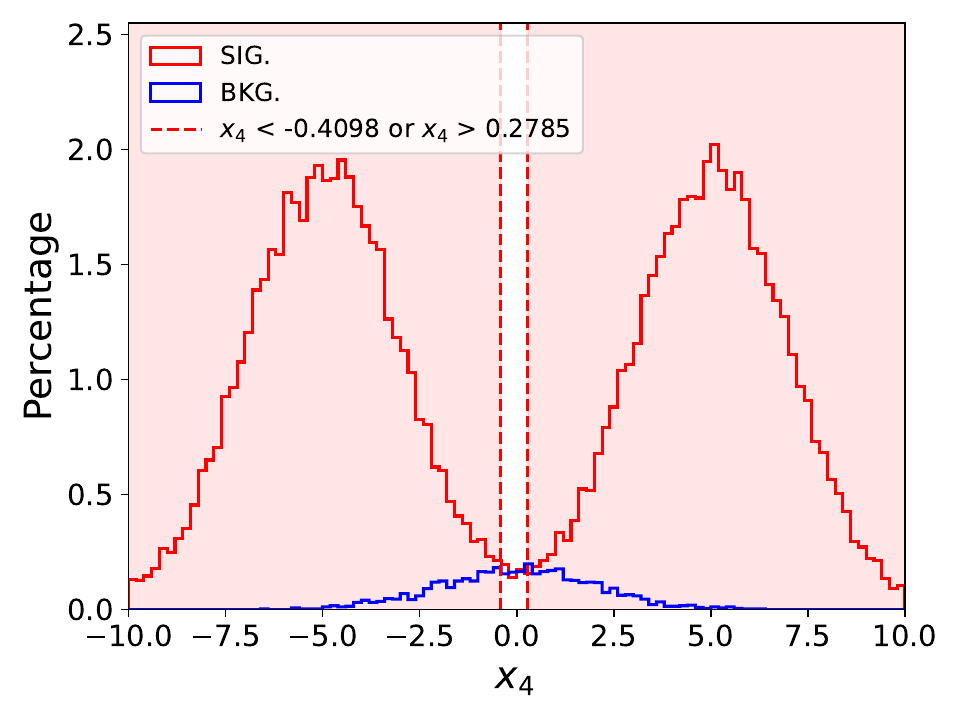}
  \includegraphics[width=.24\textwidth]{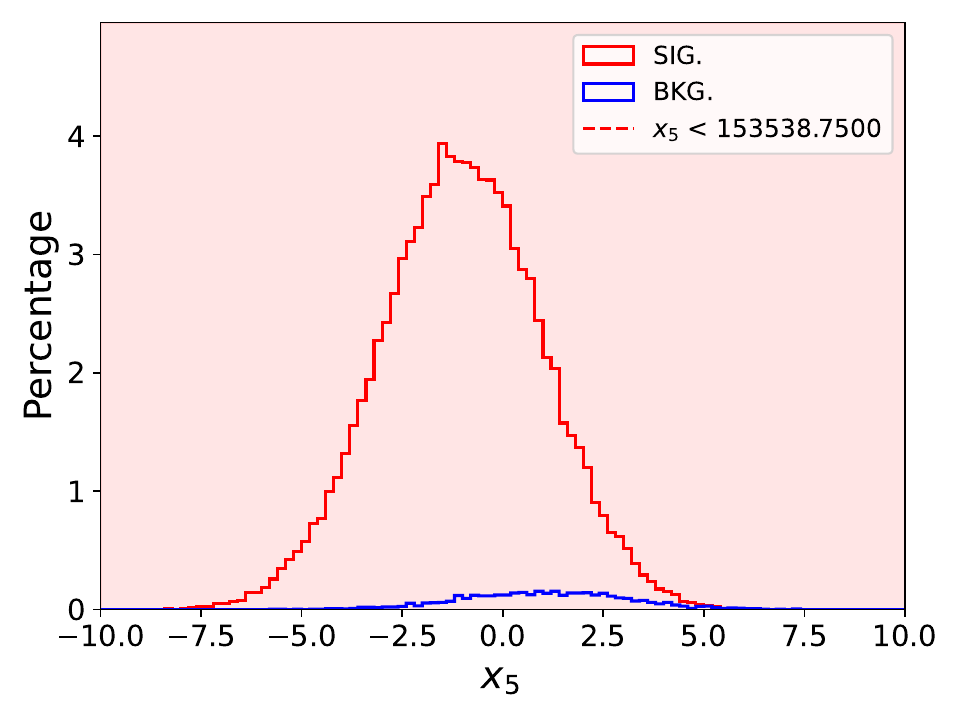}
  \includegraphics[width=.24\textwidth]{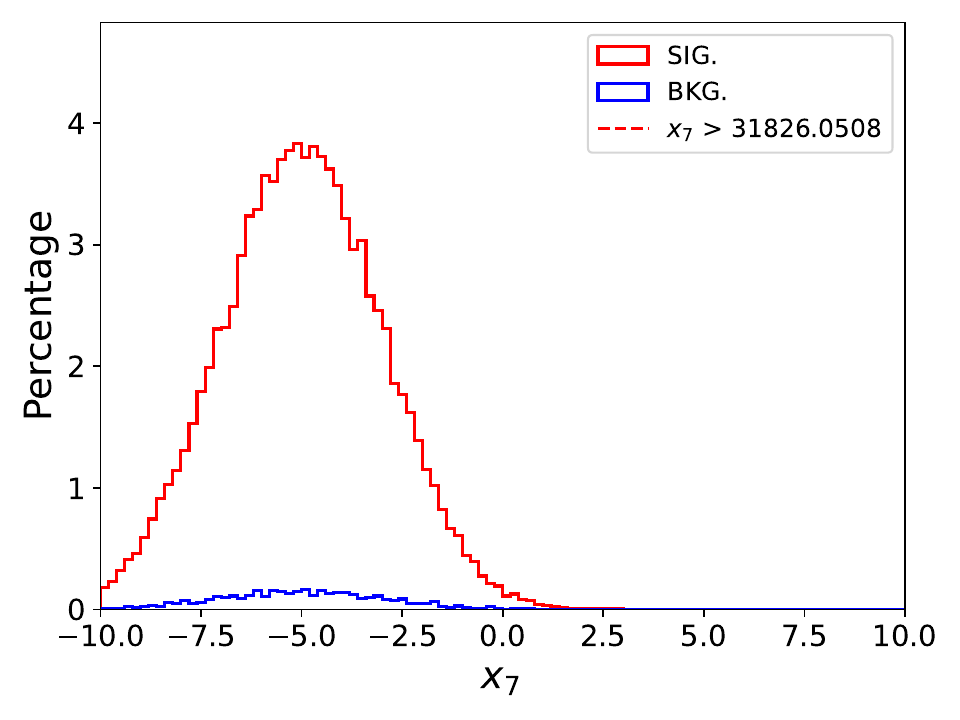}
  \includegraphics[width=.24\textwidth]{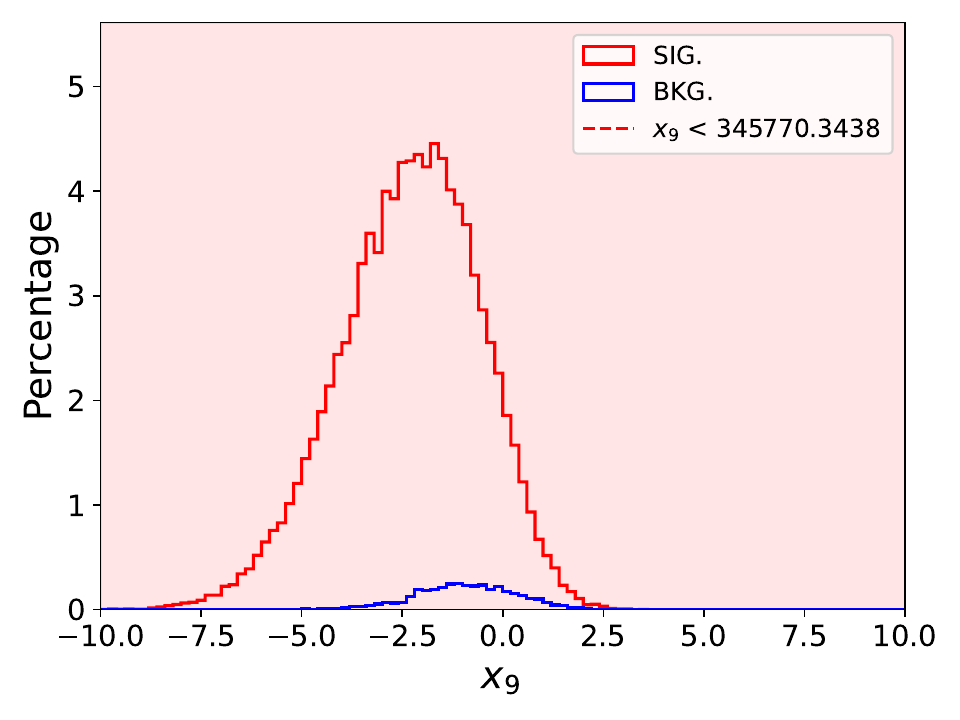}
  \caption{Learned cuts from the sequential LCF on the Mock5 dataset. The error bars are not shown due to the small error or no meaning under the zero importance.}
  \label{figure:learned_cuts-mock5-lcf_seq}
\end{figure}

\begin{figure}[htbp]
  \centering
  \includegraphics[width=.45\textwidth]{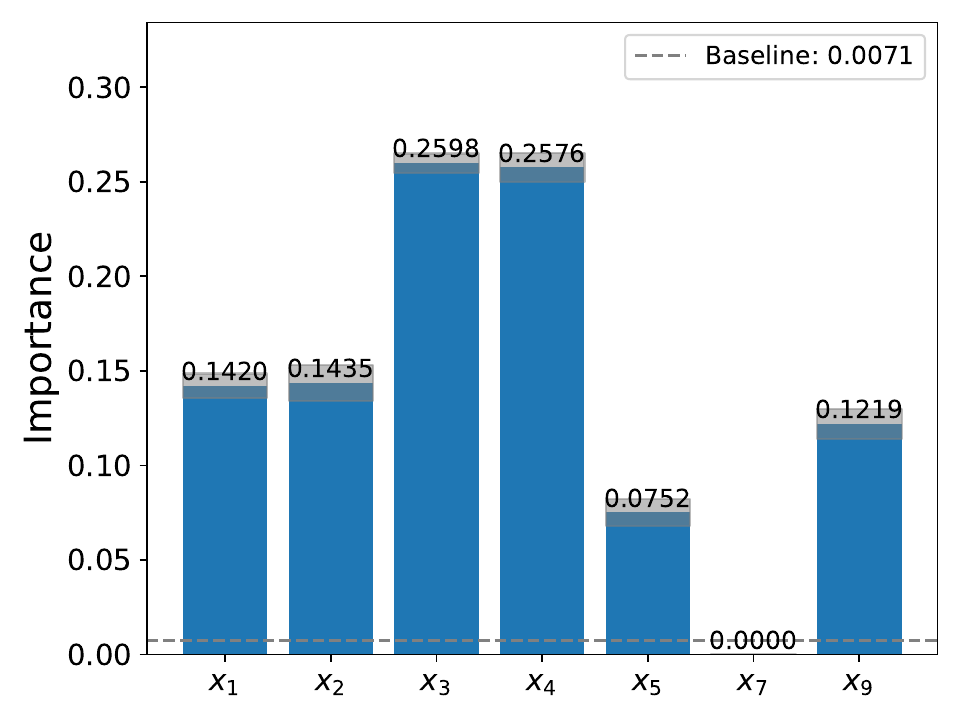} \qquad
  \includegraphics[width=.45\textwidth]{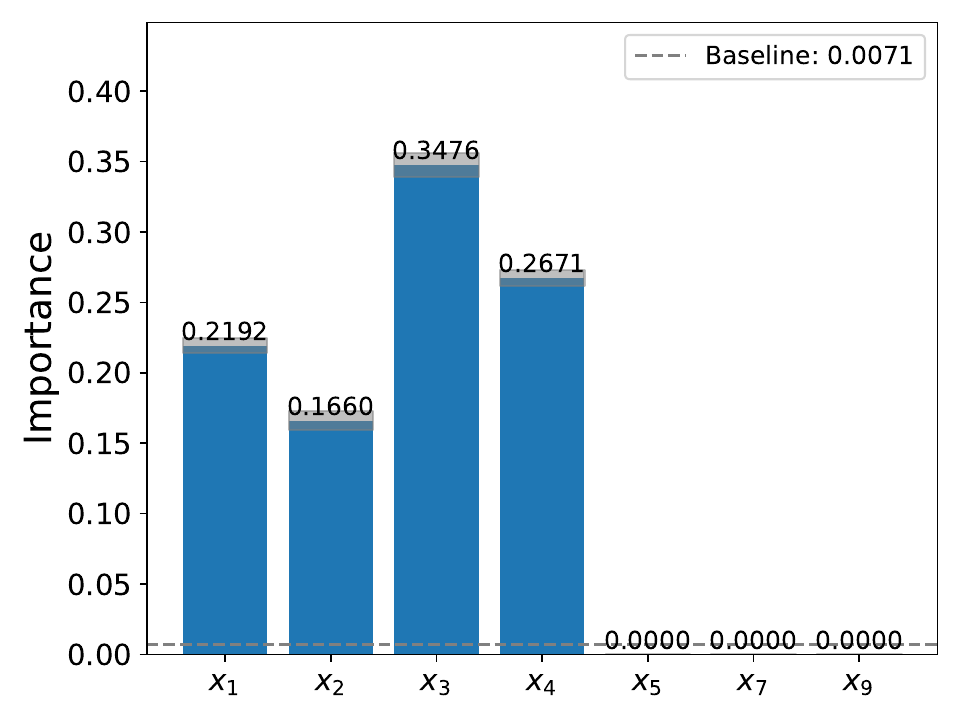}
  \caption{Learned importance from the parallel (left) and sequential (right) LCFs on the Mock5 dataset. The baseline indicates the minimum importance (default: 5\% of average importance = $1/F \times 0.05$), below which the features are ignored during inference.}
  \label{figure:learned_importance-mock5}
\end{figure}

\begin{figure}[htbp]
  \centering
  \includegraphics[width=.24\textwidth]{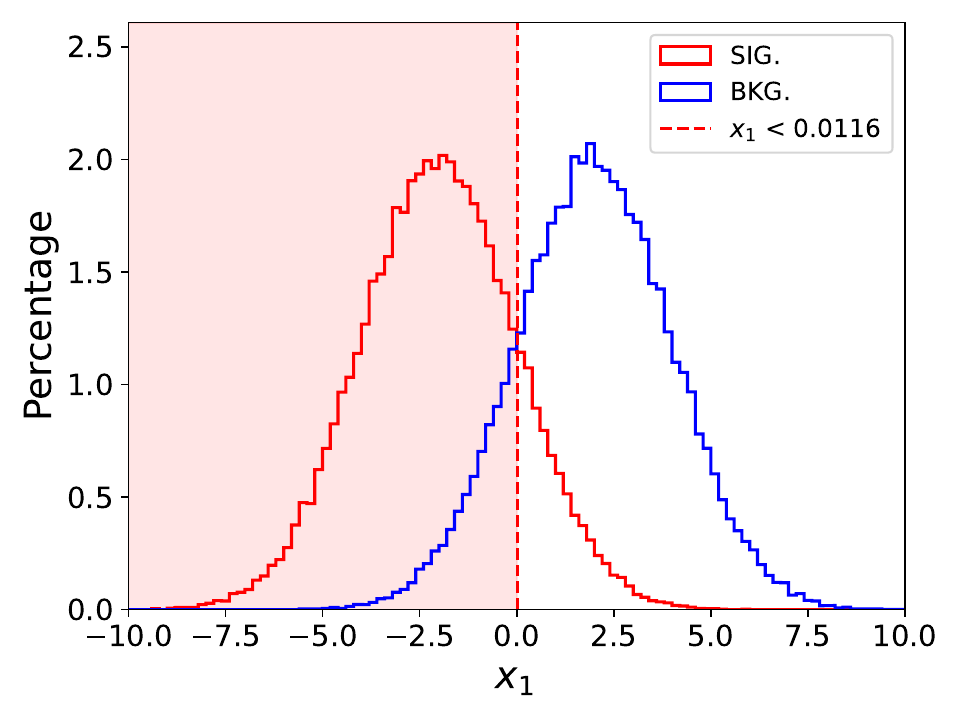}
  \includegraphics[width=.24\textwidth]{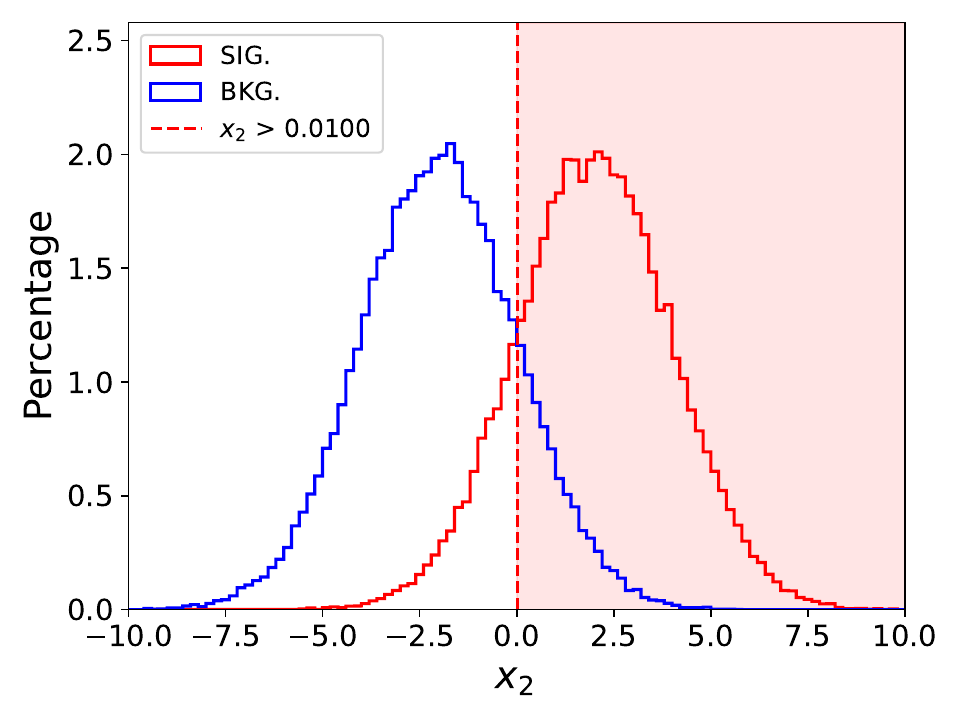}
  \includegraphics[width=.24\textwidth]{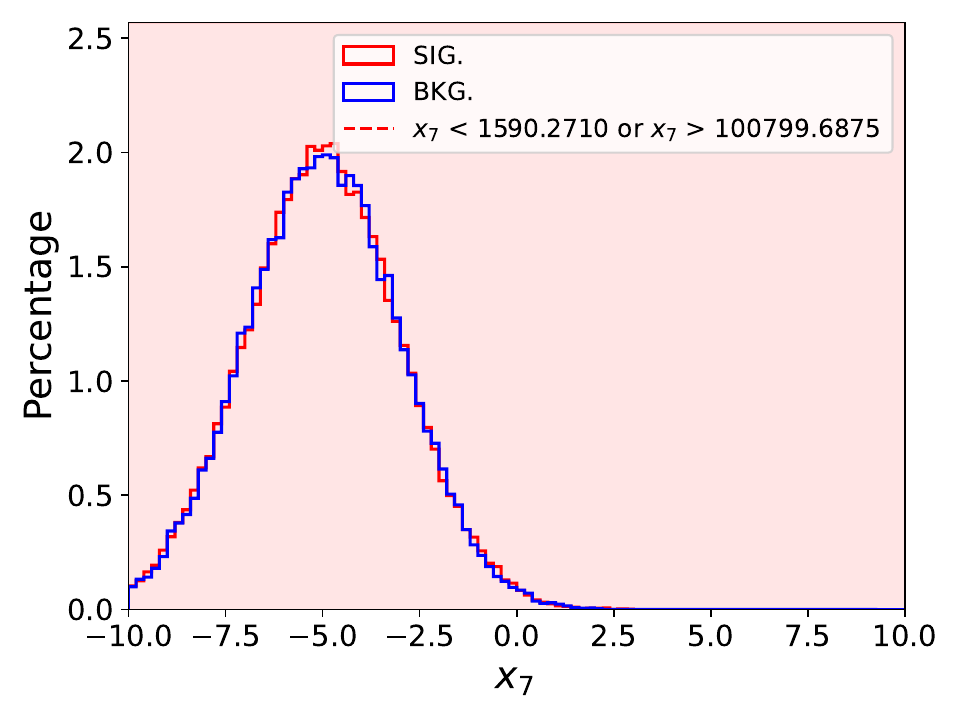}
  \includegraphics[width=.24\textwidth]{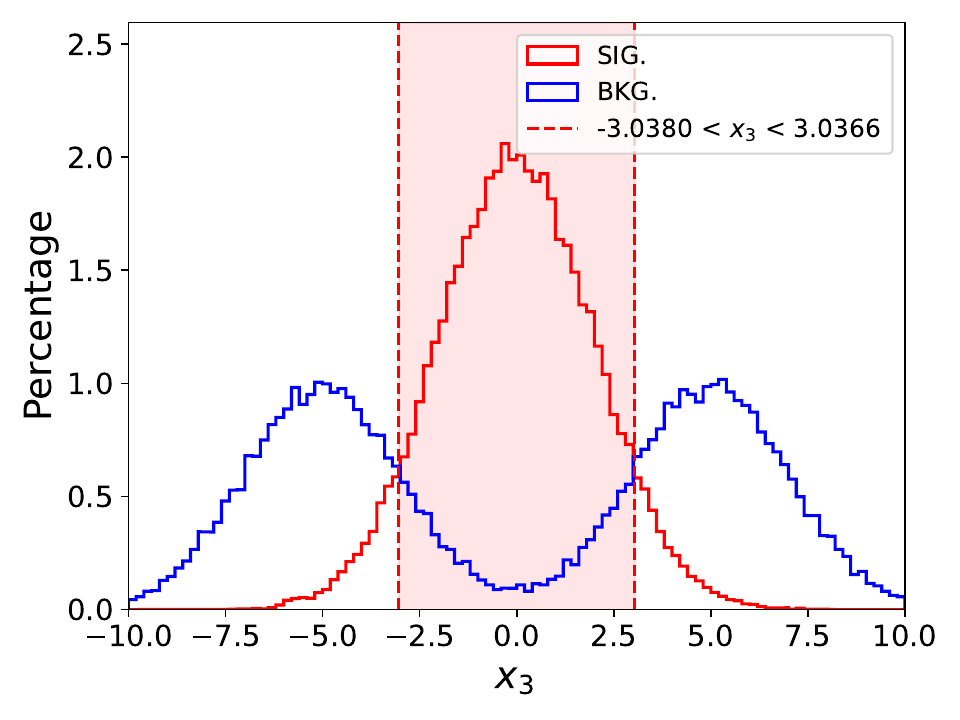}
  \includegraphics[width=.24\textwidth]{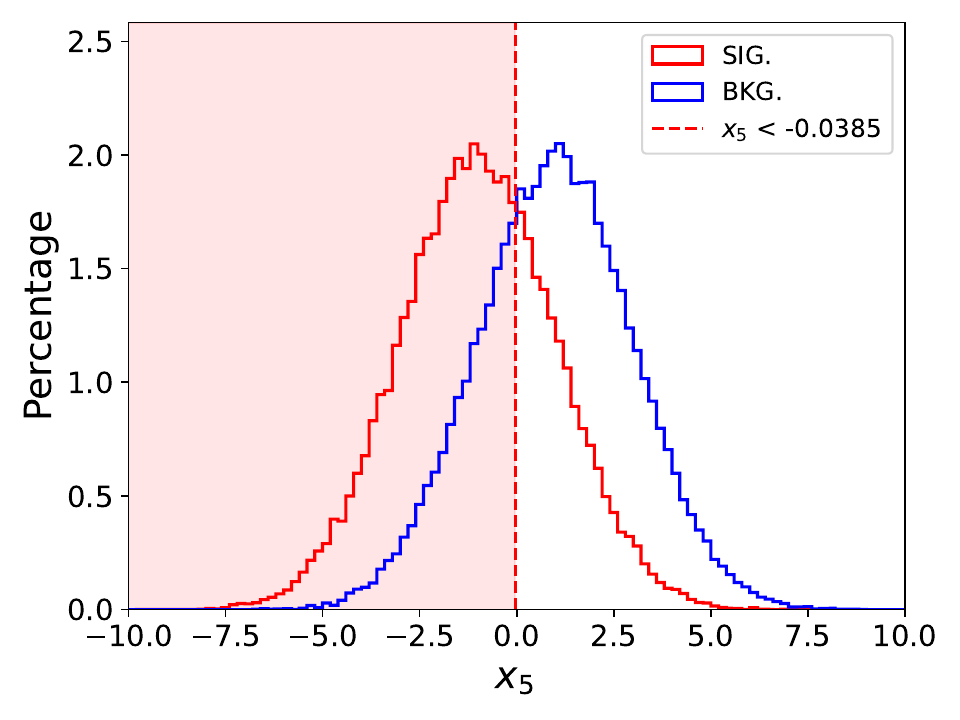}
  \includegraphics[width=.24\textwidth]{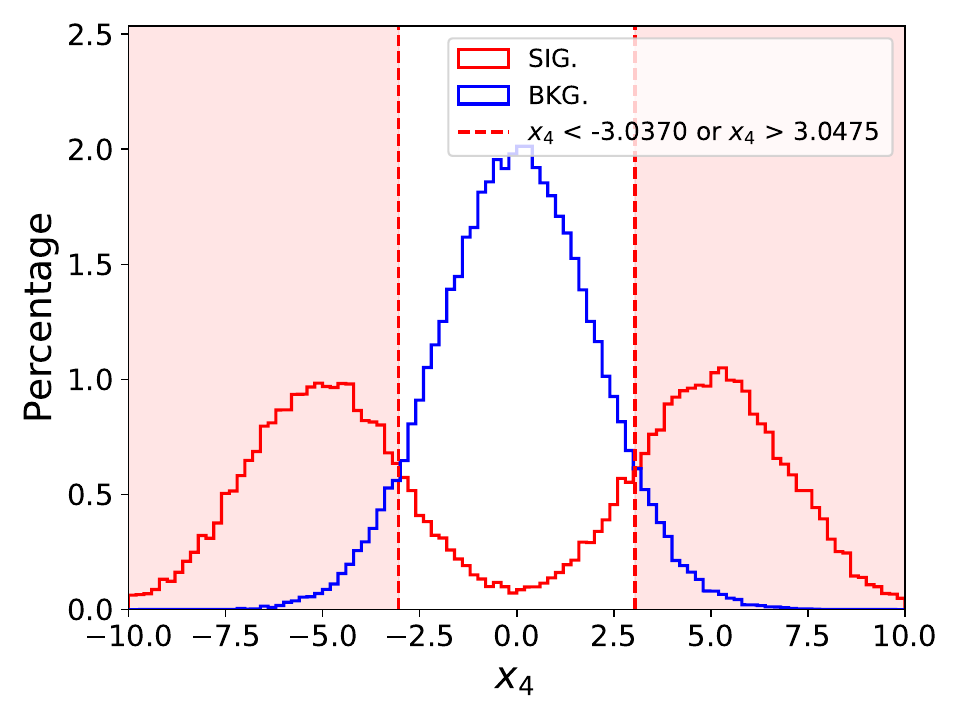}
  \includegraphics[width=.24\textwidth]{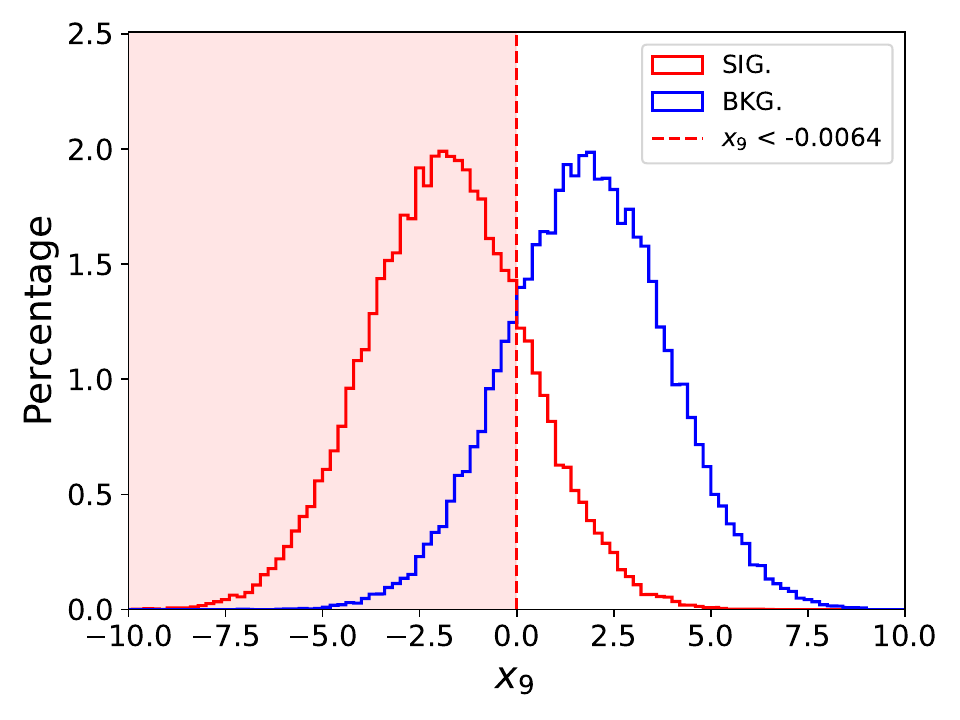}
  \caption{Learned cuts from the parallel LCF on the Mock6 dataset. The error bars are not shown due to the small error or no meaning under the zero importance.}
  \label{figure:learned_cuts-mock6-lcf_par}
\end{figure}

\begin{figure}[htbp]
  \centering
  \includegraphics[width=.24\textwidth]{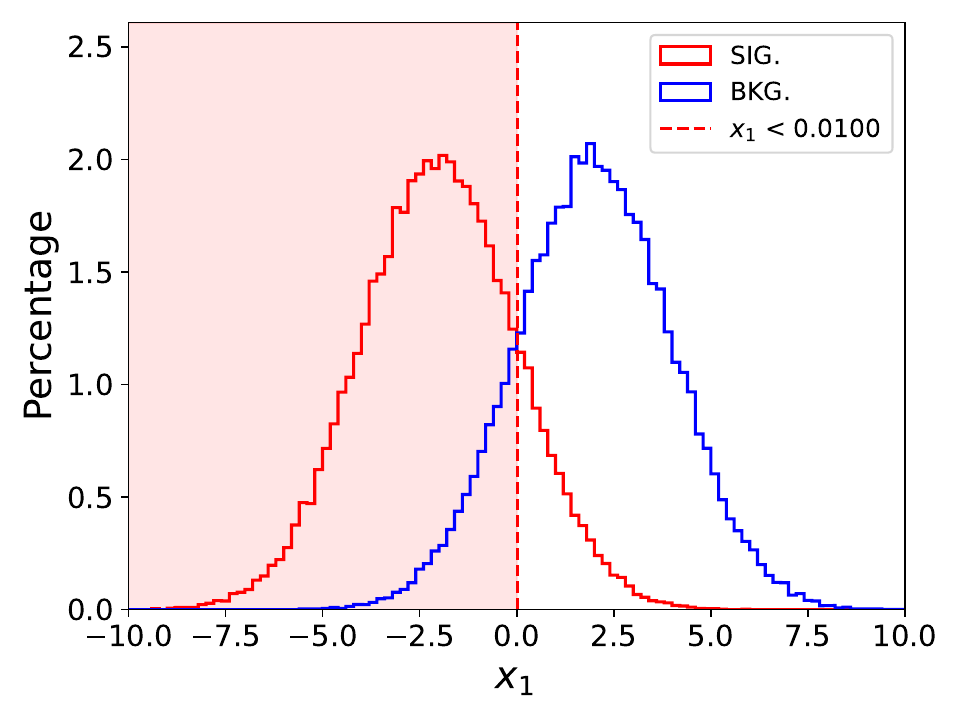}
  \includegraphics[width=.24\textwidth]{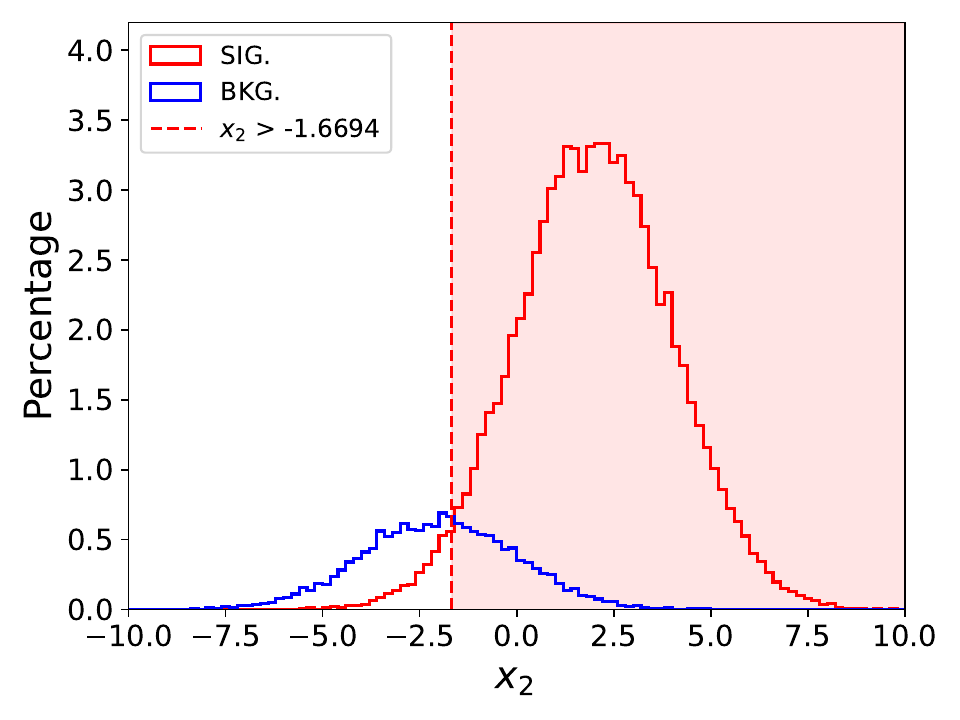}
  \includegraphics[width=.24\textwidth]{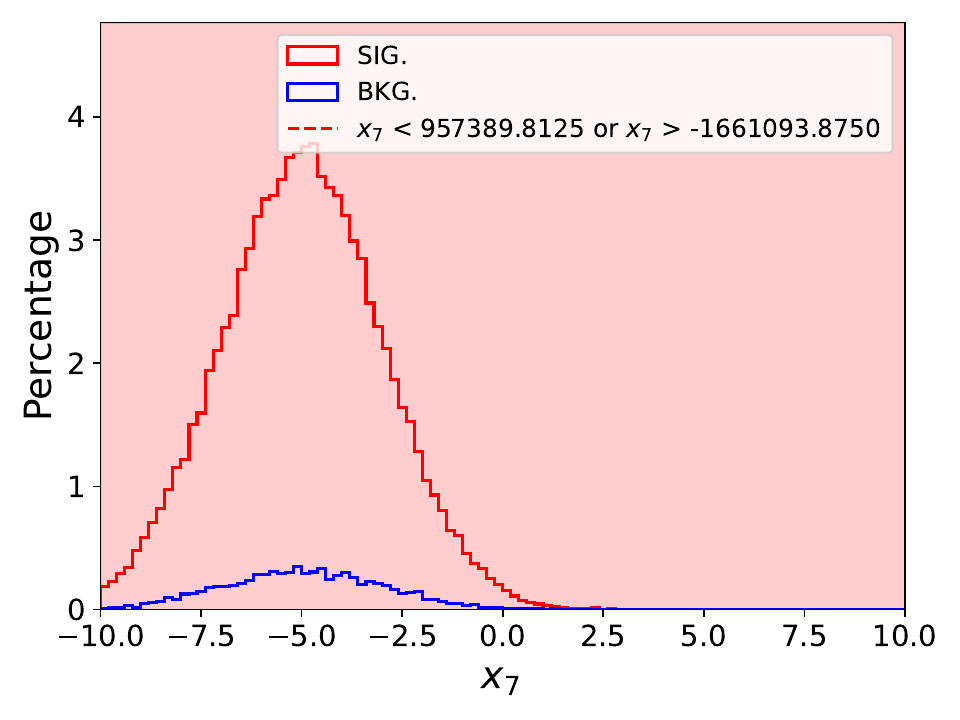}
  \includegraphics[width=.24\textwidth]{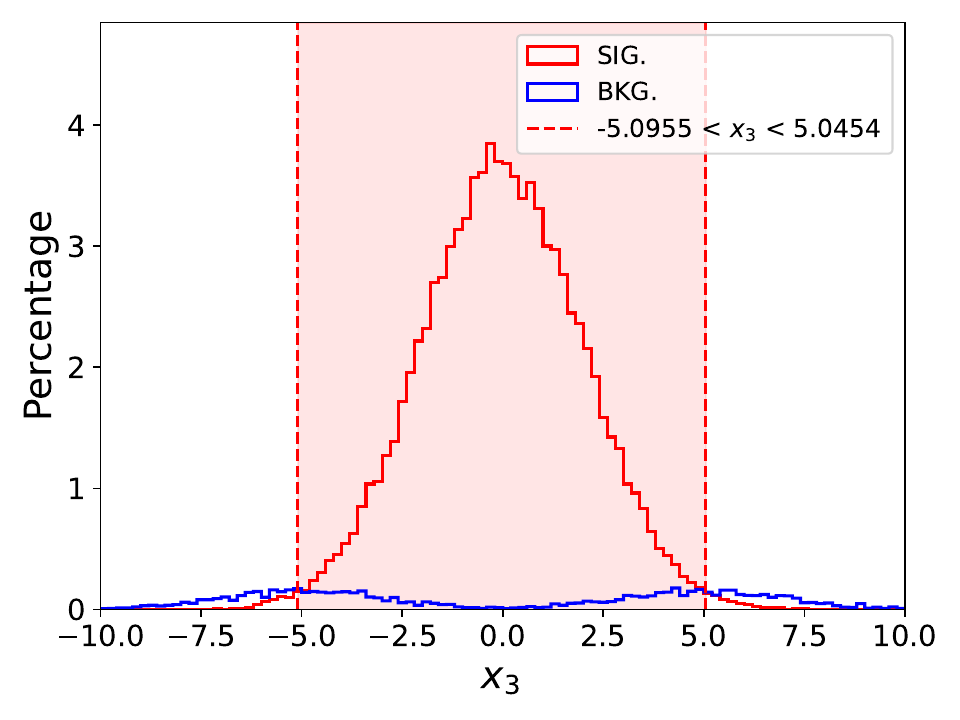}
  \includegraphics[width=.24\textwidth]{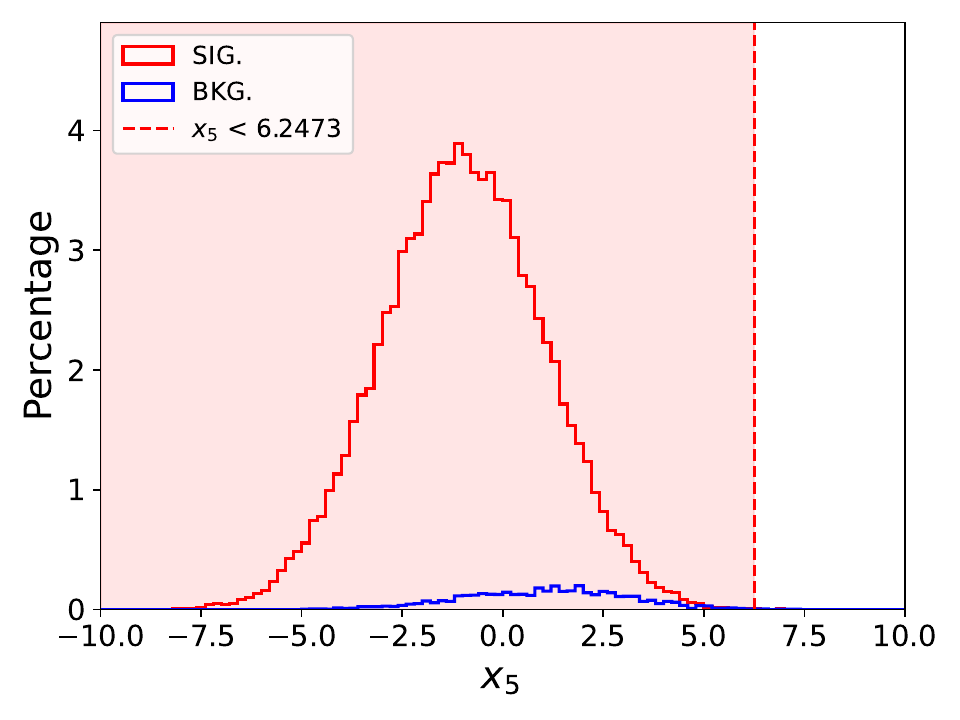}
  \includegraphics[width=.24\textwidth]{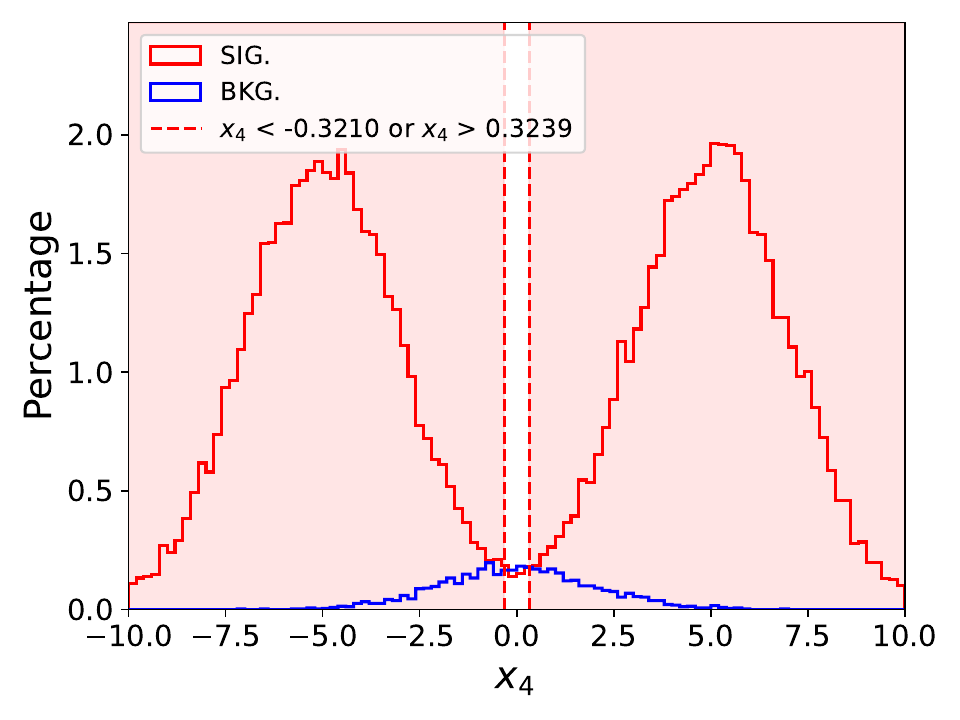}
  \includegraphics[width=.24\textwidth]{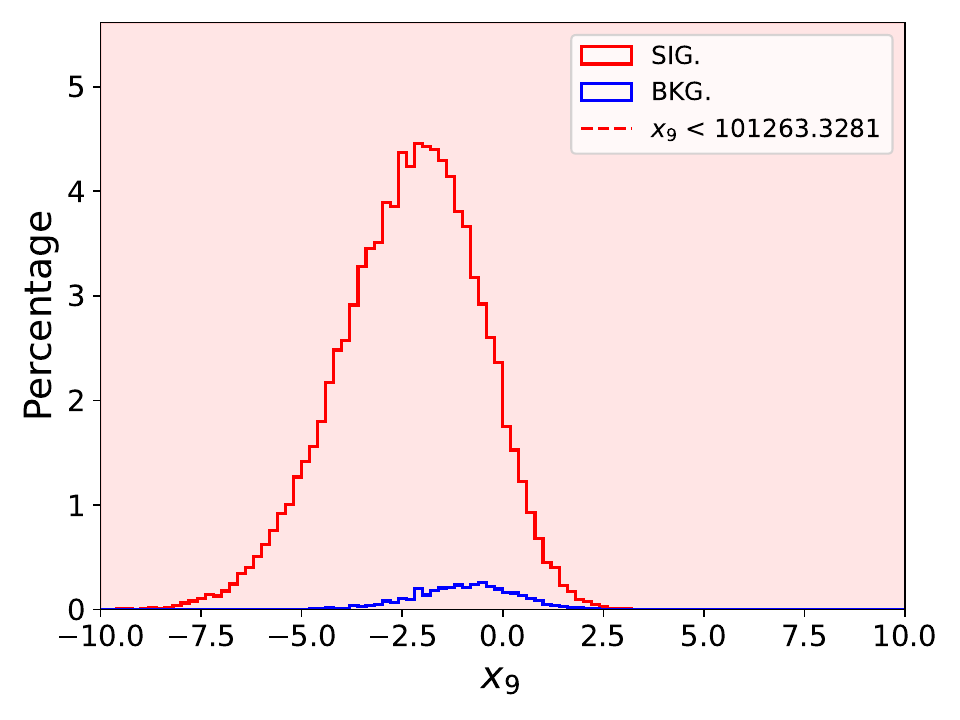}
  \caption{Learned cuts from the sequential LCF on the Mock6 dataset. The error bars are not shown due to the small error or no meaning under the zero importance.}
  \label{figure:learned_cuts-mock6-lcf_seq}
\end{figure}

\begin{figure}[htbp]
  \centering
  \includegraphics[width=.45\textwidth]{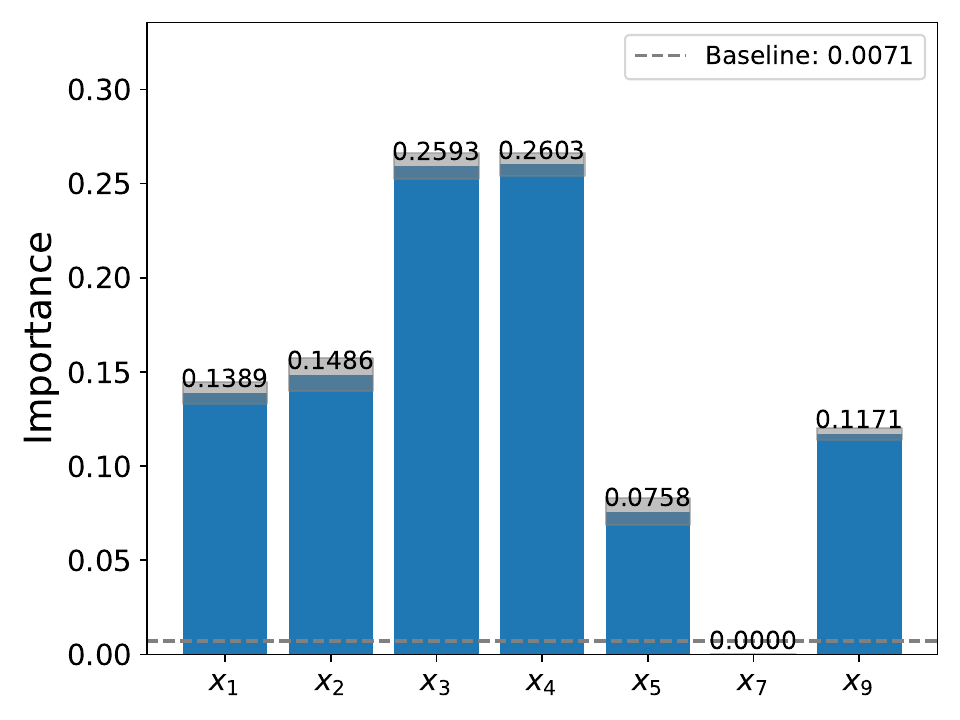} \qquad
  \includegraphics[width=.45\textwidth]{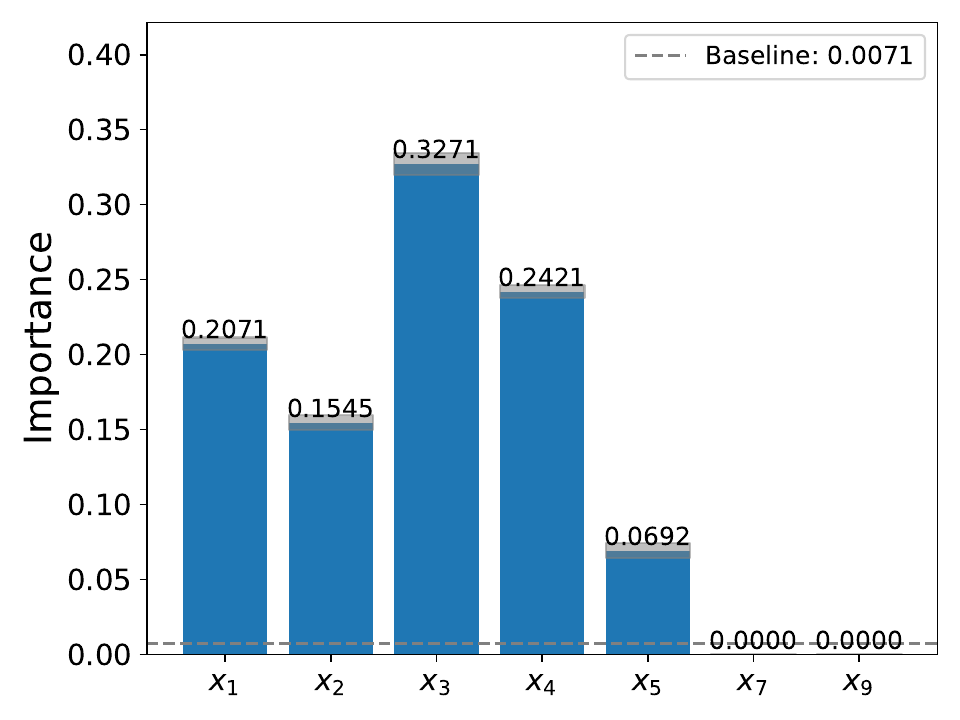}
  \caption{Learned importance from the parallel (left) and sequential (right) LCFs on the Mock6 dataset. The baseline indicates the minimum importance (default: 5\% of average importance = $1/F \times 0.05$), below which the features are ignored during inference.}
  \label{figure:learned_importance-mock6}
\end{figure}

To evaluate the generalization ability, we train the LCF models on the Mock5 and Mock6 datasets. Mock5 consists of a typical combination of features, and Mock6 introduces random reordering of features to test whether the model remains robust to different feature arrangements (here the order of features is $x_1, x_2, x_7, x_3, x_5, x_4, x_9$). The learned cuts on Mock5 are shown in figure~\ref{figure:learned_cuts-mock5-lcf_par} (parallel LCF) and figure~\ref{figure:learned_cuts-mock5-lcf_seq} (sequential LCF) with corresponding learned importance scores shown in figure~\ref{figure:learned_importance-mock5}. The figures related to Mock6 are~\ref{figure:learned_cuts-mock6-lcf_par},~\ref{figure:learned_cuts-mock6-lcf_seq}, and~\ref{figure:learned_importance-mock6}.

In the parallel strategy for Mock5, the learned cuts are consistent with those observed in previous Mock datasets except for feature $x_7$. The difference here is that the weighted baseline defined in eq.~(\ref{eq:learning_process_of_importance}) is lower than in the Mock3 dataset, which includes another redundant feature $x_8$. So the importance of $x_7$ is compressed much harder leading to zero. Unlike the randomly fixed cuts on $x_7$ in section~\ref{subsection:robustness_to_redundancy}, the learned cuts on $x_7$ here are totally random due to the zero importance score which leads to no adjustment to the gradient and ends up with a quite loose cut. The relative importance among $x_1$ to $x_4$ remains similar to Mock1. Feature $x_9$ is more important than $x_5$, which aligns with their signal-background separation. In the sequential strategy, it clearly shows that the cuts on $x_1$, $x_2$, $x_3$, $x_4$ effectively separate signal from background. In contrast, the cuts on $x_5$, $x_7$ and $x_9$ cover the full range, which matches their zero importance scores in figure~\ref{figure:learned_importance-mock5}. The first four cuts are effective and dominate the subsequent cuts, which is consistent with the results in Mock1. The weak feature $x_5$ and highly-correlated feature $x_9$ don't provide any information to the final classification according to their large boundary.

The results from the reordered dataset Mock6 are interesting. In the parallel strategy, since every feature is treated separately, the importance scores are nearly identical to those in Mock5 (figure~\ref{figure:learned_importance-mock6}). In contrast, in the sequential strategy, the presence of a less informative feature like $x_7$ early in the sequence doesn't prevent the model from converging correctly. The cut boundaries for $x_1$, $x_2$, $x_3$, $x_4$ remain clear and effective while $x_7$ and $x_9$ cover the full range as in Mock5. The cut on $x_5$ is the only one changed compared to Mock5. The key difference is $x_4$ is applied later than $x_5$ and $x_5$ doesn't generate that large loss like $x_7$, so the importance of $x_5$ is not compressed that hard like in Mock5. The change of applying order and the nature of the feature itself are the main reason for the difference.

\begin{table}[htbp]
  \centering
  \begin{tabular}{|l|c|c|c|c|c|}
    \hline
    \textbf{Model} & \textbf{TP} & \textbf{FP} & \textbf{Accuracy} & \textbf{Precision} & \textbf{Significance} \\
    \hline
    LCF (Parallel) & $16532 \pm 51$ & $6 \pm 0$ & $66.7 \pm 0.1\%$ & $100.0 \pm 0.0\%$ & $52.1 \pm 1.1$ \\
    LCF (Sequential) & $39769 \pm 40$ & $1571 \pm 25$ & $88.3 \pm 0.0\%$ & $96.2 \pm 0.1\%$ & $7.8 \pm 0.1$ \\
    \hline
  \end{tabular}
  \caption{Performance metrics of LCF models on the Mock5 dataset.}
  \label{table:metrics_mock5}
\end{table}

\begin{table}[htbp]
  \centering
  \begin{tabular}{|l|c|c|c|c|c|}
    \hline
    \textbf{Model} & \textbf{TP} & \textbf{FP} & \textbf{Accuracy} & \textbf{Precision} & \textbf{Significance} \\
    \hline
    LCF (Parallel) & $16245 \pm 94$ & $6 \pm 0$ & $66.1 \pm 0.1\%$ & $100.0 \pm 0.0\%$ & $50.8 \pm 1.0$ \\
    LCF (Sequential) & $40116 \pm 22$ & $1552 \pm 13$ & $88.4 \pm 0.0\%$ & $96.3 \pm 0.0\%$ & $7.9 \pm 0.0$ \\
    \hline
  \end{tabular}
  \caption{Performance metrics of LCF models on the Mock6 dataset.}
  \label{table:metrics_mock6}
\end{table}

Tables~\ref{table:metrics_mock5} and \ref{table:metrics_mock6} show clearly that the results of the sequential strategy are more stable than those of the parallel strategy. Specifically, the sequential strategy demonstrates lower variance across multiple training runs, as evidenced by smaller standard deviations in the significance values: $7.8 \pm 0.1$ and $7.9 \pm 0.0$ for Mock5 and Mock6 respectively, compared to $52.1 \pm 1.1$ and $50.8 \pm 1.0$ for the parallel strategy. Meanwhile, the parallel strategy achieves far better significance than the sequential strategy following the same logic stated in section~\ref{subsection:robustness_to_high_correlation}.

These findings show that our models are robust when encountering all typical features: informative, redundant, highly correlated, and robust to feature ordering.

\subsection{Benchmark on diboson classification}
\label{subsection:benchmark_on_diboson_classification}

\begin{figure}[htbp]
  \centering
  \includegraphics[width=.32\textwidth]{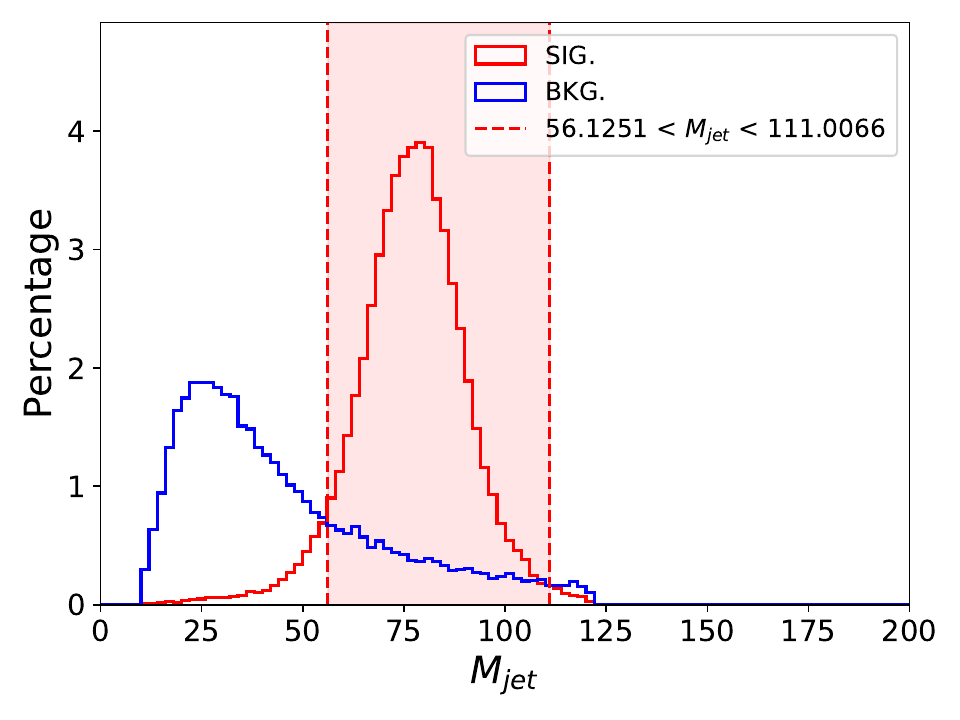}
  \includegraphics[width=.32\textwidth]{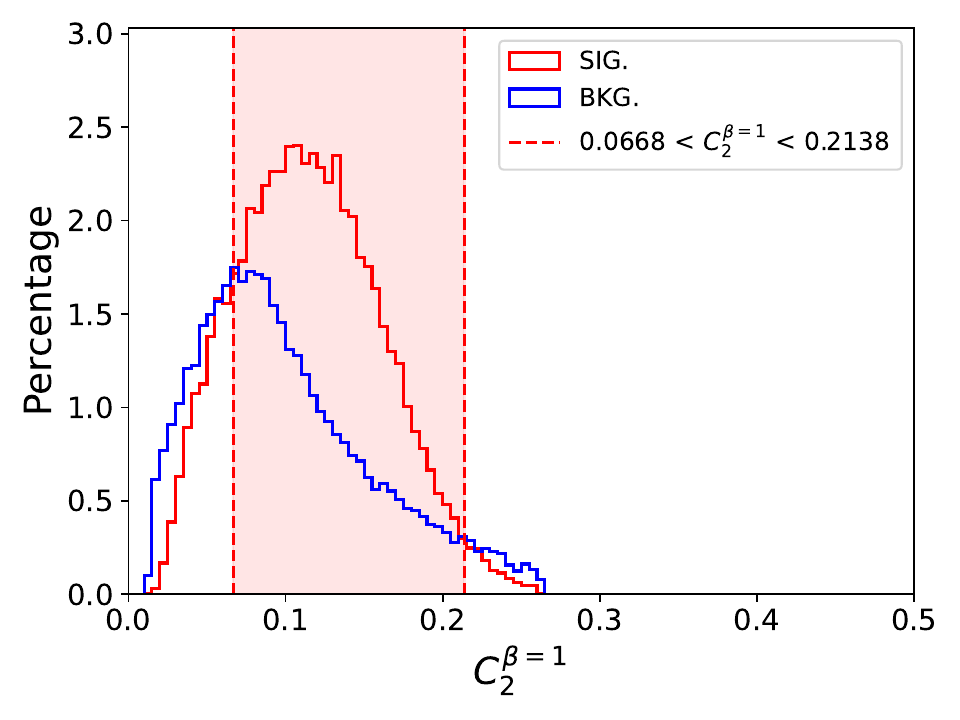}
  \includegraphics[width=.32\textwidth]{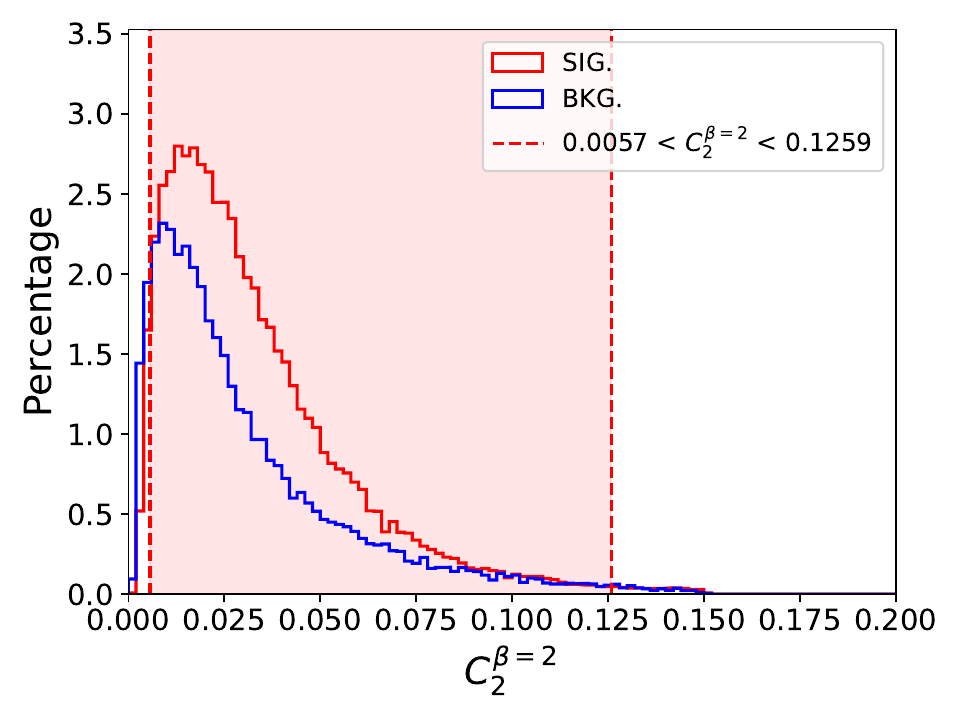}
  \includegraphics[width=.32\textwidth]{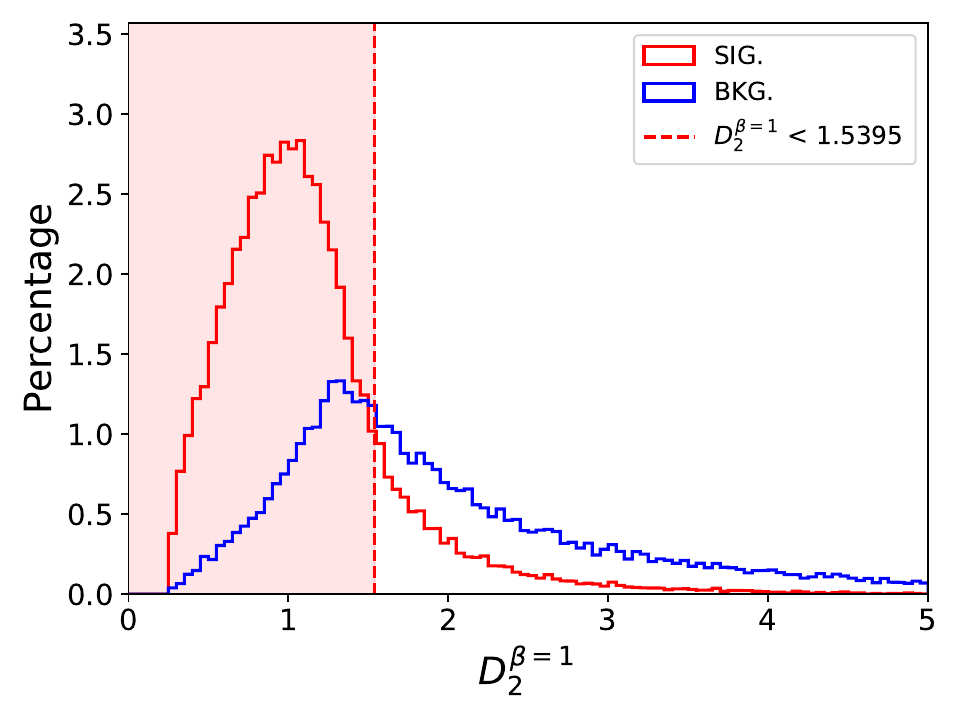}
  \includegraphics[width=.32\textwidth]{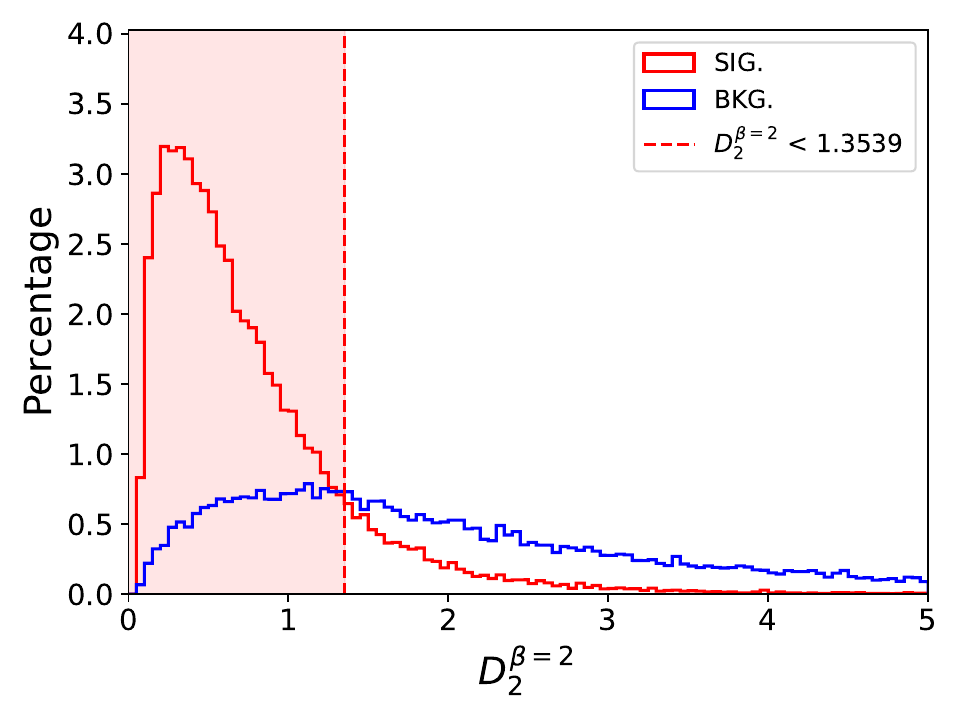}
  \includegraphics[width=.32\textwidth]{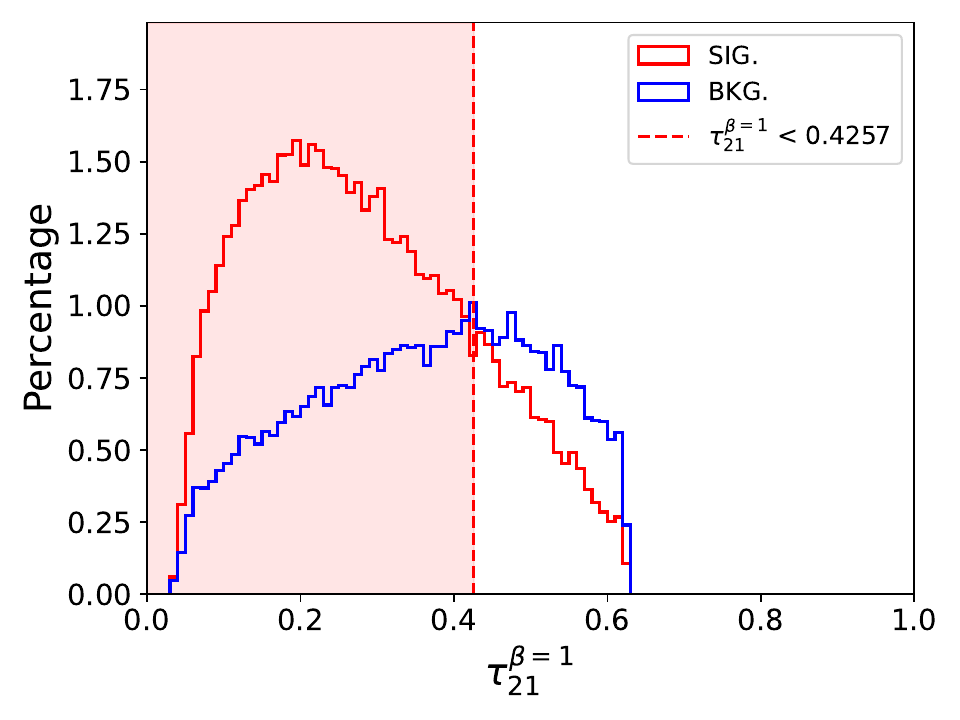}
  \caption{Learned cuts from the parallel LCF on the diboson dataset.}
  \label{figure:learned_cuts-real1-lcf_par}
\end{figure}

\begin{figure}[htbp]
  \centering
  \includegraphics[width=.32\textwidth]{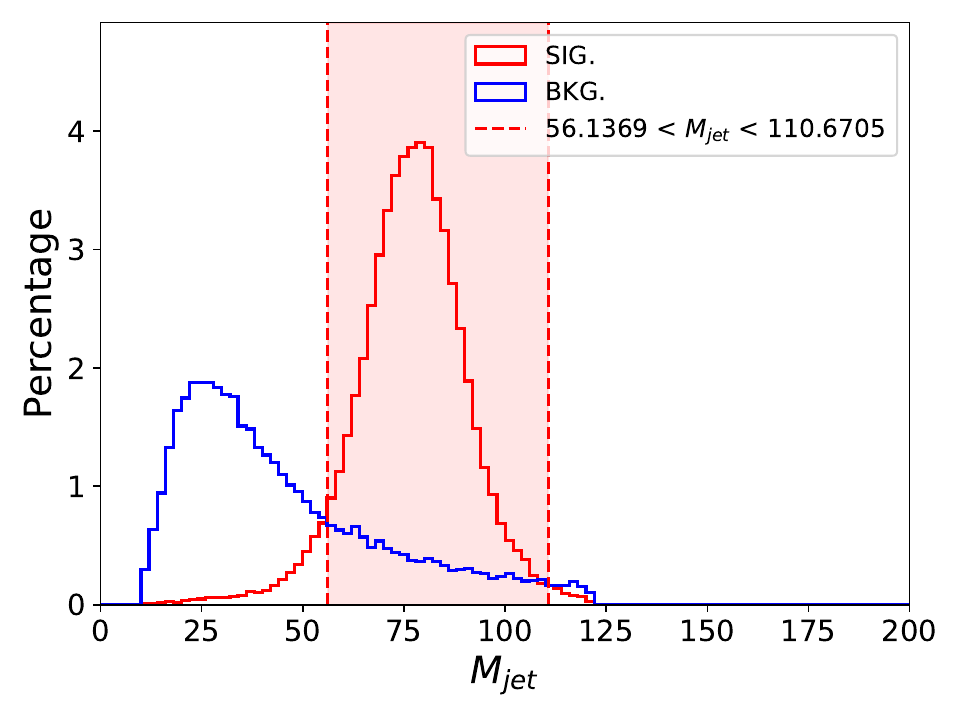}
  \includegraphics[width=.32\textwidth]{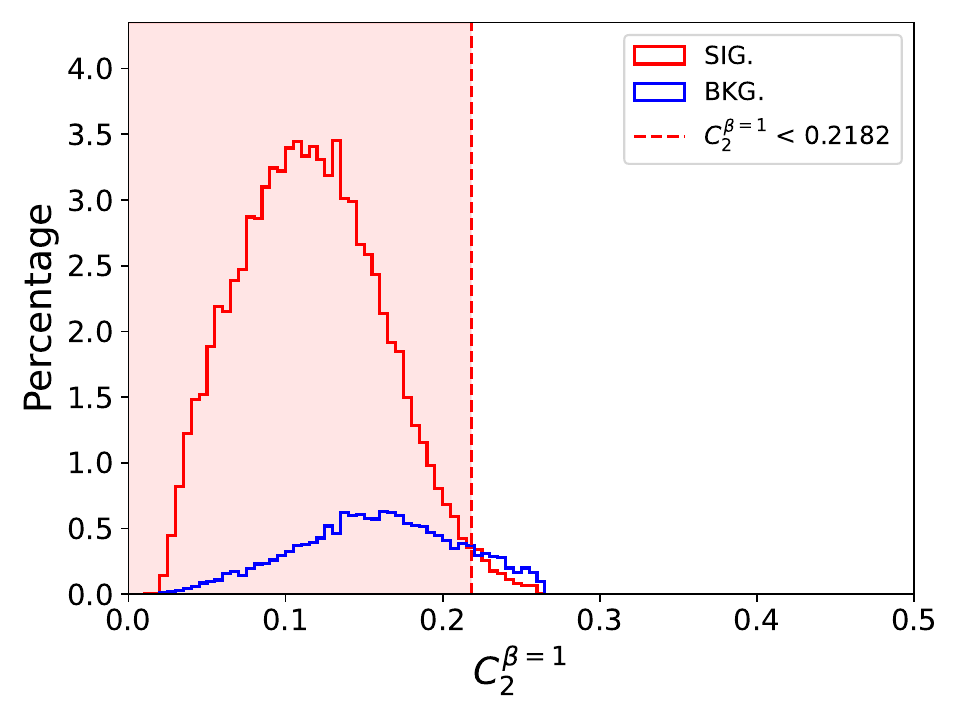}
  \includegraphics[width=.32\textwidth]{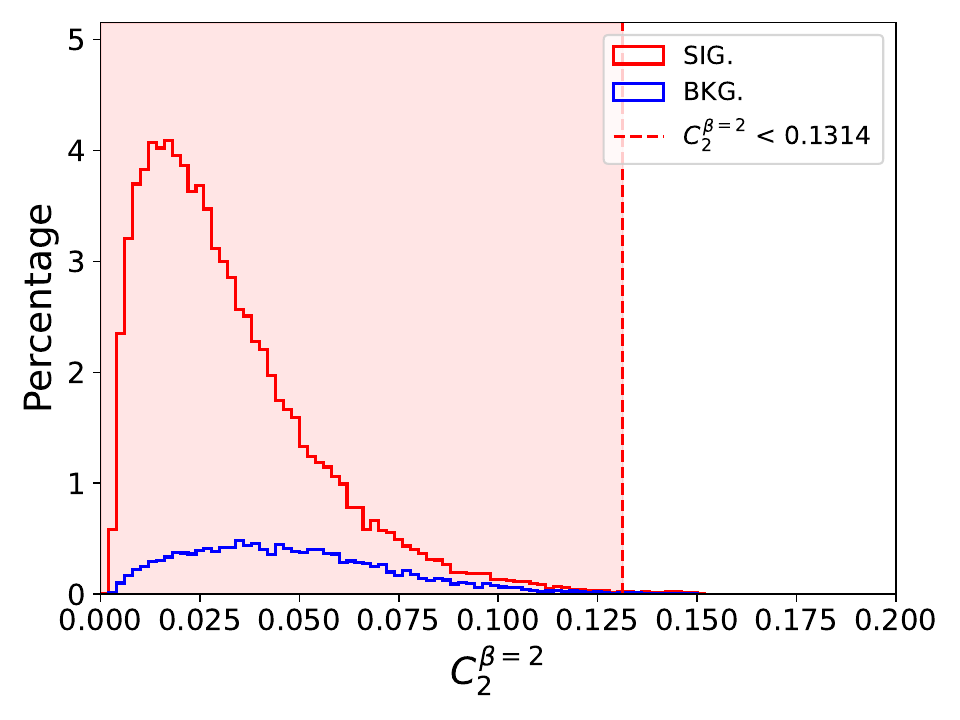}
  \includegraphics[width=.32\textwidth]{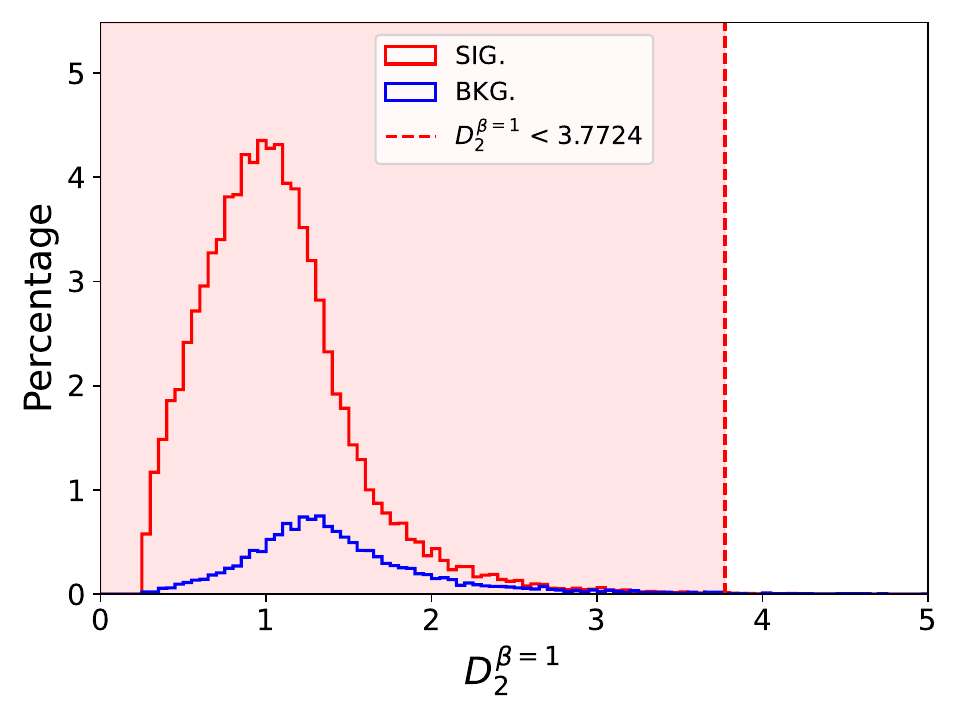}
  \includegraphics[width=.32\textwidth]{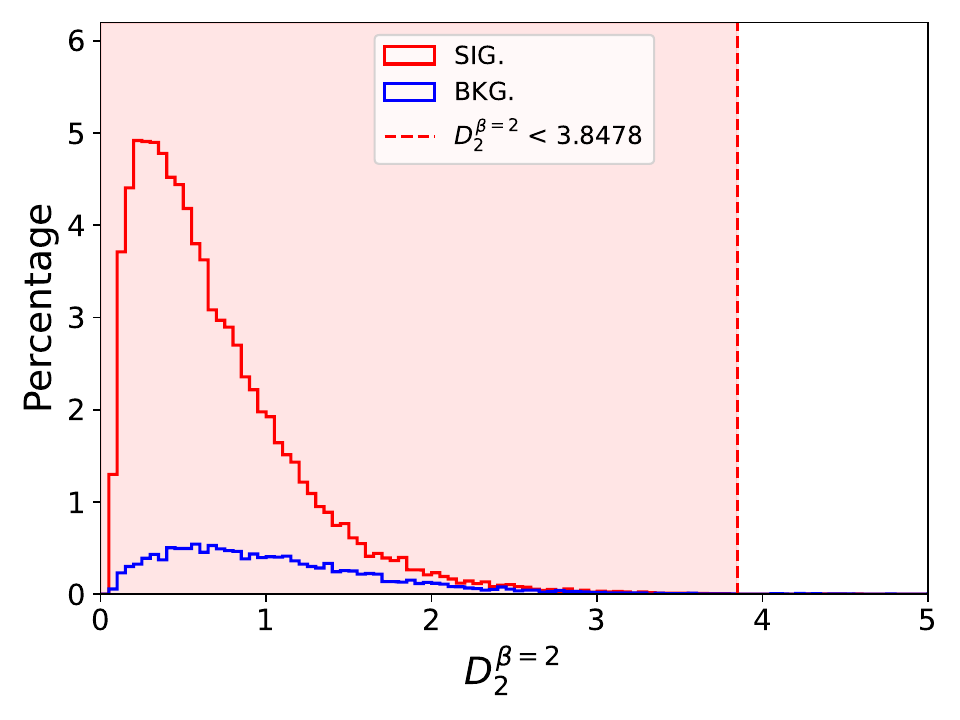}
  \includegraphics[width=.32\textwidth]{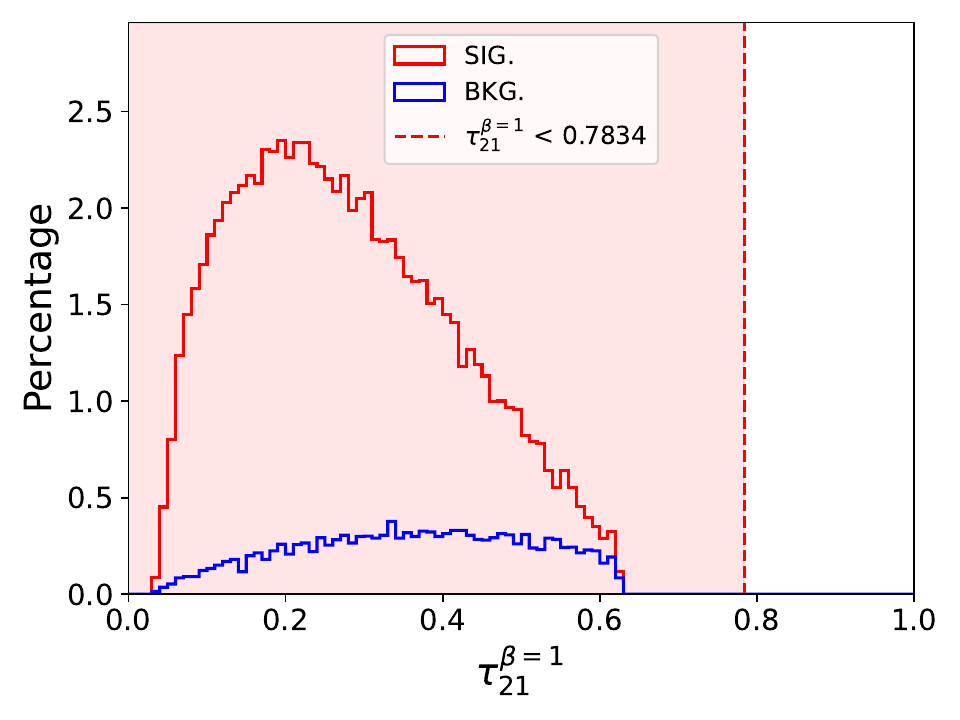}
  \caption{Learned cuts from the sequential LCF on the diboson dataset.}
  \label{figure:learned_cuts-real1-lcf_seq}
\end{figure}

\begin{figure}[htbp]
  \centering
  \includegraphics[width=0.45\textwidth]{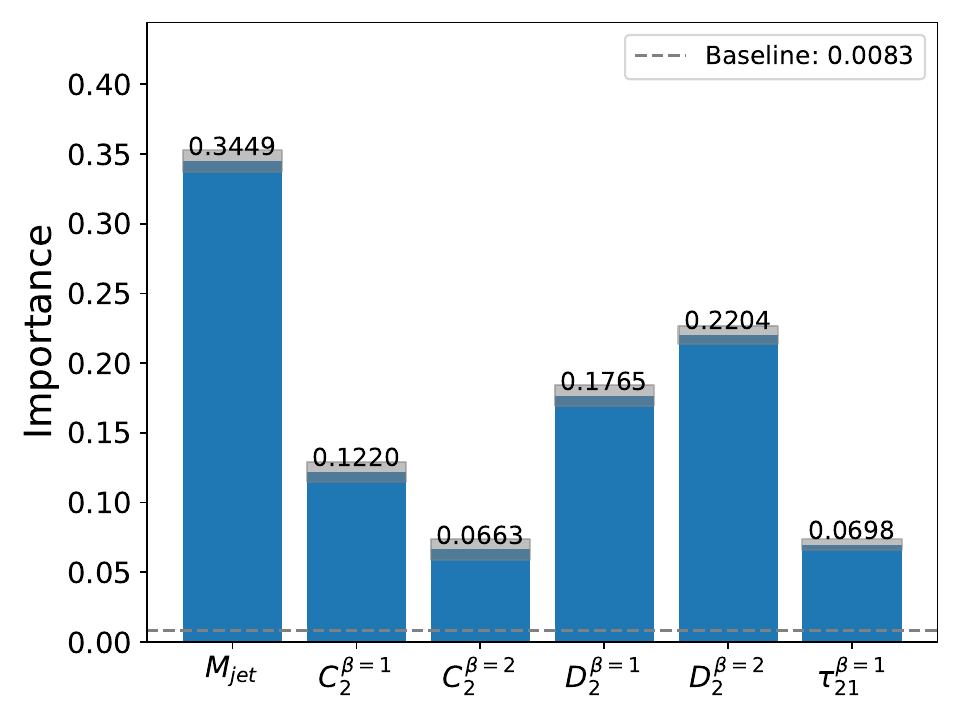}
  \includegraphics[width=0.45\textwidth]{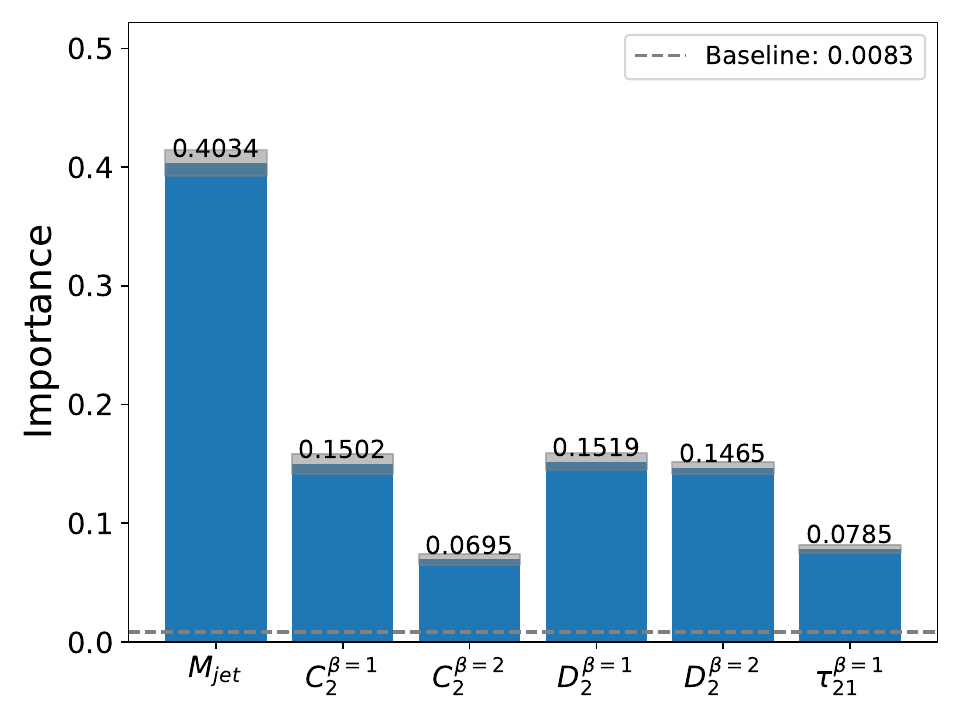}
  \caption{Learned importance from the parallel (left) and sequential (right) LCFs on the diboson dataset. The baseline indicates the minimum importance (default: 5\% of average importance = $1/F \times 0.05$), below which the features are ignored during inference.}
  \label{figure:learned_importance-real1}
\end{figure}

Finally, we evaluate the LCF model on the real diboson dataset in both strategies and also compare its performance with other baseline models. The learned cuts are shown in figure~\ref{figure:learned_cuts-real1-lcf_par} (parallel LCF) and figure~\ref{figure:learned_cuts-real1-lcf_seq} (sequential LCF). The corresponding learned importance scores are shown in figure~\ref{figure:learned_importance-real1}. The performance is summarized in table~\ref{table:metrics_diboson}.

\begin{table}[htbp]
  \centering
  \begin{tabular}{|l|c|c|c|c|c|}
    \hline
    \textbf{Model} & \textbf{TP} & \textbf{FP} & \textbf{Accuracy} & \textbf{Precision} & \textbf{Significance} \\
    \hline
    BDT & $41172 \pm 0$ & $6143 \pm 0$ & $87.5 \pm 0.0\%$ & $87.0 \pm 0.0\%$ & $6.48 \pm 0.00$ \\
    \hline
    MLP & $40819 \pm 372$ & $5480 \pm 285$ & $87.9 \pm 0.1\%$ & $88.2 \pm 0.4\%$ & $6.81 \pm 0.12$ \\
    \hline
    LCF (Parallel) & $24392 \pm 62$ & $2838 \pm 12$ & $69.7 \pm 0.1\%$ & $89.6 \pm 0.0\%$ & $5.65 \pm 0.01$ \\
    \hline
    LCF (Sequential) & $40491 \pm 29$ & $6619 \pm 31$ & $86.0 \pm 0.0\%$ & $86.0 \pm 0.0\%$ & $6.14 \pm 0.01$ \\
    \hline
  \end{tabular}
  \caption{Performance metrics of different models on the diboson dataset.}
  \label{table:metrics_diboson}
\end{table}

In the parallel strategy, the learned cuts are intuitively placed at the boundaries where signal and background distributions diverge, effectively separating the two classes across all observables. In contrast, in the sequential strategy, the first two cuts on $M_{\mathrm{jet}}$ and $C_2^{\beta=1}$ filter out most background events. This causes the signal and background to almost completely overlap in the distributions of the later observables ($C_2^{\beta=2}$, $D_2^{\beta=1}$, $D_2^{\beta=2}$, $\tau_{21}^{\beta=1}$), making their cuts less effective. To confirm whether these overlapped observables are truly negligible, we examine the learned importance scores. In both strategies, all observables are assigned fair importance scores, with none close to zero. Comparing the scores, $D_2^{\beta=2}$ ranks second in the parallel strategy and is more important than $D_2^{\beta=1}$, which aligns with the smaller overlap area in its distribution. In the sequential strategy, this trend persists, though the importance of later observables is generally reduced. Notably, $\tau_{21}^{\beta=1}$ is equally important as $C_2^{\beta=2}$ and non-zero in both strategies even if its cut covers the full range in the sequential strategy. According to eq.~(\ref{eq:learning_process_of_importance}) and the results from Mock3 (redundant features) and Mock4 (high correlated features), importance scores tend toward zero when features generate relatively high loss compared to others (like $x_7$ and $x_8$) or when highly correlated features have decreasing importance based on their application order. $\tau_{21}^{\beta=1}$ represents a different case: its signal-background separation is at a similar level to other observables, and during sequential LCF training, it generates moderate loss compared to truly redundant features. Additionally, since it is not highly correlated with other observables, its importance score does not decrease like $D_2^{\beta=1}$ and $D_2^{\beta=2}$.

The training time of the LCF model is at the same level as the BDT and MLP models, which is lightweight to train compared to modern deep neural networks. The significance of the LCF models, however, is slightly worse than the BDT and MLP models because they are simpler and shallower networks essentially. The parallel strategy this time doesn't outperform the sequential strategy in terms of significance. The parallel strategy's more conservative cut boundaries, while effective in the Mock datasets, here incur signal efficiency costs that outweigh the background reduction benefits in this diboson dataset. This demonstrates the fundamental trade-off inherent in the parallel strategy's approach. From this perspective, the sequential strategy is always more stable (lower variance of metrics across training runs) and more accurate. However, the parallel strategy gains advantages due to the cuts that achieve lower classification accuracy---it always filters out more events. Combined with results from Mock datasets, such inaccuracy doesn't lead to a consistent significance improvement. The sequential strategy is a better choice as a classifier, just like all other neural networks that are trained to reduce the loss designed for classification. On the other hand, the parallel strategy tends to achieve better significance values in most cases, although the optimal choice depends on the specific dataset characteristics. Both strategies can be quickly validated in LCF and the better one can be chosen.

\section{Conclusion}
\label{section:conclusion}
In this work, we introduce the Learnable Cut Flow (LCF), a novel neural network framework that bridges the interpretability of traditional cut-based methods with the adaptability of modern machine learning techniques in high-energy physics. By transforming the manual process of cut selection into a fully differentiable, data-driven operation, LCF offers a transparent and actionable alternative to opaque black-box models. Two widely-used strategies are also implemented via mask operations making LCF a fully interpretable model. We also propose the Learnable Importance to dynamically adjust each observable's contribution to the final classification results.

Testing on six synthetic mock datasets and a realistic diboson vs. QCD dataset demonstrated LCF's versatility and effectiveness. The model accurately identifies optimal cut boundaries across a range of feature distributions, including left, right, middle, and edge cases, while dynamically adapting to varying degrees of signal-background separation. The Learnable Importance mechanism successfully assigns higher weights to features with stronger discriminative power, effectively suppresses redundant or highly correlated observables, and remains robust to feature ordering. Although the LCF is less performant than the baseline models on the diboson dataset, it is as lightweight to train as they are, requiring fewer computational resources than modern deep neural networks. This enables the full set of optimized cuts to be obtained with reduced analyst effort and at minimal cost.

LCF's ability to emulate human cut-searching strategies while embedding them within a trainable framework marks a significant step forward. It retains the simplicity and physicist-friendly insights of traditional cut flows, yet harnesses the power of neural networks to handle complex, high-dimensional data. This hybrid approach not only enhances classification performance but also provides interpretable outputs---cut boundaries and importance scores---that can guide observable selection in real-world analyses. Future extensions of LCF could include optimization directly against signal significance---particularly valuable for BSM searches with low cross sections---and a mechanism to learn optimal feature ordering, enabling full automation while preserving interpretability.

\acknowledgments
Hao Sun is supported by the National Natural Science Foundation of China (Grant No. 12075043, No. 12147205).

\bibliographystyle{JHEP}
\bibliography{biblio.bib}

\end{document}